\documentclass{article}

\usepackage{arxiv}

\usepackage[utf8]{inputenc}
\usepackage[T1]{fontenc}

\usepackage{amsmath, amssymb, amsfonts}
\usepackage{bm}

\usepackage{graphicx}
\usepackage{subcaption}
\usepackage{caption}
\usepackage{wrapfig}
\usepackage{adjustbox}

\usepackage{booktabs}
\usepackage{tabularx}
\usepackage{array}

\usepackage{algorithm}
\usepackage{algorithmic}

\usepackage{url}
\usepackage{nicefrac}
\usepackage{microtype}
\usepackage{xcolor}
\usepackage{comment}
\usepackage{cancel}
\usepackage{placeins}

\usepackage[numbers]{natbib}

\usepackage{hyperref}
\hypersetup{
  colorlinks=true,
  linkcolor=blue,
  citecolor=blue,
  urlcolor=blue,
  filecolor=blue,
}

\title{DynaTab: Dynamic Feature Ordering as Neural Rewiring for High-Dimensional Tabular Data\thanks{
This paper has been accepted for archival publication in the PMLR proceedings of the AAAI 2026 Neuro for AI \& AI for Neuro: Towards Multi-Modal Natural Intelligence (NeuroAI) Workshop,
Code: \url{https://github.com/zadid6pretam/DynaTab},
PyPI: \texttt{pip install dynatab}.
}}

\author{
Al Zadid Sultan Bin Habib\textsuperscript{1}
\quad
Gianfranco Doretto\textsuperscript{2}
\quad
Donald A. Adjeroh\textsuperscript{1}
\\[4pt]
\textsuperscript{1}Lane Department of Computer Science and Electrical Engineering,\\
West Virginia University, Morgantown, WV 26506, USA
\\
\textsuperscript{2}Scientific Computing and Imaging Institute \& Department of Biomedical Informatics,\\
The University of Utah, Salt Lake City, UT 84112, USA
\\[4pt]
\texttt{ah00069@mix.wvu.edu}
\quad
\texttt{doretto@utah.edu}
\quad
\texttt{donald.adjeroh@mail.wvu.edu}
}

\usepackage{bibentry}
\begin{document}

\maketitle

\begin{abstract}
High-dimensional tabular data lacks a natural feature order, limiting the applicability of permutation-sensitive deep learning models. We propose DynaTab, a dynamic feature ordering-enabled architecture inspired by neural rewiring. We introduce a lightweight criterion that predicts when feature permutation will benefit a dataset by quantifying its intrinsic complexity. DynaTab dynamically reorders features via a neural rewiring algorithm and processes them through a compact, dynamic order-aware combination of separate learned positional embedding, importance-based gating, and masked attention layers, compatible with any sequence-sensitive backbone. Trained end-to-end with bespoke dynamic feature ordering (DFO) and dispersion losses, DynaTab achieves statistically significant gains, particularly on high-dimensional datasets, where it is benchmarked against 45 state-of-the-art baselines across 36 different real-world tabular datasets. Our results position DynaTab as a compelling new paradigm for high-dimensional tabular deep learning.
\end{abstract}


%\marginpar{\textcolor{red}{\hspace{1.5em} Pls update}}

\section{Introduction}
\label{sec:intro}
Developing end-to-end deep learning models for high-dimensional tabular data remains challenging due to its lack of inherent structure, unlike image or text modalities. Traditional architectures such as convolutional neural networks (CNNs) and transformers perform well in vision and natural language processing (NLP) tasks, but often underperform on tabular tasks, especially with high-dimensional data~\cite{b28}. Tree ensembles are strong baselines; deep tabular models improve with imbalance-aware training~\cite{tdl}. Models like TabNet~\cite{b1}, TabTransformer~\cite{b3}, and FT-Transformer~\cite{b5} attempt to address these issues, yet robust solutions are still limited. TabSeq~\cite{b6} introduced feature ordering to impose structure, but its fixed order cannot generalize across datasets. Mambular~\cite{mambular} highlighted the sensitivity of model behavior to random input orderings. This issue is also seen in large language models (LLMs), where sequence affects predictions in reasoning and question answering (QA) tasks~\cite{order1}. In parallel, methods such as ProtoGate~\cite{protogate} and Deep Neural Pursuit (DNP)~\cite{dnp} explored data-driven feature selection for High-Dimensional Low-Sample Size (HDLSS) contexts. These developments motivate the Column Permutation Problem (CPP) as a core challenge in tabular deep learning~\cite{cpp} for high-dimensional  datasets, supporting dynamic feature ordering with order-aware fusion as a promising solution and foundation of DynaTab.

Neuroplasticity, the brain's capacity to reorganize and adapt its neural connections~\cite{b7,b8,b9} provides a powerful model for developing flexible approaches in deep learning, such as DynaTab’s dynamic feature ordering. Through synaptic plasticity~\cite{b10,b11}, the brain strengthens or weakens connections based on stimulus relevance, allowing neurons to rewire in response to learning, experience, or recovery~\cite{b23,b24}. Similarly, structural plasticity~\cite{b13,b14,b15}, the formation and pruning of connections, enables the brain to optimize processing by reconfiguring networks for specific tasks. Mechanisms like Long-Term Potentiation (LTP) and Long-Term Depression (LTD) regulate synaptic strength~\cite{b12}, reinforcing frequently co-activated neural pathways to enhance memory and learning. Principles from Hebbian learning~\cite{b19} and Spike-Timing-Dependent Plasticity (STDP)~\cite{b18} further refine this adaptability by adjusting connection strength based on the timing of neuronal spikes, optimizing synaptic efficiency.

Feature ordering has long history in pattern recognition and is central to Incremental Attribute Learning (IAL), where features arrive sequentially and must be ranked before training~\cite{f1}. Unlike set-based treatments that assume order invariance~\cite{deepsets}, column order in practice shapes redundancy exposure and impacts our ability to capture dependencies. Empirical studies show that Fisher/correlation/entropy ranks reduce interference and error compared to unordered baselines~\cite{f2,f4}, motivating learned, task-aware ordering in machine learning~\cite{f3,f7}. Recent work highlights that tabular models can be brittle to column permutations and thus enforce order-agnostic (permutation-invariant) representations to neutralize the impact of feature ordering~\cite{ack1,ack2}. In deep tabular learning, Mambular~\cite{mambular} highlighted the importance of feature ordering, and TabSeq~\cite{b6} introduced an explicit ordering algorithm. ROTATOR-LLM~\cite{wang2025advancing} investigates feature ordering for LLM-based tabular inference. TabICL~\cite{tabicl} makes predictions over multiple different column permutations and then combines these results to restore the desired invariance. In many real-world HDLSS tasks, even simple models, such as MLP and Lasso, can outperform advanced methods (see ProtoGate~\cite{protogate}). This observation underscores that selection alone is not sufficient in \(n \ll m\) regimes (here $n$=number of samples, $m$=number of features), and that the permutation itself should be a learnable object. We therefore pose the CPP as a core challenge for deep tabular learning for high-dimensional settings to learn a data-driven feature order that reduces redundancy, exposes long-range dependencies, and provides sequence structure to downstream modules. Practically, this can be approached with attention-based pointer mechanisms and graph-aware extensions that generate permutations while encoding relational structure~\cite{pointer,gpn,pgn}. Analogous reordering ideas appear in hyperspectral compression schemes that reorder bands to expose correlations~\cite{b_54, jain2007edge} and in graph/communication layouts that minimize pairwise interaction costs~\cite{b_55}.

Inspired by these adaptive processes, we propose DynaTab, a novel deep learning model specifically designed to optimize feature ordering for high-dimensional complex tabular data. This model introduces a dynamic feature ordering approach inspired by neural rewiring principles, enabling adaptive feature arrangement that responds to data complexity. DynaTab emulates neural rewiring~\cite{b20,b21,b22} through dynamic feature ordering, adjusting feature connectivity and relevance in response to data complexity. Just as local plasticity~\cite{b16,b17} allows clusters of neurons to adapt and improve task-specific performance, DynaTab reorders feature graphs based on node centrality to optimize local relationships, creating a context-aware, flexible approach to enhance learning on complex, high-dimensional tabular data. This biologically inspired model adaptation allows tabular deep learning models to mimic the brain's adaptability in recognizing patterns and making decisions, underscoring dynamic feature ordering as an effective tool for handling heterogeneous datasets. Our key contributions are as follows:
\begin{enumerate}
  \item We predict when feature ordering offers major gains by quantifying dataset complexity and ordering sensitivity.
  \item We propose dynamic feature ordering (DFO) inspired by neural rewiring, adaptively reordering columns to optimize inter‑feature connectivity per dataset.
  \item We introduce DynaTab, an end-to-end model that integrates an order‑aware fusion of positional embeddings, importance gating, and dynamic masked attention with a plug‑and‑play sequential backbone (e.g., Transformer, DAE, LSTM, or Mamba), trained jointly under DFO and dispersion losses.
  \item We introduce a taxonomy for tabular datasets based on their sample size and feature dimensionality, and thus empirically group  tabular datasets into 5 categories. Then, we  benchmark DynaTab against 45 baselines on 36 tabular datasets of different kinds, using Friedman tests with Critical Difference (CD) diagrams and Wilcoxon Holm post hoc analysis to demonstrate significant, consistent improvements in classification and regression.  
\end{enumerate}
These contributions together tackle varying feature relevance and dataset complexity, setting a new benchmark for adaptive feature ordering and end‐to‐end model optimization in tabular deep learning.
%%%%%
\section{Related Work}
\label{sec:related}
Deep tabular models face challenges due to the lack of spatial or sequential structure. Early architectures such as TabNet~\cite{b1}, TabTransformer~\cite{b3}, and FT-Transformer~\cite{b5} employ attention-based mechanisms to model feature interactions. Tree-based models like NODE~\cite{b2} and probabilistic transformers like TabPFN~\cite{b4} achieve strong performance but face scalability issues ($\le$1000 samples). Ensemble models like XGBoost, LightGBM, and CatBoost remain competitive baselines in many scenarios. 

Ordering-based models directly motivate our approach. TabSeq~\cite{b6} sequences features by relevance; Mambular~\cite{mambular} studies column permutations with state-space layers; ProtoGate~\cite{protogate} uses prototype-guided selection in HDLSS settings. Pointer Networks~\cite{pointer} and extensions like GPN~\cite{gpn} explore differentiable output permutations. Models such as TANGOS~\cite{b42}, NDTF~\cite{ndtf}, and TabPFN v2~\cite{tabpfnv2} introduce selectors, hybrid modules, or generative pretraining. Sequential backbones are increasingly explored. MambaTab~\cite{mambatab} and MambAttention~\cite{mamb}, Mamba-based variants use state-space modeling; TabulaRNN~\cite{mamb}, Trompt~\cite{trompt}, and ModernNCA~\cite{modern} apply recurrence, memory, or attention pooling to structured inputs. ResNet-style MLPs~\cite{b5}, TabR~\cite{tabr}, and TabM~\cite{tabm} stack lightweight modules with residual connections for improved generalization. TabICL~\citep{tabicl} uses RoPE and combines column permutations to restore invariance.

Instance-wise feature selection complements our ordering approach. Models like INVASE~\cite{invase}, STG~\cite{stg}, and LSPIN/LLSPIN~\cite{spin} dynamically gate features per sample, supporting interpretability. Broader developments inform our context. PLATO~\cite{b28} proposes a graph-based meta-learning framework for tabular transfer learning. TabDDPM~\cite{tabddpm} introduces diffusion models for high-fidelity tabular synthesis. TabReD~\cite{tabred} presents temporally-split benchmarks, highlighting gaps between gradient boosted decision trees (GBDTs) and deep models. T2G-Former~\cite{b33} and HYTREL~\cite{b32} use graph-augmented transformers to model feature relationships and support structure-aware learning. These works underscore the ongoing need for flexible, data-adaptive solutions. RKNN-FS~\cite{fs1} presented a feature selection-based model for HDLSS problems, building on a proposed random $k$-nearest neighbor (RKNN) algorithm.

\begin{figure}[t]
  \centering
  \includegraphics[width=\columnwidth]{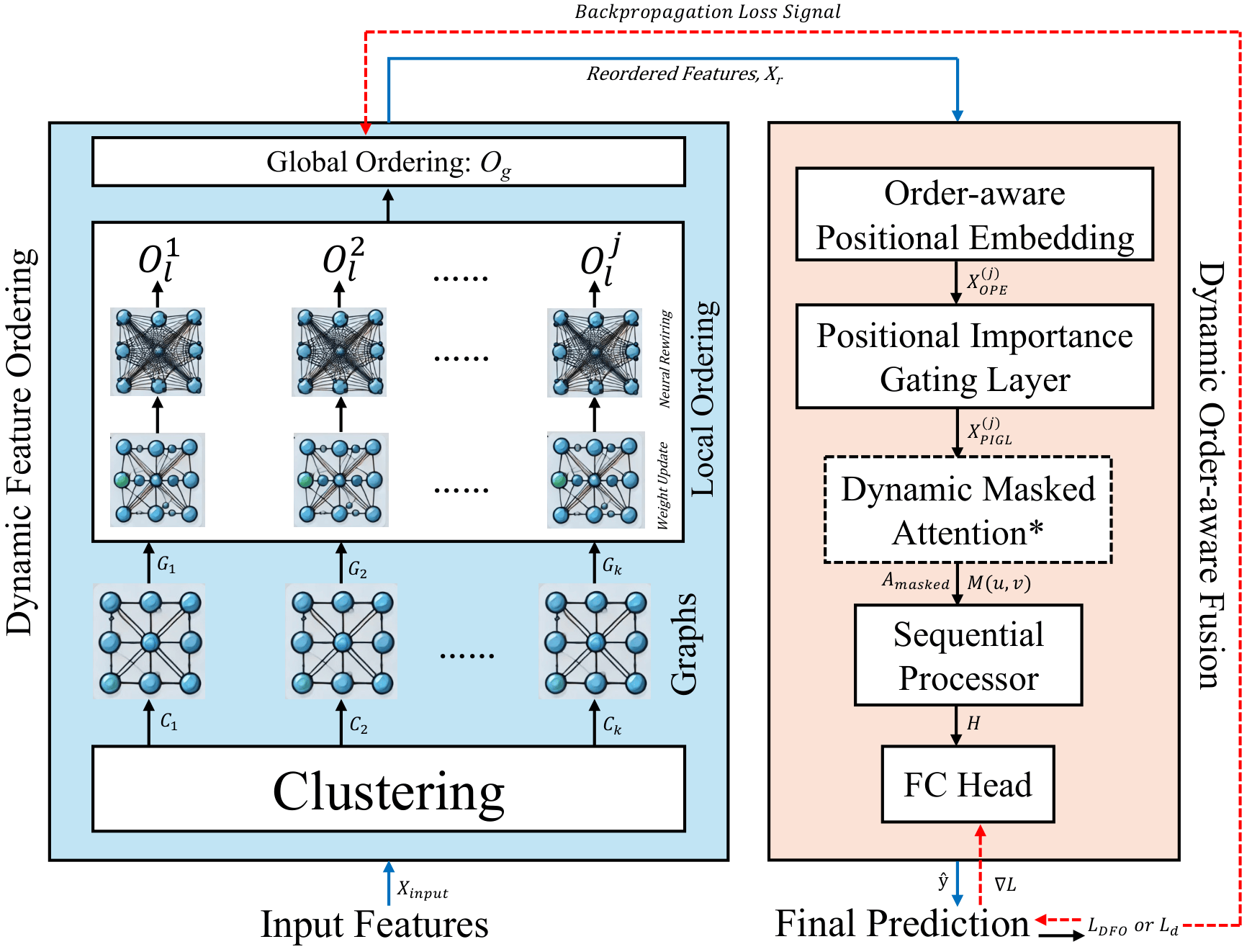}
  \caption{%
    End-to-end DynaTab. Left (light blue): Dynamic Feature Ordering produces \( X_r \) (feature graphs drawn using GPT-4o~\cite{b78}). Right (light peach): Order-aware Fusion (OPE + PIGL), DMA\textasteriskcentered, and a sequential backbone. Solid arrows: data flow; dashed red arrows: gradient flow. \textasteriskcentered\ DMA applies only to attention-based backbones.%
  }
  \label{fig:dynatabseq_arch}
\end{figure}

Our DynaTab framework (Figure~\ref{fig:dynatabseq_arch}) introduces dynamic feature ordering inspired by neural rewiring. It integrates an order-aware fusion block (OPE, PIGL, DMA) with sequential backbones (Transformer, LSTM, DAE, or Mamba), trained end-to-end using DFO and Dispersion losses. This enables context-sensitive learning across classification, regression, and coherence tasks on complex tabular datasets.
%%%%%%%%%%%
\section{Methodology}
%%%%
\begin{figure}[t]
    \centering
    \begin{subfigure}[t]{0.48\linewidth}
        \centering
        \includegraphics[width=\linewidth]{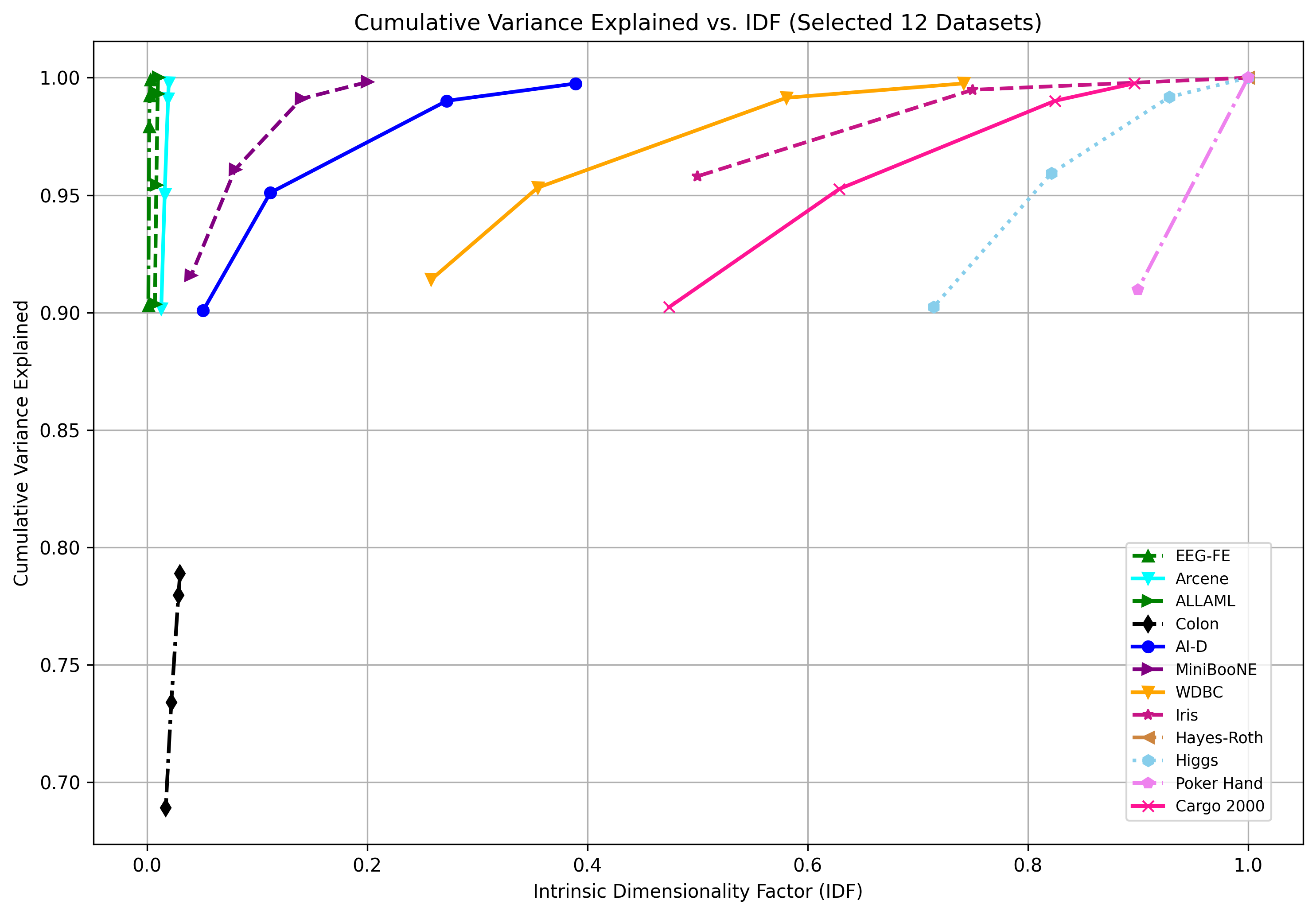}
        \caption{Cumulative Variance vs. IDF}
        \label{fig:foe1A}
    \end{subfigure}%
    \hfill
    \begin{subfigure}[t]{0.48\linewidth}
        \centering
        \includegraphics[width=\linewidth]{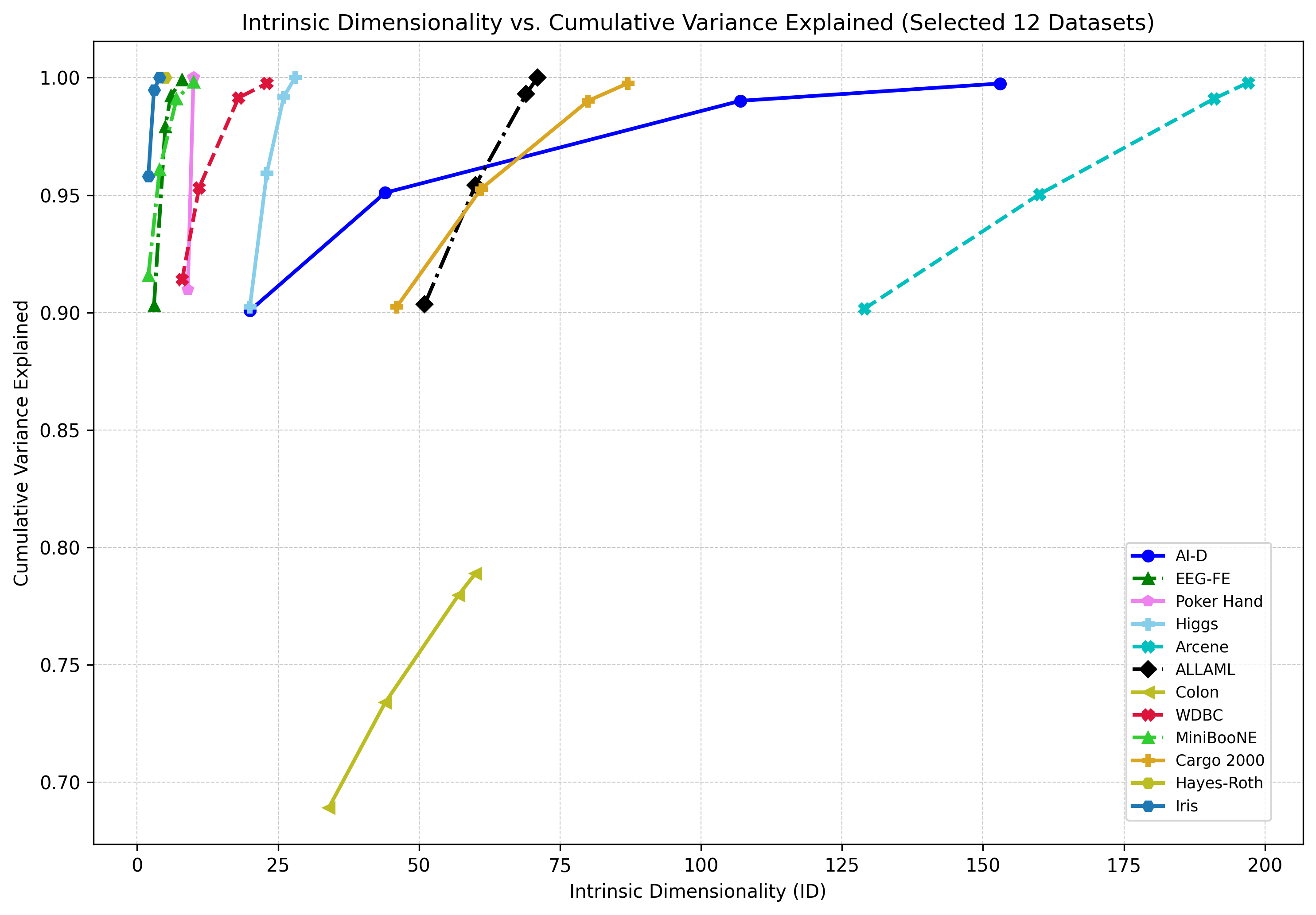}
        \caption{ID vs. Cumulative Variance}
        \label{fig:foe2A}
    \end{subfigure}
    \caption{Relationship between cumulative variance and intrinsic dimensionality across 12 selected datasets. See supplementary material for details on all 36 datasets used. %\textcolor{red}{Remove the "Cumulative Variance: " in the legends !!} >>>> removed
    }
    \label{fig:foe_combinedA}
    \vspace{-10pt} % Adjust spacing as needed
\end{figure}
%%%%%%%%
\subsection{Feature Ordering - When to Use?}
\label{sec3.1}
The Intrinsic Dimensionality Factor (IDF) estimates the benefit of feature ordering by comparing a dataset’s intrinsic dimensionality i.e., the minimal number of features capturing core variability~\cite{b44} to its total feature count. IDF is defined as the ratio of intrinsic to total feature count, 
$\text{IDF} = n_{\text{intrinsic}} / n_{\text{total}}$, where \( n_{\text{intrinsic}} \) is estimated using methods such as PCA. The dataset's complexity is then measured as:
\begin{equation}
\label{eq2}
\text{Complexity Score} = \frac{\text{Cumulative Variance at IDF}}{\text{IDF}^s}
\end{equation}
where \( s \) is a tunable sensitivity parameter. The Feature Ordering Effectiveness (FOE) quantifies the gain from ordering:
\begin{equation}
\label{eq3}
\text{FOE} = \frac{\psi}{\left( \text{AUC} \times \text{IDF} \right)^s}
\end{equation}
Here, \( \psi \) is a dataset-specific scaling factor, and AUC is the area under the IDF-variance curve. The loss function below optimizes \( \psi \) by minimizing the gap between the computed FOE and the target value (set to 1):
\begin{equation}
\label{eq4}
\text{Loss}(\psi) = \left( \frac{\psi}{(\text{AUC})^s} - 1 \right)^2
\end{equation}
Setting \( s = 2 \) introduces quadratic sensitivity~\cite{b52}, increasing the penalty for low-variance datasets. AUC is estimated using the trapezoidal rule~\cite{b53} for efficient integration of discrete IDF–variance pairs. While \( \psi \) and AUC vary across datasets, FOE maintains an inverse relationship with IDF (Equation~\ref{eq3}), making it a reliable proxy for ordering effectiveness.
%in Equation \ref{eq5}:
\begin{equation}
\label{eq5}
\text{FOE} \propto \frac{1}{\text{IDF}} \quad \text{(for fixed } \psi \text{ and AUC)}
\end{equation}
Although \(\psi\) and AUC vary, FOE consistently decreases with increasing IDF (Equation~\ref{eq3}). A lower IDF indicates higher redundancy, 
%and complexity,
making feature ordering more effective. In contrast, higher IDF reflects better-structured datasets with limited gain from ordering. Thus, the success probability \( P_{\text{success}} \) is inversely related to IDF (Eq.~\ref{eq6}) which yields in Eq.~\ref{eq7}.
\begin{equation}
\label{eq6}
\text{FOE} \propto \frac{1}{\text{IDF}} \propto P_{\text{success}}
\end{equation}
%%%%%%
\begin{equation}
\label{eq7}
P_{\text{success}} = 1 - \text{IDF} = 1 - \frac{n_{\text{intrinsic}}}{n_{\text{total}}}
\end{equation}
As intrinsic dimensionality approaches the total, \( P_{\text{success}} \) drops; lower intrinsic dimensionality suggests greater ordering benefit (Equations~\ref{eq6},~\ref{eq7}). Figure~\ref{fig:foe_combinedA} (see also  Supplementary Material) shows the cumulative variance–IDF relationship. Figure~\ref{fig:foe1A} illustrates that datasets with low IDF reach high cumulative variance quickly, indicating redundancy. Figure~\ref{fig:foe2A} presents absolute intrinsic dimensionality across datasets. See the supplementary material for FOE-based dataset rankings and further analysis.
\subsection{Dynamic Feature Ordering as Neural Rewiring}
Feature ordering aims to find an optimal feature arrangement within and across clusters that minimizes disorganization~\cite{b6}. For a dataset \( X \in \mathbb{R}^{n \times m} \), we define graphs \( \{G_1, \dots, G_k\} \) for \( k \) clusters, where each \( G_j = (V_j, E_j) \) represents cluster \( j \), with vertices \( u \in V_j \) as features and edges \( (u, v) \in E_j \) weighted by relationship strength \( w(u, v) \). The optimal local ordering \( \mathcal{O}_{\text{l}}^j \) minimizes intra-cluster dispersion:
\begin{equation}
\label{eq8}
F_j(\mathcal{O}_{\text{l}}^j) = \arg \min_{\mathcal{O}_{\text{l}}^j} \sum_{(u, v) \in E_j} w(u, v) \cdot | \mathcal{O}_{\text{l}}^j(u) - \mathcal{O}_{\text{l}}^j(v) |
\end{equation}
where \( \mathcal{O}_{\text{l}}^j(u) \) is the position of feature \( u \) in \( \mathcal{O}_{\text{l}}^j \). The global ordering \( \mathcal{O}_{\text{g}} \) aggregates local orderings to minimize a weighted global dispersion cost (Eq.~\ref{eqb}) and the optimal global ordering is then obtained (Eq.~\ref{eq10}).
\begin{equation}
\label{eqb}
G(\mathcal{O}_{\text{g}}) = \sum_{j=1}^{k} \alpha_j \cdot F_j(\mathcal{O}_{\text{l}}^j), \quad \sum_{j=1}^{k} \alpha_j = 1
\end{equation}
%%%%%%
\begin{equation}
\label{eq10}
\mathcal{O}_{\text{g}}^* = \arg \min_{\mathcal{O}_{\text{g}}} G(\mathcal{O}_{\text{g}})
\end{equation}
\textbf{Local Ordering with Neural Rewiring:}  
Local ordering minimizes intra-cluster dispersion by optimizing \( F_j(\mathcal{O}_{\text{l}}^j) \) (Equation~\ref{eq8}). To guide this process, we compute centrality scores e.g., degree, betweenness, or eigenvector centrality~\cite{b54, b55} which quantify a feature’s influence in the graph. High-centrality features are prioritized by strengthening their connections and pruning weaker ones, simulating structural plasticity~\cite{b13, b14}. Edge weights are then updated based on feature centrality, producing an optimized feature graph and revised ordering \( \mathcal{O}_{\text{l}}^j \), in line with synaptic plasticity~\cite{b10, b11}. Inspired by Hebbian learning~\cite{b19}, we model edge weight updates as:
\begin{equation}
w_{uv}^{(t+1)} = w_{uv}^{(t)} + \Delta w_{uv}
\end{equation}
where \( \Delta w_{uv} \) reflects centrality-driven adjustment:
\begin{equation}
\label{eq:weight_adjustment}
\Delta w_{uv} = \lambda C(u) C(v) - \epsilon w_{uv}^{(t)}
\end{equation}
Here, \( \lambda \) is the learning rate, \( \epsilon \) a regularization term, and \( C(u), C(v) \) the centrality scores of features \( u, v \) respectively. This promotes connections between influential features while suppressing less relevant ones~\cite{b56, b57}, enhancing information flow and structural coherence. As in neuroplasticity~\cite{b64}, this rewiring reinforces key connections to minimize dispersion (Equation~\ref{eq8}) and align clusters into a cohesive global order (Equation~\ref{eq10}).

%\\
\textbf{Local Ordering and Quality Maximization:}  
The rewiring process aims to maximize a local quality metric \( Q \), halting when further updates yield minimal improvement. At each step \( t \), edge weights are updated as:
\begin{equation}
w_{uv}^{(t+1)} = w_{uv}^{(t)} + \Delta w_{uv} \quad \text{such that} \quad Q(\mathcal{O}_{\text{l}}^{(t+1)}) \geq Q(\mathcal{O}_{\text{l}}^{(t)})
\end{equation}
Iterations stop when the quality change drops below a threshold \( \epsilon \), indicating convergence:
\begin{equation}
\left| \frac{\partial Q}{\partial \mathcal{O}_{\text{l}}^{(t)}} \right| = |Q(\mathcal{O}_{\text{l}}^{(t+1)}) - Q(\mathcal{O}_{\text{l}}^{(t)})| < \epsilon
\end{equation}
The threshold \( \epsilon \) controls early stopping and defines when additional rewiring becomes non-beneficial.\\
\textbf{Edge Pruning and Rewiring:}  
Edges with weights below a threshold \( \theta \) are pruned:
\begin{equation}
w_{uv}^{(t+1)} = 0 \quad \text{if} \quad w_{uv}^{(t)} < \theta
\end{equation}
New connections are then formed between high-centrality nodes \( u', v' \) as:
\begin{equation}
w_{u'v'}^{(t+1)} = \lambda C(u') C(v') - \epsilon w_{u'v'}^{(t)}
\end{equation}
This rewiring promotes high-centrality features within each local ordering \( \mathcal{O}_{\text{l}}^j \), guiding the final global ordering \( \mathcal{O}_{\text{g}} \).\\
\textbf{Quality Metric \( Q \):}  
We define a general metric \( Q \) to evaluate how well a feature ordering aligns with pairwise relationships captured by an edge metric \( \zeta(u, v) \), which may represent KL divergence, Euclidean or Manhattan distance, variance difference, or correlation-based dissimilarity (see ablation studies in supplementary). In the feature graph \( G_j \) for cluster \( C_j \), each edge \( (u, v) \in E_j \) is weighted by \( \zeta(u, v) \). The quality of a local ordering \( \mathcal{O}_{\text{l}}^j \) is:
\begin{equation}
Q(\mathcal{O}_{\text{l}}^j) = \sum_{(u,v) \in E_j} \zeta(u, v) \cdot d_{\mathcal{O}_{\text{l}}^j}(u, v)
\end{equation}
\begin{equation}
d_{\mathcal{O}_{\text{l}}^j}(u, v) = \big|\mathcal{O}_{\text{l}}^j(u) - \mathcal{O}_{\text{l}}^j(v)\big|
\end{equation} 
%\marginpar{\textcolor{red}{$\mathcal{O}()$ same as $\mathcal{O}_l()$?? Also chk Alg1!! And nxt eqn too!!}} >>>> fixed this one and algorithm 1 also
This encourages similar features (low \( \zeta(u, v) \)) to appear closer in the ordering. The system iteratively rewires the graph and updates \( \mathcal{O}_{\text{l}}^j \) to reduce  \( Q(\mathcal{O}_{\text{l}}^j) \), accepting changes that lower total dissimilarity. Algorithm~\ref{algo1} outlines the neural rewiring procedure for local ordering.
%%%%%
\begin{algorithm}[t]
\caption{Neural Rewiring for Local Ordering}
\begin{algorithmic}[1]
\label{algo1}
\STATE \textbf{Input:} Feature graph $G_j = (V_j, E_j)$, quality metric $Q$, mutation probability $p$, tolerance $\epsilon$, pruning threshold $\theta$
\STATE \textbf{Output:} Rewired local feature ordering $\mathcal{O}_{\text{l}}^j$
\STATE Initialize $\mathcal{O}_{\text{l}}^j = \arg \min_{\mathcal{O}} \sum_{(u,v) \in E_j} w(u, v) \cdot | \mathcal{O}_{\text{l}}^j(u) - \mathcal{O}_{\text{l}}^j(v) |$
\STATE Calculate centrality $C(v)$ for each node $v \in V_j$

\WHILE{stopping criteria not met}
    \STATE Compute current quality $Q_{\text{c}} = Q(\mathcal{O}_{\text{l}}^j)$
    \FOR{each edge $(u,v) \in E_j$}
        \IF{$w_{uv}^{(t)} < \theta$} 
            \STATE Prune edge $(u,v)$ and rewire based on centrality of $u'$ and $v'$:
            $
            w_{u'v'}^{(t+1)} = \lambda C(u') C(v') - \epsilon \cdot w_{u'v'}^{(t)}
            $
        \ENDIF
    \ENDFOR
    \STATE Update $\mathcal{O}_{\text{l}}^{j, (t+1)}$ based on rewired graph
    \IF{$|\Delta Q| < \epsilon$} 
        \STATE Stop rewiring
    \ENDIF
\ENDWHILE

\STATE Return $\mathcal{O}_{\text{l}}^j$
\end{algorithmic}
\end{algorithm}
%second algorithm
\begin{algorithm}[t]
\caption{Dynamic Feature Ordering}
\label{algo2}
\begin{algorithmic}[1]
\STATE \textbf{Input:} Dataset $X$, clustering algorithm, number of clusters $k$, edge metric, sorting order, tolerance $\epsilon$, mutation probability $p$
\STATE \textbf{Output:} Global feature ordering $\mathcal{O}_{\text{g}}$
\STATE Cluster data (e.g., KMeans) with clusters $C_j$ and centroids $\mu_j$:
$
c_i = \arg \min_j \| x_i - \mu_j \|^2, \quad 
\mu_j = \frac{1}{|C_j|} \sum_{x_i \in C_j} x_i
$
\STATE Build feature graph $G_j = (V_j, E_j)$ for each cluster $C_j$
\FOR{each cluster $C_j$}
    \STATE Apply Neural Rewiring to get local ordering $\mathcal{O}_{\text{l}}^j$
    \STATE Mutate ordering with probability $p$ if $|\Delta Q| < \epsilon$
\ENDFOR
\STATE Compute cluster importance $\alpha_j$ such that $\sum_{j=1}^{k} \alpha_j = 1$
\STATE Integrate local orderings into global ordering:
$
\mathcal{O}_{\text{g}} = \arg \min_{\mathcal{O}_{\text{g}}} 
\sum_{j=1}^{k} \alpha_j \cdot F_j(\mathcal{O}_{\text{l}}^j)
$
\STATE \textbf{Return} $\mathcal{O}_{\text{g}}$
\end{algorithmic}
\end{algorithm}
%%%%
% top m feature selection adaptively like protogate
% but somewhere messed up while changing the code which is degrading the performance horribly, so going back to older approach and choosing PCA derived IDF features for very high dimensionality
%%%
\begin{comment}
    \textbf{Adaptive Top-\(m\) Feature Selection:} 
To improve efficiency for high-dimensional data, we incorporate an adaptive feature-selection step post neural rewiring-based local ordering. Each feature \( f \) within cluster \( j \) is assigned a learnable gating score \( g_f^{(j)} \in \mathbb{R} \). Using a temperature-controlled softmax (Eq.~\ref{topm}), we approximate selecting the top-\(m\) influential features per cluster from the local orders. Gradual annealing (\(\tau \rightarrow 0\)) makes this selection deterministic, significantly reducing complexity and GPU memory consumption.
\begin{equation}
\label{topm}
p_f^{(j)} = \frac{\exp(g_f^{(j)}/\tau)}{\sum_{f'} \exp(g_{f'}^{(j)}/\tau)}
\end{equation}
\end{comment}
%%%%%

\textbf{Global Ordering:}  
The global ordering \( \mathcal{O}_{\text{g}} \) integrates local cluster orderings to minimize total feature dispersion and improve model performance (Equation~\ref{eq10}). It is defined as:
\[
\mathcal{O}_{\text{g}} = \arg\min_{\mathcal{O}} \sum_{j=1}^{k} \alpha_j \cdot F_j(\mathcal{O}_{\text{l}}^j)
\]
where \( F_j(\mathcal{O}_{\text{l}}^j) \) measures local feature dispersion~\cite{b58, b59}, and \( \alpha_j \) is the weight for cluster \( j \), derived from inter-centroid distances. This formulation encourages cohesion across clusters by emphasizing more densely connected regions of the feature space.

Local dispersion for cluster \( j \) is computed as:
\begin{equation}
D(\mathcal{O}_{\text{l}}^j) = \sum_{(u, v) \in E_j} w(u, v) \cdot |\mathcal{O}_{\text{l}}^j(u) - \mathcal{O}_{\text{l}}^j(v)|
\end{equation}
where \( w(u, v) \) captures the proximity importance between features \( u \) and \( v \), reinforcing related features being placed closer together (see also Equation~\ref{eq8}). An alternate view defines dispersion of feature \( j \) under ordering \( \mathcal{O}_{\text{l}} \) as \( D(\mathcal{O}_{\text{l}}^j) = \sum_{k=1}^{d} W_{jk} \) where \( W_{jk} \) is the entry of the adjacency (distance) matrix. Reducing this dispersion directly reduces feature variance as \( \mathrm{Var}_{j}(X_u) \propto D(\mathcal{O}_{\text{l}}^j) \). Algorithm~\ref{algo2} summarizes the complete dynamic feature ordering process.
%%%
\subsection{Dynamic Order-aware Fusion}
The DynaTab backbone includes a specialized Dynamic Order-aware Fusion (DOF) module that processes globally reordered features through an order-sensitive pipeline. To mitigate permutation invariance and capture ordering effects, DOF incorporates three components before the sequential processor: Order-aware Positional Embedding (OPE), Positional-Importance Gating Layer (PIGL), and Dynamic Masked Attention (DMA), inspired by attention mechanisms, biological gating, and sequence modeling. DOF maintains the inductive bias from DFO and supports robust representation learning via a flexible sequential processor, instantiated as a Transformer, Denoising Autoencoder (DAE), LSTM, hybrid DAE-MHA-LSTM, or Mamba.
%\\

\textbf{Order-aware Positional Embedding (OPE):} Inspired by positional encoding in Transformers~\cite{b66}, OPE injects dynamic positional information into globally reordered tabular features. Given a reordered sequence \( X_r \in \mathbb{R}^m \) from the global ordering \( \mathcal{O}_{\text{g}} \), each feature \( X_r^{(f)} \) is augmented with a learned embedding based on its position:
\begin{equation}
X_{\text{OPE}}^{(f)} = X_r^{(f)} + PE(\mathcal{O}_{\text{g}}(f)), \quad PE(\mathcal{O}_{\text{g}}(f)) \in \mathbb{R}^d
\end{equation}
Here, \( PE(\mathcal{O}_{\text{g}}(f)) \) denotes the positional embedding for feature \( f \), and \( d \) is the embedding dimension. Unlike fixed sinusoidal encodings, OPE learns embeddings that adapt to feature permutations, preserving semantic structure and enhancing downstream performance.
%\\

\textbf{Positional Importance Gating Layer (PIGL):} Inspired by biological gating and attention-free networks~\cite{veness2021gated}, PIGL modulates each feature based on an ordering-derived importance score \( \gamma_f \), computed from metrics like dispersion or ordering weights. A sigmoid-activated gate controls the contribution of each feature:
\begin{equation}
X_{\text{PIGL}}^{(f)} = \sigma(W_g \gamma_f + b_g) \cdot X_{\text{OPE}}^{(f)}
\end{equation}
where \( \sigma \) is the sigmoid function, and \( W_g, b_g \) are learnable gating parameters. This operation is permutation-invariant yet guided by \( \gamma_f \), reinforcing salient features and suppressing less informative ones.
%\\

\textbf{Dynamic Masked Attention (DMA):} Inspired by directional masking in Transformers~\cite{shaw2018self}, DMA enforces attention flow consistent with the global feature order. In attention-based architectures (e.g., Transformer, DAE-MHA-LSTM), a directional mask \( M \in \mathbb{R}^{m \times m} \) is constructed using the global ordering \( \mathcal{O}_{\text{g}} \). For features \( f_p \) and \( f_q \), the mask is defined as:
\begin{equation}
\begin{split}
M_{\mathrm{SW}}(f_p, f_q) = 
\begin{cases}
0, & 
\begin{aligned}
&\text{if } \mathcal{O}_{g}(f_p) \le \mathcal{O}_{g}(f_q) \\
&\quad \text{and } \lvert p - q \rvert \le w,
\end{aligned}\\[6pt]
-\infty, & \text{otherwise.}
\end{cases}
\end{split}
\end{equation} 
%\marginpar{\textcolor{red}{What is $w$??}}
This mask is added to attention logits to block reverse attention:
\begin{equation}
A_{\text{DMA}} = \text{softmax}\left( \frac{QK^\top}{\sqrt{d_k}} + M_{\mathrm{SW}} \right)V
\end{equation}
where \( Q, K, V \) are query, key, and value matrices, and \( d_k \) is the key dimension. We use a full causal mask ($w = m$), so each feature can attend to all earlier ones. DMA enforces order-aware directionality in feature interactions.

\textbf{Sequential Processor and FC Head:} The processor operates on reordered inputs \( X \in \mathbb{R}^{n \times m \times d_{\mathrm{model}}} \), using any backbone (Transformer~\cite{b66}, DAE~\cite{b65}, LSTM~\cite{b67}, DAE-MHA-LSTM, or Mamba~\cite{gu2023mamba}) to produce a pooled output \( \bar{h} \), followed by a lightweight fully connected head for prediction. Training uses class-weighted DFO or dispersion loss, enabling effective modeling even when \( m \gg n \). 

%\\
\textbf{DFO Loss Function:} The DFO loss combines a prediction loss with penalties for local feature smoothness and global ordering coherence. The default prediction term is binary cross-entropy (BCE) for binary classification:
\begin{equation}
L_{\text{b}} = - \frac{1}{N} \sum_{i=1}^{N} \left[ y_i \log(y_i^{\text{pred}}) + (1 - y_i) \log(1 - y_i^{\text{pred}}) \right]
\end{equation}
For multi-class classification, BCE is replaced by categorical cross-entropy (CCE)~\cite{b63}, and for regression tasks, mean squared error (MSE) is used. The feature dispersion penalty encourages smooth transitions across reordered features:
\begin{equation}
P_d(X_r) = \frac{1}{N} \sum_{i=1}^{N} \sum_{j=1}^{M-1} \left| X_r^{(i,j)} - X_r^{(i,j+1)} \right|
\end{equation}
\begin{align}
\label{ore}
P_{\text{g}}(\mathcal{O}_{\text{g}}) &= \frac{1}{|V|} \sum_{(u,v) \in V} \alpha_{u,v} \cdot d(\mathcal{O}_{\text{g}}(u), \mathcal{O}_{\text{g}}(v)) \\
\alpha_{u,v} &= \frac{1}{\|c_u - c_v\| + \epsilon}
\end{align}
Global ordering coherence is encouraged via centroid-proximity weighting (Eq.~\ref{ore}). The full DFO loss balances these components:
\begin{equation}
L_{\text{DFO}} = \lambda_{\text{d}} P_d(X_r) + \lambda_{\text{g}} P_{\text{g}}(\mathcal{O}_{\text{g}}) + (1 - \lambda_{\text{d}} - \lambda_{\text{g}}) L_{\text{b}}
\end{equation}

\textbf{Dispersion Loss Function:} A simplified version excludes the global penalty:
\begin{equation}
L_{\text{d}} = \lambda_{\text{reg}} P_d(X_r) + (1 - \lambda_{\text{reg}}) L_{\text{b}}
\end{equation}

Customized loss functions enable end-to-end training that jointly promotes ordering quality, coherence, and prediction accuracy. This reinforces dynamic feature ordering as a biologically inspired mechanism akin to neural rewiring.
%%%%
%% tablle
%%%%%%
\begin{table*}[t]
\caption{Mean accuracy ($\mu\pm\sigma$) on HDLSS (top) and HDHSS (bottom) datasets and average rank ($\mu\pm\sigma$) of the top-6 models in each regime. Rankings are based on cross-dataset performance. ``$+$'' indicates ResNet50-extracted features; ``$^{\dagger}$'' denotes 11K subsample. Model symbols: 
ProtoGate$^{\dagger}$~\cite{protogate}, 
TabulaRNN$^{\ddagger}$,~\cite{mamb}, 
LGBM$^{\mathsection}$~\cite{lgbm}, 
LSPIN$^{\|}$~\cite{spin}, 
LLSPIN$^{\#}$~\cite{spin}, 
TabNet$^{\star}$~\cite{b1}.
See supplementary material for the full results with all baselines.}
\label{tab:combined_hldss_hdhss_top6}
\centering
\scriptsize
\resizebox{\textwidth}{!}{%
\begin{tabular}{l c c c c c c}
\toprule
\multicolumn{7}{c}{\textbf{HDLSS}} \\
\midrule
Dataset & 1. DynaTab (Ours) & 2. Lasso & 3. ProtoGate$^{\dagger}$ & 4. MLP & 5. TabulaRNN$^{\ddagger}$ & 6. LGBM$^{\mathsection}$ \\
\midrule
GLI-85        & $85.96 \pm 5.77$ & $85.88 \pm 4.71$ & $82.48 \pm 5.68$ & $85.41 \pm 8.00$ & $79.68 \pm 6.68$ & $85.88 \pm 11.53$ \\
SMK\_CAN\_187 & $61.31 \pm 3.37$ & $61.19 \pm 13.72$ & $60.16 \pm 5.10$ & $59.05 \pm 7.44$ & $60.02 \pm 3.18$ & $58.85 \pm 10.14$ \\
ALLAML        & $92.31 \pm 5.77$ & $87.24 \pm 3.39$ & $86.12 \pm 3.34$ & $89.98 \pm 9.17$ & $88.92 \pm 2.02$ & $85.81 \pm 5.67$ \\
Prostate-GE   & $90.91 \pm 8.91$ & $91.18 \pm 6.39$ & $90.58 \pm 5.72$ & $89.20 \pm 6.07$ & $90.50 \pm 6.00$ & $91.38 \pm 5.71$ \\
Arcene        & $83.00 \pm 6.71$ & $81.00 \pm 3.39$ & $81.50 \pm 5.10$ & $78.40 \pm 4.05$ & $81.50 \pm 5.10$ & $80.50 \pm 5.79$ \\
TOX-171       & $88.71 \pm 3.53$ & $91.86 \pm 6.03$ & $92.34 \pm 5.67$ & $94.48 \pm 4.28$ & $85.80 \pm 4.70$ & $81.98 \pm 6.25$ \\
Colon         & $85.71 \pm 8.91$ & $79.40 \pm 10.18$ & $83.95 \pm 9.82$ & $83.95 \pm 9.80$ & $84.20 \pm 6.50$ & $76.60 \pm 11.67$ \\
Lung          & $92.75 \pm 1.28$ & $94.47 \pm 4.39$ & $93.44 \pm 6.37$ & $96.47 \pm 2.69$ & $90.50 \pm 4.80$ & $93.42 \pm 5.91$ \\
\midrule
Avg.\ Rank ($\mu\pm\sigma$) & $2.63 \pm 2.26$ & $5.13 \pm 4.29$ & $5.50 \pm 3.74$ & $5.88 \pm 4.22$ & $8.00 \pm 5.86$ & $9.00 \pm 6.19$ \\
\midrule
\multicolumn{7}{c}{\textbf{HDHSS}} \\
\midrule
Dataset & 1. DynaTab (Ours) & 2. LSPIN$^{\|}$ & 3. MLP & 4. LGBM$^{\mathsection}$ & 5. TabNet$^{\star}$ & 6. LLSPIN$^{\#}$ \\
\midrule
HAM10000$^{+}$           & $84.48 \pm 0.38$ & $81.37 \pm 1.13$ & $80.05 \pm 0.95$ & $80.33 \pm 1.33$ & $73.51 \pm 3.62$ & $79.60 \pm 1.19$ \\
DeepLesion$^{+\dagger}$  & $94.48 \pm 0.25$ & $93.14 \pm 0.29$ & $92.93 \pm 4.06$ & $94.73 \pm 2.84$ & $93.02 \pm 0.20$ & $92.95 \pm 0.63$ \\
MNIST$^{+\dagger}$       & $96.68 \pm 0.16$ & $96.77 \pm 0.40$ & $96.20 \pm 0.76$ & $95.42 \pm 0.43$ & $96.88 \pm 0.38$ & $96.36 \pm 0.34$ \\
Fashion\,MNIST$^{+\dagger}$ & $88.52 \pm 0.21$ & $87.46 \pm 0.53$ & $87.85 \pm 0.65$ & $87.25 \pm 0.68$ & $87.99 \pm 0.68$ & $87.89 \pm 0.29$ \\
CIFAR-10$^{+\dagger}$                  & $88.58 \pm 0.45$ & $87.17 \pm 1.02$ & $88.10 \pm 0.62$ & $87.20 \pm 0.75$ & $87.25 \pm 0.63$ & $86.59 \pm 0.63$ \\
Dog\,vs\,Cat$^{+\dagger}$ & $99.20 \pm 0.15$ & $99.25 \pm 0.13$ & $99.22 \pm 0.15$ & $99.12 \pm 0.15$ & $99.09 \pm 0.13$ & $99.19 \pm 0.29$ \\
\midrule
Avg.\ Rank ($\mu\pm\sigma$) & $2.50 \pm 2.35$ & $4.67 \pm 2.80$ & $5.67 \pm 3.01$ & $6.33 \pm 3.67$ & $6.83 \pm 4.62$ & $7.33 \pm 3.01$ \\
\bottomrule
\end{tabular}%
} % end resizebox
\end{table*}
%%%%%%%
\section{Experimental Results}
\textbf{When Feature Ordering Matters:}
While Deep Sets~\cite{deepsets} handles unordered data 
%(e.g., point clouds, MIL, chemoinformatics), 
(e.g., point clouds,  chemoinformatics), we focus on high-dimensional tabular data where column order impacts learning. Optimal permutations reduce redundancy, reveal feature dependencies, and improve performance. This is especially relevant for high-dimensional biological profiles (e.g., gene expression), electroencephalograms (EEGs) and sensor streams, remote sensing, climate 
%records
data, and multimodal or engineered feature tables, typical domains marked by sparsity, redundancy, and latent structure.
%, all well suited to sequence-based models like DynaTab. 
%\marginpar{\textcolor{red}{Pls check last para. Split. Re-write!!}}
%\newline

\textbf{Categorization Rules:}
Let \(n\) be the number of samples, \(m\) the number of features, and \(\rho = \frac{m}{n}\) the feature-to-sample ratio. Each dataset is assigned to one of five categories based on empirical \(\rho\) ranges: \newline
\\
%\medskip
%%%%
\begin{tabular}{@{}lp{0.7\linewidth}@{}}
\textbf{HDLSS:}   & \(m>1000,\;n<1000,\;\rho>2\).\\
\textbf{HDHSS:}   & \(m>1000,\;n>10^4,\;0.005<\rho\le2\).\\
\textbf{LDHSS:}   & \(m\le100,\;n>10^4,\;\rho\le0.01\).\\
\textbf{LDLSS:}   & \(m\le100,\;n\le1000,\;\rho\le0.05\).\\
\textbf{MixedRegime:}   & otherwise.
\end{tabular}
%%%%

\textbf{Datasets:} We evaluate DynaTab on 36 datasets across five structural regimes defined by sample size and dimensionality. The HDLSS (high dimensionality, low sample size) group includes 8 biological datasets (e.g., Arcene, Colon, GLI-85, SMK\_CAN\_187, etc.) downloaded from~\cite{hdlsss}. The HDHSS (high dimensionality, high sample size) group includes 
%consists of
6 image datasets (e.g., HAM10000, MNIST, DeepLesion, etc.). We use ResNet-50 embeddings for the images. The LDLSS (low dimensionality, low sample size) group includes 6 low-dimensional datasets (e.g., Iris, Pima Indian, Glass), while the LDHSS (low dimensionality, high  sample size)  includes 5 large-sample datasets (e.g., Higgs, Adult Census, MiniBooNE). The MixedRegime category spans 8 varied datasets from clinical, geospatial, and chemical domains e.g. ADNI, MOF, EEG-FE~\cite{b73}, CNAE-9, AI-D~\cite{b68}, EEG-PD~\cite{b71}, Water. We also evaluate on 3 regression datasets: DrivFace (HDLSS), Cargo, and GHG (MixedRegime). Full dataset sources and details are provided in the supplementary. 

%\newline
%%%% baselines without citation 
%%%%
\begin{figure}[t]
\centering
\includegraphics[width=\columnwidth]{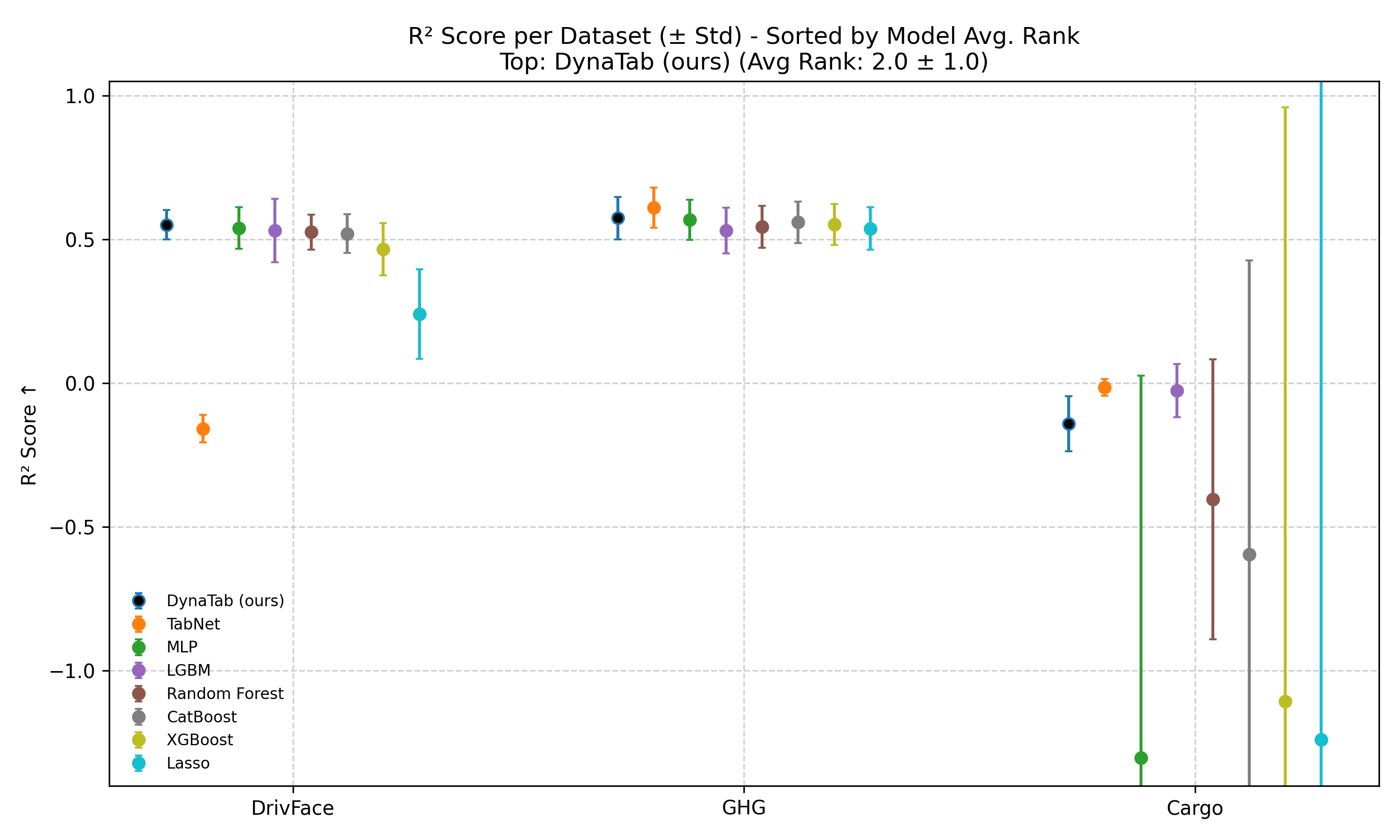}
\caption{R² score ($\pm$ std) across three regression datasets, with models ranked by their avg. performance. Our proposed DynaTab achieves the best avg. rank across datasets.}
\label{fig:r2_dotplot}
\end{figure}
%%%%%

%\marginpar{\textcolor{red}{TabNet, TabTransformer are also tabular models too!!}}
\textbf{Baseline Models:} We benchmark DynaTab against a diverse set of models: classical (Naive Bayes, KNN, SVM, Decision Tree, Lasso, RF), ensembles (AdaBoost, GBM, XGBoost, LightGBM, CatBoost), deep models (MLP, CNN, DeepFM, DCN), probabilistic/tabular models (TabPFN, TabPFN v2, NODE, ENODE, NDTF, TANGOS, TabSeq, ProtoGate, TabNet, TabTransformer, FT-Transformer, SAINT, AutoInt), and others (ResNet Tabular, ModernNCA, TabR, Mamba-variants, TabM, Trompt, TabulaRNN, LSPIN, INVASE, STG). 

%\newline
%%%%% evaluation
\textbf{Evaluation Metrics:} We report 5-fold cross-validation accuracy (mean~$\pm$~std) for all models, using 5$\times$5 nested cross-validation for HDLSS and LDLSS datasets (matching the ProtoGate's protocol~\citep{protogate} for HDLSS), and standard 5-fold cross-validation for HDHSS, LDHSS, and MixedRegime datasets. To compare methods across datasets, we compute average rank~\cite{fried} over the benchmarks and select the top 10 for further analysis for each dataset categories. We report Critical Difference (CD) diagrams~\cite{cd} and perform Friedman-Nemenyi~\cite{nemen} and Wilcoxon-Holm tests~\cite{wilcoxon} (significance level $\alpha = 0.05$) to assess pairwise significance, visualized via heatmaps. 

%\newline
%%%%%
\textbf{Implementation Details:} We tuned all models with Optuna~\citep{optuna} for 150 trials per dataset following standard tabular deep learning practice of Gorishniy et al.~\citep{tabm,tabr,embed,b5} and evaluated using task-specific cross-validation. Experiments ran on PyTorch 2.4.1+cu121 with AMP across two systems: (1) Windows (Intel, 64GB RAM, RTX 2000 Ada, 16GB VRAM) and (2) TITAN cluster (x86\_64, 188GB RAM, TITAN RTX, 24GB VRAM). DDP/DP mitigated out of memory (OOM) issues in transformer models. Full configurations, sources, and DynaTab’s optimal hyperparameters are provided in the supplement.
%\newline
%%%%%

\textbf{Computational Complexity:} DynaTab performs end-to-end training on dynamically reordered features by integrating clustering, graph construction, and rewiring. Feature ordering is computed once per training loop via KMeans (\( \mathcal{O}(nkd) \)), edge evaluation (\( \mathcal{O}(mk^2) \)), and centrality-based rewiring (\( \mathcal{O}(km^2) \)). The reordered input is processed per epoch through the DOF backbone (OPE, PIGL, DMA, and a sequential processor) at cost \( \mathcal{O}(nmd) \). Space complexity is \( \mathcal{O}(nmd + m^2) \), dominated by the backbone and graph storage. Overall, DynaTab scales linearly with number of samples \(n\), and quadratically with number of features \(m\). (See Table~\ref{tab:complexity}).
It also supports optional dimensionality reduction for further improved efficiency on high-dimensional data. 

%\newline
%\vspace{4pt}
%%%% results
\begin{table}[t]
\caption{Time and space complexity of DynaTab with integrated feature ordering.}
\label{tab:complexity}
\centering
\scriptsize
\setlength{\tabcolsep}{1.5pt}
\begin{tabular}{lcc}
\toprule
\textbf{Component} & \textbf{Time Complexity} & \textbf{Space Complexity} \\
\midrule
KMeans Clustering & \( \mathcal{O}(nkd) \) & \( \mathcal{O}(nd + kd) \) \\
Edge Metric Computation & \( \mathcal{O}(mk^2) \) & \( \mathcal{O}(m^2) \) \\
Graph Rewiring (per cluster) & \( \mathcal{O}(km^2) \) & \( \mathcal{O}(m^2) \) \\
DOF Sequential Backbone & \( \mathcal{O}(nmd) \) & \( \mathcal{O}(nmd) \) \\
\midrule
\textbf{Total (DynaTab)} & \( \mathcal{O}(nkd + mk^2 + km^2 + nmd) \) & \( \mathcal{O}(nmd + m^2) \) \\
\bottomrule
\end{tabular}
\end{table}

%%%%
\textbf{Performance Summary:} As shown in Table~\ref{tab:combined_hldss_hdhss_top6} and Figure~\ref{fig:r2_dotplot}, DynaTab achieves strong performance across diverse data regimes. On eight HDLSS datasets, it ranks best or second-best in most tasks, with the lowest average rank ($2.63 \pm 2.26$). Similarly, on six HDHSS datasets, it ranks first or second on all but one task (average rank of $2.50 \pm 2.35$), showing robustness in extreme high-dimensional settings. Notably, Lasso and MLP remain highly competitive across several datasets in HDLSS and HDHSS settings, which is consistent with the findings of ProtoGate~\citep{protogate} for HDLSS that highlight the strength of these baselines relative to heavier architectures. In MixedRegime datasets, DynaTab ranks first on five of eight tasks, yielding the best average rank ($6.00 \pm 8.72$), outperforming methods like LSPIN, TabTransformer, and REAL-X. On regression datasets, it achieves the best average rank ($2.00 \pm 1.00$), ranking first on DrivFace and second on GHG, and third on Cargo. However, on LDHSS and LDLSS datasets, tree-based models such as CatBoost, XGBoost, and GBM outperform DynaTab, which ranks tenth and fourteenth respectively, reflecting its limitations in low-dimensional regimes. Full results and comparisons for all datasets against 45 baselines are provided in the supplementary material.
%\newline
%%%%

\begin{figure}[t]
  \centering
  \includegraphics[width=\columnwidth]{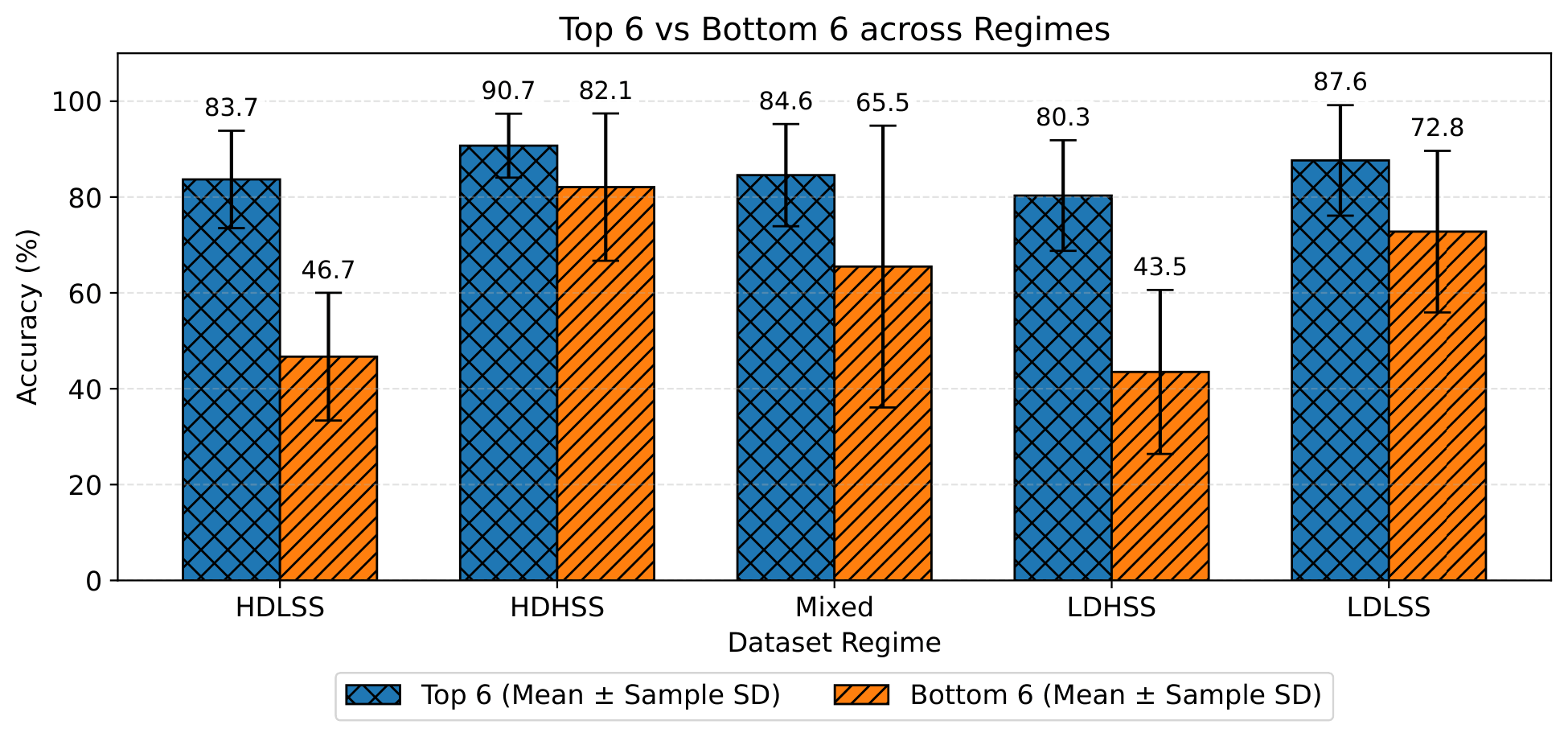}
  \caption{Top-6 vs.\ Bottom-6 mean accuracies across dataset regimes
  (HDLSS, HDHSS, Mixed, LDHSS, LDLSS). Error bars show sample standard
  deviation across the aggregated cells in each group.}
  \label{fig:top6_bottom6_regimes}
\end{figure}

\textbf{Regime-level Performance Patterns:} Figure~\ref{fig:top6_bottom6_regimes} summarizes cross-regime performance and shows that high dimensional regimes, particularly HDLSS, remain under-served in tabular deep learning. In HDLSS, the Top-6 average accuracy is 83.7\% while the Bottom-6 drops to 46.7\% (gap \(\approx 37.0\) points), the largest gap among all regimes. LDHSS exhibits a similar gap (80.3\% vs. 43.5\%, \(\Delta\approx 36.8\)), whereas HDHSS narrows to \(\approx 8.6\) points and LDLSS to \(\approx 14.9\) points. The Bottom-6 groups also show larger variability, for example sample SD \(=29.4\) in MixedRegime and \(=17.1\) in LDHSS, indicating instability and hyperparameter brittleness. These patterns are consistent with models struggling when \(m \gg n\), motivating capacity-controlled, structure-aware, and scarcity-tuned training methods. \newline
% %%%%%
%%%%
%%%%% Statistical significance
\textbf{Statistical Significance and Limitations:}  
On HDLSS datasets, DynaTab achieved the best average rank (2.12), outperforming Lasso (3.06), ProtoGate (3.50), and MLP (3.56), while TabulaRNN (4.19) and LGBM (4.56) ranked lower. A Nemenyi test with critical difference (CD = 2.40 at $\alpha = 0.05$) showed only TabulaRNN differed significantly from DynaTab (Holm-adjusted $p < 0.05$). The Friedman test ($\chi^2 = 8.5$, $p = 0.13$) was not significant, suggesting no global performance difference across models. Despite strong results on high-dimensional data, DynaTab faces memory constraints when scaling to large sample sizes (e.g., \(>100\text{K}\)), especially during ordering and Transformer-based sequential processing. To address this, we adopt Mamba for efficiency, though DP/DDP may still exacerbate OOM issues in HDHSS settings. Additionally, DynaTab underperforms on low-dimensional datasets, where classical models often suffice. See supplementary material for statistical plots and additional regime-wise analysis.
%\newline
%%%%%%

\begin{figure}[t]
    \centering
    \includegraphics[width=\columnwidth]{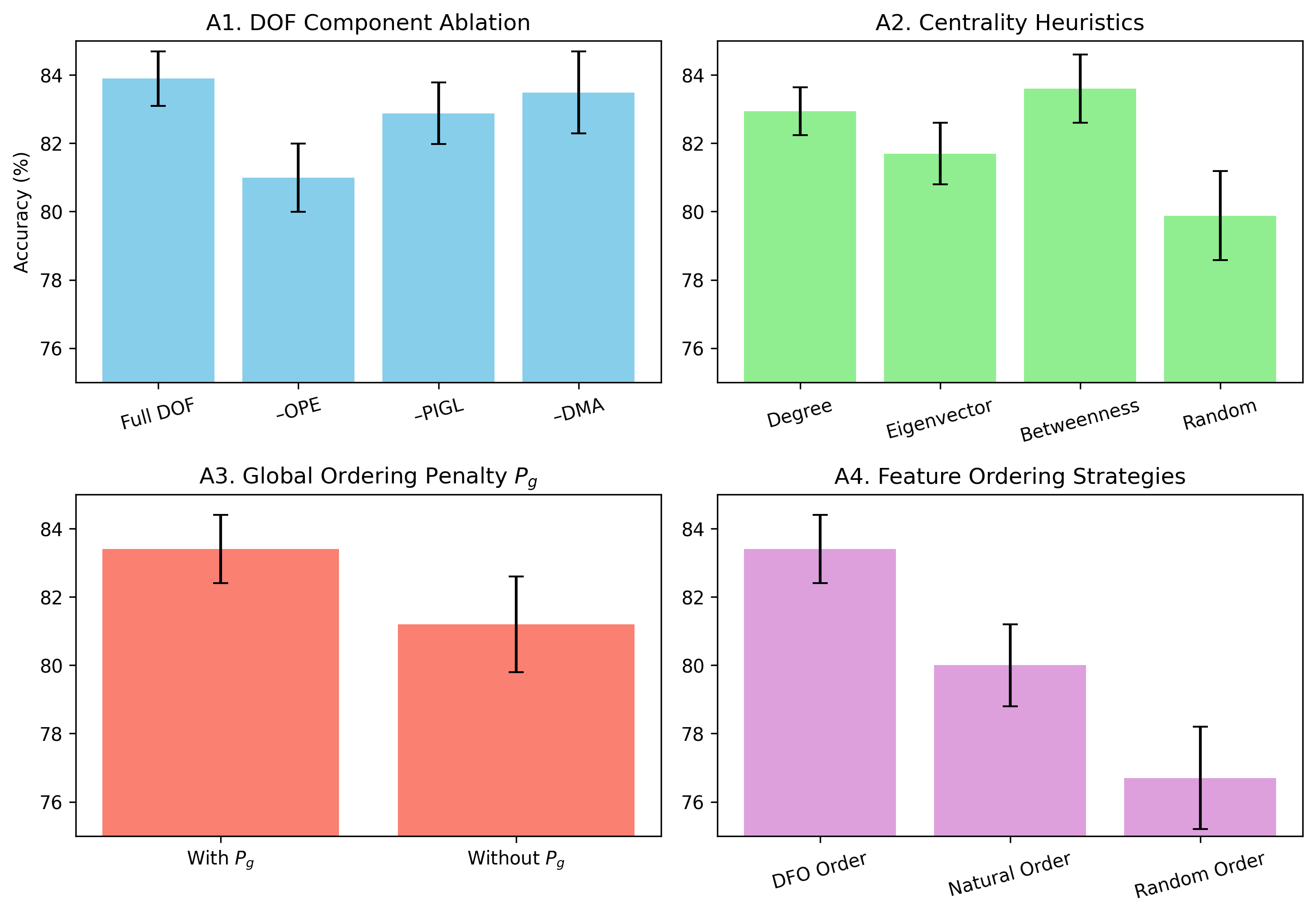}
    \caption{Key ablations on fusion, ordering, and rewiring strategies in DynaTab (See supplementary for more ablation).}
    \label{fig:ablation}
\end{figure}

\textbf{Ablation Studies:} We run targeted ablations to assess the contribution of core DynaTab components (Figure~\ref{fig:ablation}). Removing any part of the DOF fusion module (OPE, PIGL, or DMA) reduces performance (A1), noting their complementary roles. Centrality-based rewiring (A2) consistently outperforms random, validating our structural prior. Disabling the global penalty $P_g$ (A3) leads to clear accuracy drop. Finally, our learned DFO ordering (A4) surpasses natural and random feature orders, confirming the impact of ordering-aware processing. See Table~\ref{tab:dynatabseq_backbones} for backbone-specific performance, and supplementary for further ablations. \newline
\begin{table}[t]
\caption{DynaTab with different backbones as sequential processor on AI-D (case 5) dataset.}
\label{tab:dynatabseq_backbones}
\centering
\scriptsize
\setlength{\tabcolsep}{4pt}
\begin{tabular}{lcc}
\toprule
Backbone & Acc (\%) $\pm$ std & AUC $\pm$ std \\
\midrule
DAE-MHA-LSTM                       & $80.91\pm6.76$ & $0.8257\pm0.0820$ \\
Transformer                        & $83.40\pm5.10$ & $0.8470\pm0.0740$ \\
DAE                                & $81.44\pm1.04$ & $0.8445\pm0.0420$ \\
LSTM                               & $79.09\pm3.96$ & $0.8242\pm0.0589$ \\
Mamba                              & $82.18\pm2.86$ & $0.8315\pm0.0801$ \\
%MHA-DAE (Orig.\ TabSeq backbone)   & $77.78\pm1.16$ & $0.8394\pm0.0619$ \\
\bottomrule
\end{tabular}
\end{table}
%%%%
%%%%% figure
%%%%%
\vspace{-6mm}
\section{Conclusion}
We presented DynaTab, a neural rewiring-based model that dynamically orders features based on intrinsic structure and fuses them with sequence-aware backbones. A theoretical criterion using IDF and FOE guides when ordering is beneficial. DynaTab rewires local feature graphs into a global sequence, improving inductive bias and performance across 36 tabular datasets. It gains strong results on high-dimensional complex and mixed-regime tasks, with statistically significant gains (Friedman, CD, Wilcoxon-Holm). Limitations include lower gains on low-dimensional or extreme sample-size regimes. Future work includes scaling, efficient backbones, and extensions to multi-view and continual learning.
\section*{Acknowledgment}
This work was supported in part by grants from the US National Science Foundation (Award \#1920920, \#2125872, and \#2223793).
%%%%%
\bibliographystyle{plain}
\bibliography{aaai2026}
%\input{../../ReproducibilityChecklist/LaTeX/ReproducibilityChecklist.tex}
%\input{AnonymousSubmission/LaTeX/suppl}
\clearpage
%\runningtitle{DynaTab}
\newpage

% Supplementary material: To improve readability, you must use a single-column format for the supplementary material.
%\onecolumn
% No \documentclass, no \usepackage, no \maketitle here.

\onecolumn

% (Optional) If you want Appendix-style numbering like A, A.1, etc.
% \appendix  % <- try this first; many classes switch to A, B, ... automatically

% If your class doesn't letter sections the way you want, force it:

% Manual title block (since \maketitle can't be used twice)
\begin{center}
{\LARGE Supplementary Materials\\[4pt]
DynaTab: Dynamic Feature Ordering as Neural Rewiring for High-Dimensional Tabular Data\par}
\vspace{8pt}
\end{center}

This supplementary document supports our main paper \textit{DynaTab: Dynamic Feature Ordering as Neural Rewiring for High-Dimensional Tabular Data} (Proceedings, AAAI 2026 First International Workshop on Neuro for AI \& AI for Neuro: Towards Multi-Modal Natural Intelligence). Specifically, it includes:
\begin{itemize}
    \item Extended Analysis of Feature Ordering - When to Use?
    \item Extended Analysis of Datasets
    \item Baseline Details and Hyperparameters for Selected Models
    \item DynaTab Hyperparameters
    \item Discussion on HDLSS Benchmark Results
    \item Discussion on HDHSS Benchmark Results
    \item Discussion on Mixed Regime Benchmark Results
    \item Discussion on LDHSS Benchmark Results
    \item Discussion on LDLSS Benchmark Results
    \item Extended Analysis for Statistical Significance
    \item Extended Analysis on Ablation Studies
\end{itemize}
%%%%%%%%
\setcounter{section}{0}
\renewcommand{\thesection}{A}
\renewcommand{\thesubsection}{A.\arabic{subsection}}
\setcounter{figure}{0}\renewcommand{\thefigure}{A.\arabic{figure}}
\setcounter{table}{0}\renewcommand{\thetable}{A.\arabic{table}}
\section{Extended Analysis of Feature Ordering - When to Use?}
\label{app:when}
% ...
%\section{Extended Analysis of Feature Ordering - When to Use?}
%%%%% figure from main paper
\begin{figure*}[htbp]
  \centering
  \begin{subfigure}[t]{0.49\textwidth}
    \centering
    \includegraphics[width=\linewidth]{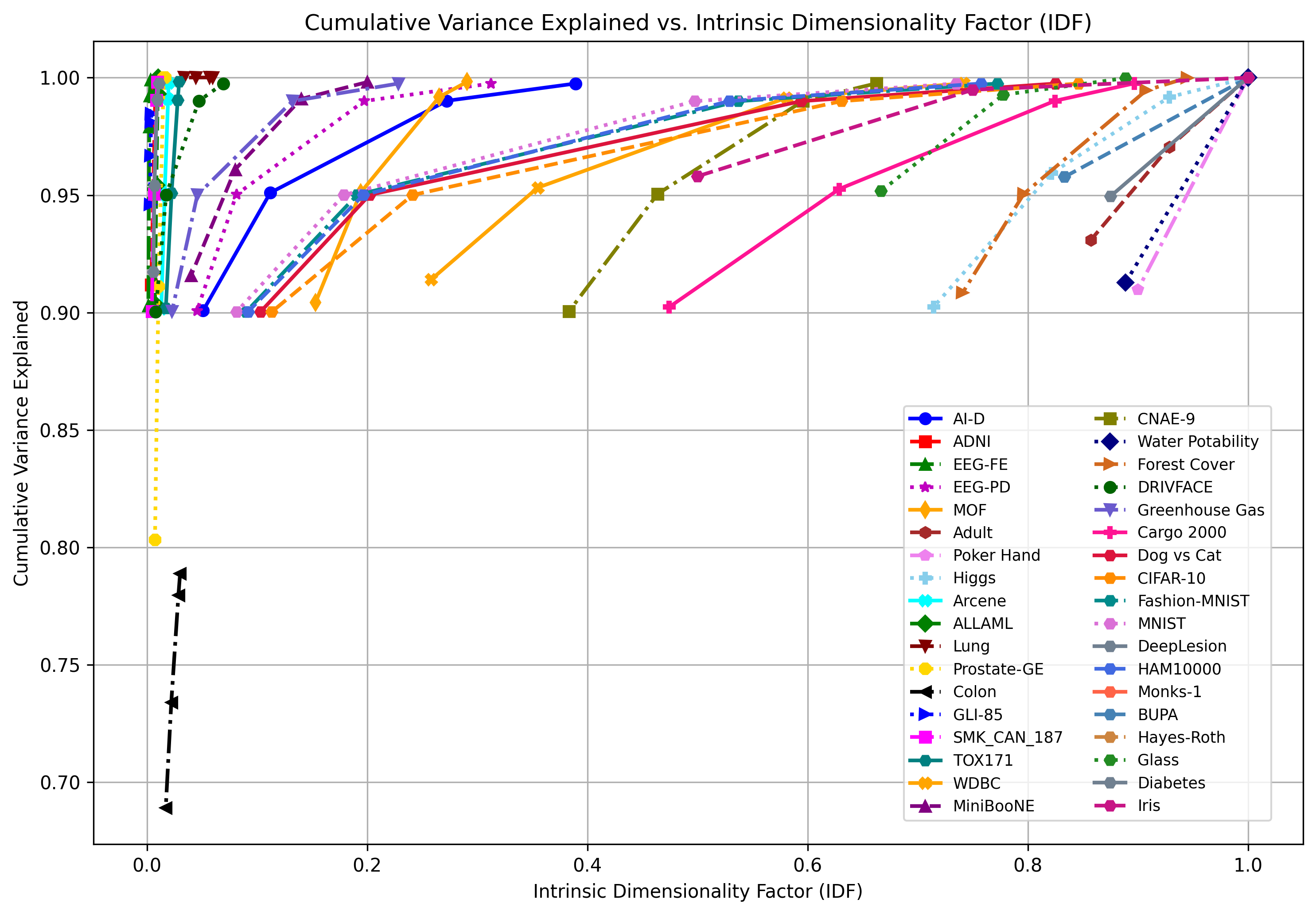}
    \caption{Cumulative Variance vs. IDF}
    \label{fig:foe1}
  \end{subfigure}\hfill
  \begin{subfigure}[t]{0.49\textwidth}
    \centering
    \includegraphics[width=\linewidth]{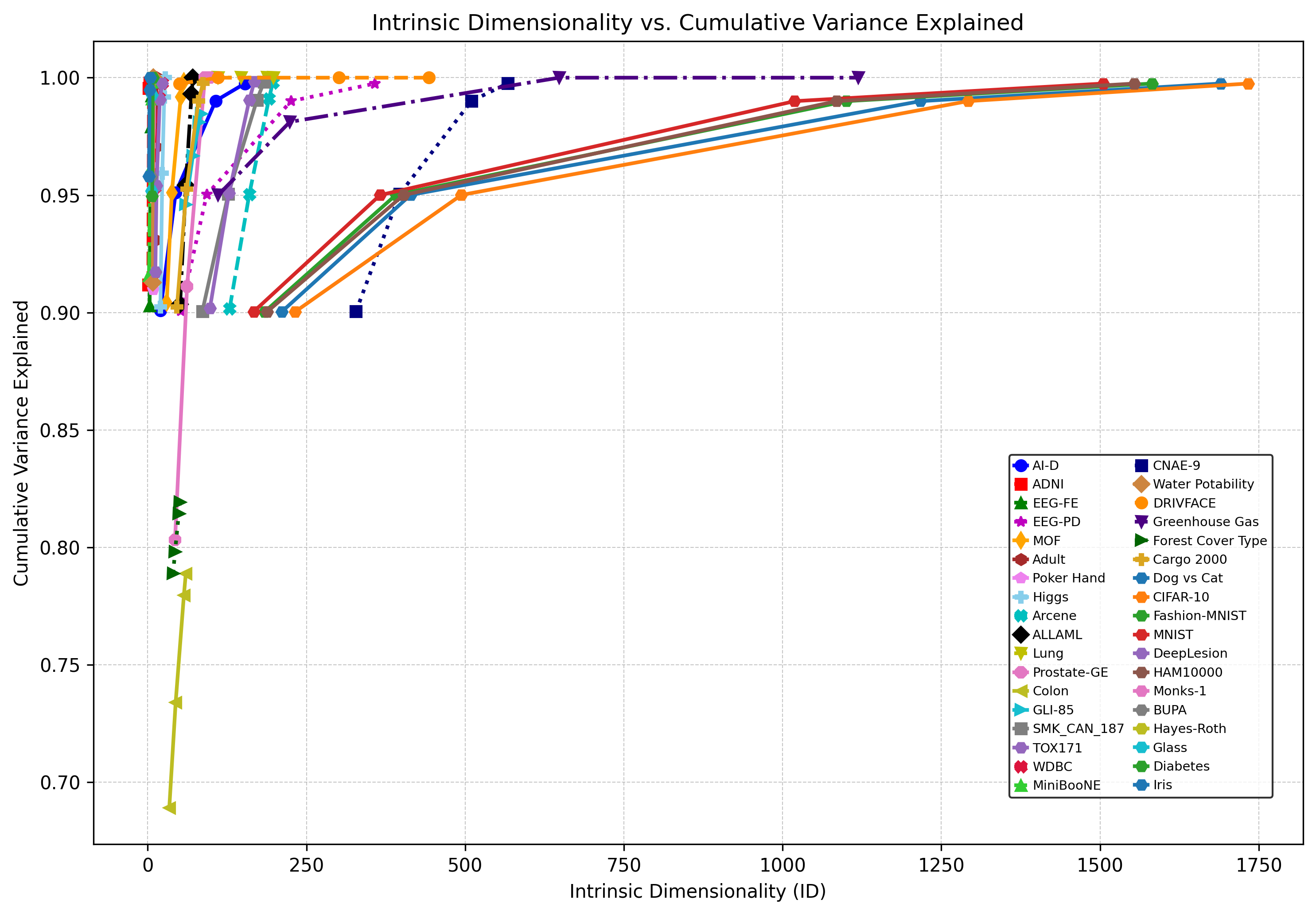}
    \caption{ID vs. Cumulative Variance}
    \label{fig:foe2}
  \end{subfigure}
  \caption{Relationship between cumulative variance and intrinsic dimensionality across all 36 datasets.}% \textcolor{red}{Let's try to use the same marker for the same dataset, for consistency, and ease of comparison !!}}
  \label{fig:foe_combined}
  \vspace{-6pt}
\end{figure*}
%%%%%%%
\label{sec:supfo}
%%%% big ranking table
\begin{table*}[t]
\caption{Ranking of datasets based on FOE score (* = optimized value). ``undefined'' means FOE was infinite and treated as bottom in ranking.}
  \label{tab:foe_singlecol_extended}
  \centering
  \scriptsize
  \renewcommand{\arraystretch}{0.95}
  \setlength{\tabcolsep}{4pt} % increase horiz spacing a bit
  \begin{tabular}{rlccccc|@{\hspace{10pt}}rlccccc}
    \toprule
    \multicolumn{7}{c}{ } & \multicolumn{7}{c}{ }\\[-6pt]
    Rank & Dataset & FOE & Complexity & $\psi^*$ & Mean IDF & AUC &
    Rank & Dataset & FOE & Complexity & $\psi^*$ & Mean IDF & AUC \\
    \midrule
     1 & EEG-FE         & 214326.59 & 651305.76 & 3.63e-06  & 0.00216 & 0.00191 &
    19 & HAM10000       & 6.46      & 107.26    & 0.41792   & 0.39352 & 0.6465  \\
     2 & GLI-85         & 90256.19  & 130481.38 & 1.08e-06  & 0.00332 & 0.00104 &
    20 & Fashion-MNIST  & 6.33      & 111.53    & 0.43986   & 0.39758 & 0.6632  \\
     3 & SMK\_CAN\_187  & 19818.83  & 48663.27  & 2.14e-05  & 0.00710 & 0.00463 &
    21 & Dog vs Cat     & 5.38      & 84.82     & 0.49220   & 0.43115 & 0.7016  \\
     4 & ADNI           & 17290.68  & 63064.61  & 5.50e-05  & 0.00760 & 0.00742 &
    22 & CIFAR-10       & 4.77      & 70.16     & 0.50455   & 0.45789 & 0.7103  \\
     5 & DeepLesion     & 14996.21  & 27611.93  & 2.65e-05  & 0.00817 & -0.00514 &
    23 & WDBC           & 4.27      & 13.72     & 0.22119   & 0.48387 & 0.4703  \\
     6 & ALLAML         & 12898.16  & 17653.29  & 7.18e-06  & 0.00880 & 0.00268 &
    24 & CNAE-9         & 3.61      & 6.13      & 0.07229   & 0.52629 & 0.2689  \\
     7 & Prostate-GE    & 6633.18   & 15459.78  & 7.32e-05  & 0.01228 & 0.00856 &
    25 & Iris           & 2.12      & 3.83      & 0.24351   & 0.68750 & 0.4935  \\
     8 & Arcene         & 3490.53   & 5418.36   & 4.19e-05  & 0.01693 & 0.00648 &
    26 & Cargo 2000     & 2.01      & 4.01      & 0.16437   & 0.70619 & 0.4054  \\
     9 & TOX171         & 1722.34   & 3102.51   & 1.35e-04  & 0.02410 & 0.01162 &
    27 & Glass          & 1.78      & 2.14      & 0.04784   & 0.75000 & 0.2187  \\
    10 & Colon          & 1683.01   & 2384.56   & 9.32e-05  & 0.02438 & 0.00965 &
    28 & Forest         & 1.39      & 1758.23   & 0.04024   & 0.84722 & 0.2006  \\
    11 & DRIVFACE       & 800.17    & 1.66      & 0.03841   & 0.03535 & 0.1960  &
    29 & Higgs          & 1.33      & 1.77      & 0.07584   & 0.86607 & 0.2754  \\
    12 & Lung           & 423.18    & 890.30    & 6.90e-04  & 0.04861 & 0.02627 &
    30 & BUPA           & 1.19      & 1.38      & 0.02662   & 0.91667 & 0.1631  \\
    13 & Greenhouse     & 87.12     & 14749.63  & 0.00358   & 0.10714 & 0.05986 &
    31 & Adult          & 1.12      & 1.27      & 0.01912   & 0.94643 & 0.1383  \\
    14 & MiniBooNE      & 75.61     & 572.35    & 0.02426   & 0.11500 & 0.1558  &
    32 & Diabetes       & 1.07      & 1.24      & 0.01485   & 0.96875 & 0.1218  \\
    15 & EEG-PD         & 39.54     & 437.50    & 0.06769   & 0.15903 & 0.2602  &
    33 & Water Potability & 1.06    & 1.16      & 0.01129   & 0.97222 & 0.1063  \\
    16 & AI-D           & 23.54     & 347.82    & 0.10789   & 0.20611 & 0.3285  &
    34 & Poker          & 1.05      & 1.12      & 0.00912   & 0.97500 & 0.0955  \\
    17 & MOF            & 19.62     & 38.60     & 0.01759   & 0.22577 & 0.1326  &
    35 & Monks-1        & undefined & 1.00      & 1.0       & 1.00000 & 0.0000  \\
    18 & MNIST          & 7.18      & 135.41    & 0.40324   & 0.37317 & 0.6350  &
    36 & Hayes-Roth     & undefined & 1.00      & 1.0       & 1.00000 & 0.0000  \\
    \bottomrule
  \end{tabular}
\end{table*}
%\section*{Extended Analysis of Feature Ordering: When and Why It Helps}
This section synthesizes the empirical evidence from the full suite of datasets to answer the core question: under what conditions does feature ordering meaningfully 
%succeed?
improve performance?
We leverage four complementary perspectives: (1) intrinsic dimensionality required to retain varying cumulative variance (Table \ref{tab:intrinsic_dimensionality_all}), (2) success probability as a function of intrinsic dimensionality factor (IDF) (Figure \ref{fig:feature_graph} and the transposed view in Figure \ref{fig:foe_combined}), (3) the compounded effectiveness summarized by the FOE-based ranking (Table \ref{tab:foe_singlecol_extended}), and (4) detailed head-to-head contrasts for illustrative pairs (Figure \ref{fig:feature_graph4}). Together, they reveal nuanced structure about data compactness, complexity, and amenability to feature ordering.
%%%%%%
\subsection{Intrinsic Dimensionality and Compactness}
Table \ref{tab:intrinsic_dimensionality_all} enumerates the number of principal components needed to achieve four cumulative variance thresholds (99.75\%, 99\%, 95\%, 90\%) for each dataset. Datasets such as ADNI and EEG-FE stand out immediately: even at the stringent 99.75\% level, their intrinsic dimensionality is tiny (3 and 8 components, respectively), indicating extremely compact representations. This is visually corroborated in Figure \ref{fig:feature_graph2} (Cumulative Variance vs Number of Components) and in the alternate projection in Figure \ref{fig:foe2}, where their curves spike to near-total variance with minimal components. Such sharp concentration suggests that the variance is dominated by a few coherent directions, making them favorable candidates for interventions that rely on identifying and preserving key directions, such as feature ordering. In contrast, datasets such as Dog vs. Cat, CIFAR-10, Fashion-MNIST, and HAM10000 require hundreds to over a thousand components to reach high-variance thresholds, reflecting diffuse information spread across many directions. Their cumulative variance curves rise much more gradually (Figure \ref{fig:feature_graph2}), signifying that there is no small subspace capturing the bulk of the variability. This inherently limits how concisely their structure can be exploited, making effective feature ordering more challenging unless the ordering can reliably isolate the few most informative features among many. GLI-85, updated in the success probability table to have very high success probabilities (e.g., 0.9962 at 99.75\%), also shows moderately low intrinsic dimensionality (84 components for 99.75\%), situating it in a middle regime: not as concentrated as ADNI/EEG-FE, but much more so than the image and high-variability datasets. Its performance will be better understood in later subsections when juxtaposing intrinsic compactness with ordering effectiveness.
%%%%%%
%%%%% hyper parameters
\begin{table*}[t]
\centering

\begin{minipage}[t]{0.49\textwidth}
\caption{Optuna-tuned parameters for DynaTab on the AI-D (Case 5) dataset.}
\label{tab:optuna_aid}
\centering
\scriptsize
\setlength{\tabcolsep}{4pt}
\begin{tabular}{ll|ll}
\toprule
Parameter & Value & Parameter & Value \\
\midrule
Num Clusters       & 12            & Backbone           & Transformer \\
Order              & Descending    & Mutation Prob      & 0.2 \\
Tolerance          & 0.021         & Loss Mode          & DFO \\
$\lambda_{\text{disp}}$   & 0.4     & $\lambda_{\text{global}}$ & 0.3 \\
Learning Rate      & 0.001         & d\_model           & 256 \\
nhead              & 4             & Num Layers         & 3 \\
Window Size        & 64            & Metric             & euclidean \\
\bottomrule
\end{tabular}
\end{minipage}\hfill
\begin{minipage}[t]{0.49\textwidth}
\caption{Optuna-tuned parameters for DynaTab on the Pima Indian dataset.}
\label{tab:optuna_pima}
\centering
\scriptsize
\setlength{\tabcolsep}{4pt}
\begin{tabular}{ll|ll}
\toprule
Parameter & Value & Parameter & Value \\
\midrule
Num Clusters       & 2            & Backbone           & Transformer \\
Order              & Ascending    & Mutation Prob      & 0.0 \\
Tolerance          & 0.001        & Loss Mode          & DFO \\
$\lambda_{\text{disp}}$   & 0.0     & $\lambda_{\text{global}}$ & 0.0 \\
Learning Rate      & 0.001        & d\_model           & 128 \\
nhead              & 8            & Num Layers         & 2 \\
Window Size        & 32           & Metric             & manhattan \\
\bottomrule
\end{tabular}
\end{minipage}

\end{table*}
%%%%%%%
\begin{table*}[t]
\centering

\begin{minipage}[t]{0.32\textwidth}
\captionof{table}{Optuna-tuned parameters for DynaTab on the Arcene dataset.}
\label{tab:optuna_arcene}
\centering
\tiny
\setlength{\tabcolsep}{3pt}
\begin{tabular}{ll|ll}
\toprule
Parameter & Value & Parameter & Value \\
\midrule
Num Clusters       & 20           & Backbone           & Mamba \\
Order              & Descending   & Mutation Prob      & 0.2 \\
Tolerance          & 0.021        & Loss Mode          & DFO \\
$\lambda_{\text{disp}}$ & 0.1    & $\lambda_{\text{global}}$ & 0.1 \\
Learning Rate      & 0.0001       & d\_model           & 128 \\
Num Layers         & 3            & Dropout            & 0.4 \\
Metric             & manhattan    &                    &      \\
\bottomrule
\end{tabular}
\end{minipage}\hfill
\begin{minipage}[t]{0.32\textwidth}
\captionof{table}{Optuna-tuned parameters for DynaTab on the MiniBooNE dataset.}
\label{tab:optuna_miniboone}
\centering
\tiny
\setlength{\tabcolsep}{3pt}
\begin{tabular}{ll|ll}
\toprule
Parameter & Value & Parameter & Value \\
\midrule
Num Clusters       & 5            & Backbone           & Transformer \\
Order              & Descending   & Mutation Prob      & 0.2 \\
Tolerance          & 0.021        & Loss Mode          & DFO \\
$\lambda_{\text{disp}}$ & 0.1     & $\lambda_{\text{global}}$ & 0.1 \\
Learning Rate      & 0.0001       & d\_model           & 128 \\
nhead              & 8            & Num Layers         & 2 \\
Window Size        & 32           & Metric             & manhattan \\
\bottomrule
\end{tabular}
\end{minipage}\hfill
\begin{minipage}[t]{0.32\textwidth}
\captionof{table}{Optuna-tuned parameters for DynaTab on the HAM10000 dataset.}
\label{tab:optuna_ham10000}
\centering
\tiny
\setlength{\tabcolsep}{3pt}
\begin{tabular}{ll|ll}
\toprule
Parameter & Value & Parameter & Value \\
\midrule
Num Clusters       & 20           & Backbone           & Mamba \\
Order              & Descending   & Mutation Prob      & 0.2 \\
Tolerance          & 0.021        & Loss Mode          & DFO \\
$\lambda_{\text{disp}}$ & 0.1     & $\lambda_{\text{global}}$ & 0.1 \\
Learning Rate      & 5e-5         & d\_model           & 128 \\
Num Layers         & 3            & Dropout            & 0.4 \\
Batch Size         & 64           & Accum. Steps       & 2 \\
Clip Norm          & 1.0          & Metric             & manhattan \\
\bottomrule
\end{tabular}
\end{minipage}

\end{table*}
%%%%%%%
\subsection{Intrinsic Dimensionality Factor (IDF) and Success Probability}
The Intrinsic Dimensionality Factor (IDF), defined as the ratio of intrinsic dimensionality to the true (ambient) dimensionality, normalizes these counts and allows cross-dataset comparison. Figure \ref{fig:feature_graph} shows success probability of feature ordering as a function of IDF for each dataset; Figure \ref{fig:foe_combined} offers a transposed, multi-view layout that emphasizes per-dataset trends across the same axis space. A consistent and informative pattern emerges: datasets with very low IDF (i.e., compact intrinsic representations like ADNI, EEG-FE, GLI-85, ALLAML, and DeepLesion) maintain high success probability across a wide range of variance thresholds. Their curves in Figure \ref{fig:feature_graph} are relatively flat and positioned near the top, indicating that even as more variance is required (which typically increases intrinsic dimensionality and hence IDF), the degradation in success probability is marginal. This suggests that when variance is concentrated in a few directions, feature ordering can reliably prioritize those directions early, yielding high success even if one expands the retained subspace slightly. By contrast, datasets with large IDF such as CIFAR-10, Dog vs Cat, Fashion-MNIST, and Adult exhibit steep declines in success probability as IDF increases. Their curves in Figure~\ref{fig:foe_combined} exhibit a clear decreasing trend with IDF (see Fig.~\ref{fig:foe_combined}(b)): as higher intrinsic dimensionality (thus larger IDF) is needed to explain the same variance, the success probability of feature ordering drops rapidly. This indicates that when variance is dispersed across many directions, so that no small subset dominates, ordering becomes a weaker proxy for true feature importance, and early selected features are more likely to miss critical structure, leading to lower success. Some datasets occupy intermediate regimes (e.g., MOF, EEG-PD, WDBC): their curves show moderate slopes and partial retention of ordering benefit. Their performance depends on the variance threshold at looser thresholds (e.g., 90\%), ordering regains more utility because the effective IDF is smaller, but at tighter ones, ordering becomes less reliable. %\marginpar{\textcolor{red}{Pls Chk "negative slope"!!}}
%%%%%%%
\subsection{Success Probability at Intrinsic Dimensionality and the Effect of Compactness}
Table \ref{tab:success_probability_full} (with the corrected GLI-85 row) reports success probability evaluated exactly at the intrinsic dimensionality for each variance level. This isolates the performance when the ordering is tasked to capture the minimal subspace necessary for the desired variance, making it a clean reflection of whether the ordering aligns with the true dominant directions. Datasets like EEG-FE, ALLAML, SMK\_CAN\_187, Prostate-GE, and DeepLesion achieve extremely high success probabilities (often exceeding 0.99) even at the most stringent 99.75\% threshold, reinforcing that their intrinsic structure is both compact and ordered in a way that the selected features are strongly informative. GLI-85’s high success across all variants (0.9962 to 0.9973) further solidifies it as a dataset where feature ordering captures the key dimensions almost perfectly. On the lower end, datasets such as Monks-1, Hayes-Roth, and some of the small tabular datasets (e.g., BUPA, Iris at certain thresholds) have either zero or very low success probabilities. Their extreme values (e.g., zero for Monks-1 and Hayes-Roth at most thresholds) point to pathological regimes where either variance is too uniformly spread relative to dimension (making ordering noisy), or the notion of ordering does not meaningfully separate signal from noise given the dataset structure. For example, Monks-1 has intrinsic dimensionality equal to its ambient dimension (IDF=1) across thresholds, leaving virtually no ordering leverage. Intermediate datasets like AI-D, EEG-PD, and MOF show moderate success suggesting that while there is some alignment between ordering and informative directions, it is not sufficiently sharp to be a panacea, especially at higher variance retention (e.g., 99.75\%).
%%%%%%
%1st figure
\begin{figure*}[t]
\centering
\begin{subfigure}[t]{0.5\textwidth}
  \centering
  \includegraphics[width=\linewidth]{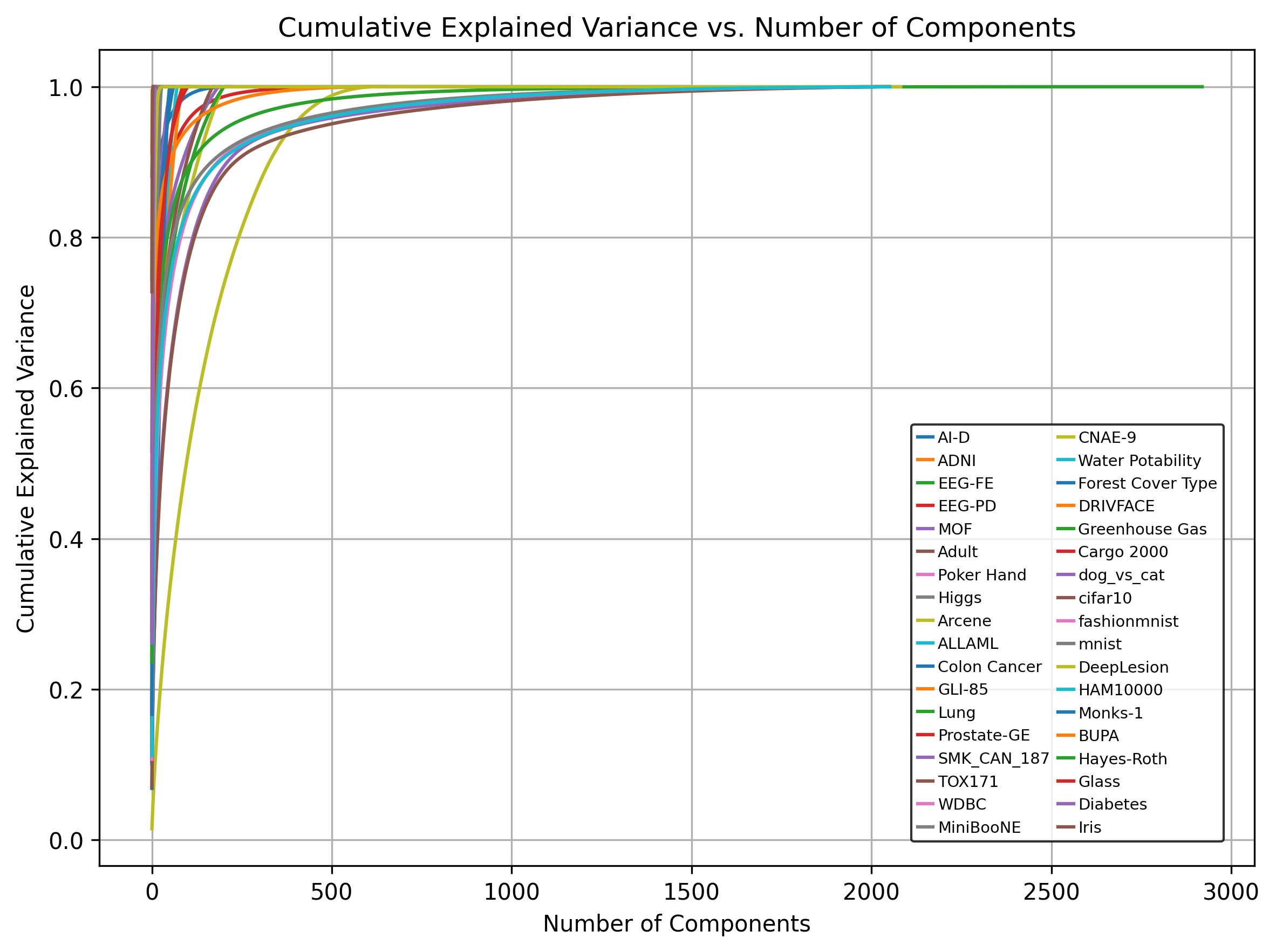}
  \caption{}
  \label{fig:feature_graph2}
\end{subfigure}%
\begin{subfigure}[t]{0.5\textwidth}
  \centering
  \includegraphics[width=\linewidth]{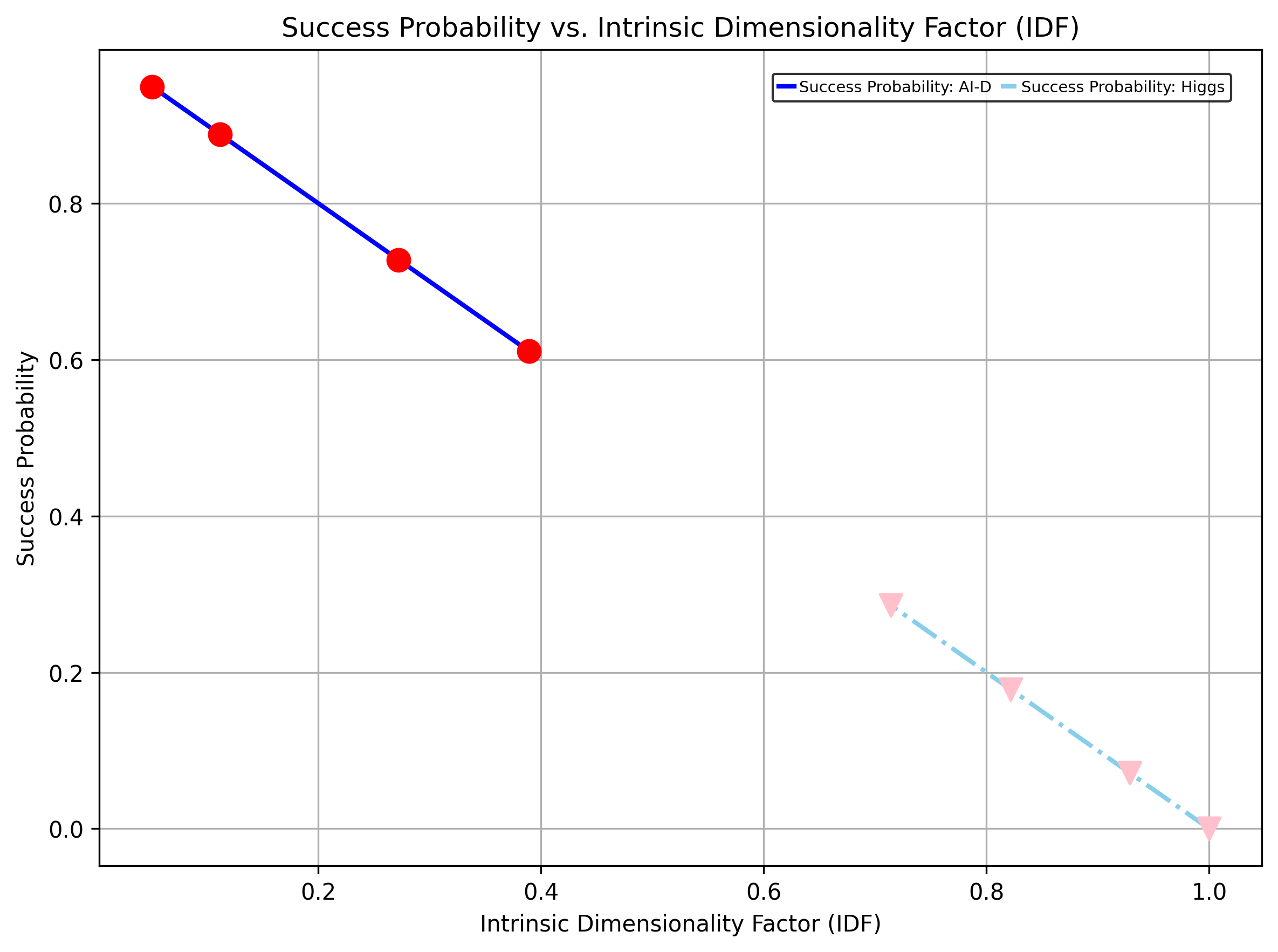}
  \caption{}
  \label{fig:feature_graph4}
\end{subfigure}
\vspace{-2mm}
\caption{Additional analyses: (a) cumulative variance vs.\ number of components across datasets; (b) head-to-head success probability of feature ordering vs.\ IDF for AI-D and Higgs.}
\label{fig:feature_graphs_combined}
\end{figure*}
%%%%%%%
\subsection{Feature Ordering Effectiveness (FOE) and Dataset Complexity}
Table \ref{tab:foe_singlecol_extended} aggregates the above findings into a composite ranking based on the FOE score, along with auxiliary metrics: complexity, optimal $\psi^*$ (ordering parameter), mean IDF, and AUC. The FOE score encapsulates the net benefit of ordering relative to complexity and variance retention behavior. Top-ranked datasets (EEG-FE, GLI-85, SMK\_CAN\_187, ADNI, DeepLesion, ALLAML) share the empirical fingerprints identified earlier: low mean IDF, high compactness, and correspondingly high success probabilities. Their high complexity values despite compactness reveal a subtle point: complexity here is not merely about dimensional spread but about nuanced structure (e.g., subtle patterns, correlations, biological signal) that makes feature ordering valuable. High FOE alongside nontrivial complexity suggests ordering both uncovers and leverages structured signal that might otherwise be obscured in naive representations. Mid-tier datasets (e.g., Colon, TOX171, DRIVFACE, Lung) exhibit a mix of moderate mean IDF and non-negligible complexity. Their FOE scores, while lower, are still positive, indicating that ordering provides benefit, albeit less spectacular and more sensitive to the choice of parameters (reflected in $\psi^*$ variability). Lower-ranked or “undefined” datasets (such as Monks-1 and Hayes-Roth) expose the limits of the approach. Their FOE is infinite/undefined because the ordering fails to improve over the trivial baseline in any meaningful way, and their mean IDF is extreme (close to 1), implying that signal is not concentrated in a way that ordering can exploit. In combination with low effective complexity (relative to how FOE is computed), this yields minimal (or no) gain, underlining a boundary regime where feature ordering should either be deprioritized or rethought.
\subsection{Pairwise and Head-to-Head Illustrations}
Figure \ref{fig:feature_graph4} provides a focused head-to-head comparison, e.g., between AI-D and Higgs. The contrast is instructive: AI-D, despite having a moderate IDF, shows a steeper decline in success probability as the IDF increases, whereas in the extreme case of Higgs (which has very low compactness IDF near 1 for high thresholds), the challenge is that ordering has very little room to improve, and success probabilities collapse accordingly. Such pairwise views help clarify that two datasets with similar ambient sizes may behave drastically differently depending on intrinsic structure; thus, global heuristics (like “larger dimension means harder”) are insufficient without the nuanced lens of IDF and ordering alignment.
%%%%%%%
\subsection{Synthesis and Practical Implications}
Bringing the threads together, the following practical insights emerge:
%%%%%%%
\begin{enumerate}
  \item \textbf{High Benefit Regime:} Datasets with low mean IDF and high FOE (e.g., EEG-FE, ADNI, GLI-85, ALLAML, DeepLesion) are ideal candidates for feature ordering. They combine compact variance concentration with structured complexity, meaning ordering can rapidly isolate informative features and yield large relative gains.
  \item \textbf{Conditional Benefit Regime:} Datasets with moderate IDF and FOE (e.g., Colon, TOX171, DRIVFACE, Lung) benefit, but performance depends on tuning (optimal $\psi^*$) and variance target. For looser variance thresholds (lower intrinsic dimensionality), ordering is more effective; for tighter ones, the window narrows.
  \item \textbf{Low/No Benefit Regime:} Datasets exhibiting high IDF (close to 1), low FOE, or pathological characteristics (e.g., Monks-1, Hayes-Roth, and small tabular sets whose success probability remains nearly constant and low across variance thresholds) provide little to no benefit from feature ordering in the current formulation. %\marginpar{\textcolor{red}{"flat success"?? }}
  \item \textbf{Interpretability of Ordering:} The consistent alignment between compact intrinsic structure and success probability indicates that feature ordering is implicitly discovering principal directions (or strong proxies thereof) when those directions dominate variance. Failure cases often arise when variance is diffuse, meaning ordering needs auxiliary signals (e.g., supervised guidance or domain priors) to differentiate among many weakly informative features.
  \item \textbf{Role of Complexity:} High dataset complexity does not preclude success; rather, it often coexists with high ordering benefit when that complexity is organized (structured) rather than random. Thus, complexity metrics should be interpreted jointly with IDF and FOE to decide whether effort on feature ordering is warranted.
\end{enumerate}
%%%%%%
\subsection{Recommendations for Usage}
\begin{itemize}
  \item \textbf{Pre-screening:} Compute intrinsic dimensionality and IDF at the desired variance level. If IDF is low (e.g., $< 0.1 $) and success probability at that point is empirically high (or inferred from similar datasets), prioritize feature ordering in pipelines.
  \item \textbf{Adaptive Thresholding:} For datasets with moderate IDF, consider relaxing variance retention targets to reduce IDF and improve ordering reliability trading off a bit of variance for more stable ordering gains.
  \item \textbf{Fallback Strategies:} For datasets in the low-benefit regime, either avoid feature ordering or augment it with domain-specific feature importance signals, since naive ordering is unlikely to yield robust improvements.
  \item \textbf{Hyperparameter Tuning:} Use the $\psi^*$ (optimal ordering parameter) profile from Table \ref{tab:foe_singlecol_extended} as starting points when adapting to new but similar datasets; small deviations in structure can cause the effective ordering scale to shift.
\end{itemize}
%%%%%%%
\subsection{Caveats and Future Directions}
\begin{figure*}[t]
    \centering
    \includegraphics[width=\textwidth]{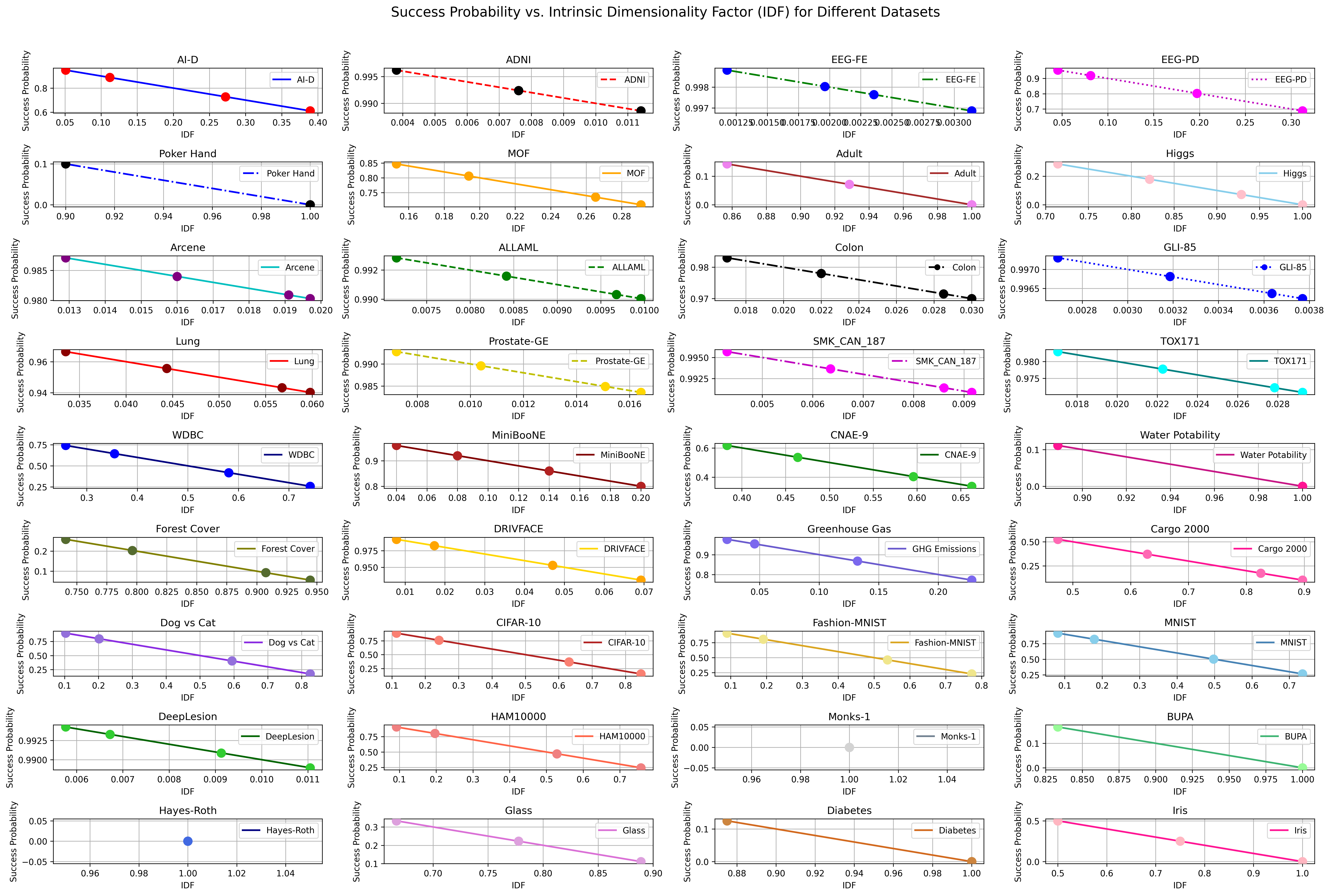} % Replace with your image path
    \caption{Success probability of Feature Ordering vs IDF for Different Datasets.}
    \label{fig:feature_graph}
    \vspace{-1em} % Adjusts space after figure to comply with page limits
\end{figure*}

While the current analysis unifies variance concentration, ordering alignment, and complexity, it remains largely empirical. Future work could formalize the theoretical link between ordering heuristics and the spectral properties of the data covariance in regimes of diffuse variance. Additionally, developing hybrid orderings that incorporate supervised signals or nonlinear embeddings may recover utility for datasets currently in the low-benefit regime.
\\
\begin{table*}[t]
\caption{Intrinsic dimensionality (number of components) required to reach various cumulative variance thresholds for all datasets.}
\label{tab:intrinsic_dimensionality_all}
\centering
\renewcommand{\arraystretch}{0.9}
\setlength{\tabcolsep}{3pt}
\begin{tabular}{lcccc|@{\hspace{10pt}}lcccc}
\toprule
\textbf{Dataset} & \textbf{99.75\%} & \textbf{99\%} & \textbf{95\%} & \textbf{90\%} &
\textbf{Dataset} & \textbf{99.75\%} & \textbf{99\%} & \textbf{95\%} & \textbf{90\%} \\
\midrule
AI-D               & 153  & 107  & 44  & 20  & TOX171             & 168  & 160  & 128 & 98  \\
ADNI               & 3    & 2    & 2   & 1   & DRIVFACE           & 443  & 301  & 111 & 50  \\
EEG-FE             & 8    & 6    & 5   & 3   & Dog vs Cat         & 1690 & 1217 & 414 & 211 \\
EEG-PD             & 358  & 226  & 93  & 52  & CIFAR-10           & 1733 & 1292 & 494 & 232 \\
MOF                & 57   & 52   & 38  & 30  & Fashion-MNIST      & 1582 & 1100 & 391 & 184 \\
WDBC               & 23   & 18   & 11  & 8   & MNIST              & 1505 & 1019 & 366 & 167 \\
CNAE-9             & 567  & 510  & 397 & 328 & DeepLesion         & 23   & 19   & 14  & 12  \\
Water Potability   & 9    & 9    & 9   & 8   & HAM10000           & 1554 & 1085 & 403 & 188 \\
Forest Cover Type  & 51   & 49   & 43  & 40  & Monks-1            & 6    & 6    & 6   & 6   \\
Greenhouse Gas     & 1119 & 648  & 224 & 111 & BUPA               & 6    & 6    & 5   & 5   \\
Cargo 2000         & 87   & 80   & 61  & 46  & Hayes-Roth         & 5    & 5    & 5   & 5   \\
Arcene             & 197  & 191  & 160 & 129 & Glass              & 8    & 7    & 6   & 6   \\
ALLAML             & 71   & 69   & 60  & 51  & Diabetes           & 8    & 8    & 8   & 7   \\
Colon Cancer       & 60   & 57   & 44  & 34  & Iris               & 4    & 3    & 2   & 2   \\
GLI-85             & 84   & 81   & 71  & 60  & Water Potability   & 9    & 9    & 9   & 8   \\
Lung               & 198  & 188  & 147 & 111 & Forest Cover Type  & 51   & 49   & 43  & 40  \\
Prostate-GE        & 98   & 90   & 62  & 43  & Greenhouse Gas     & 1119 & 648  & 224 & 111 \\
SMK\_CAN\_187      & 183  & 172  & 127 & 86  & Cargo 2000         & 87   & 80   & 61  & 46  \\
\bottomrule
\end{tabular}
\end{table*}
%%%%%%
We showcase the success probability of feature ordering methods for each dataset at different variance levels. Datasets such as EEG-FE and ADNI, which have lower intrinsic dimensionality and fewer required components (as seen in Figure \ref{fig:feature_graph}), achieve higher success probabilities consistently across all variance levels. In contrast, datasets like Poker Hand and Adult show poor success probabilities, particularly at higher variance levels, indicating limited benefits from feature ordering for these datasets. Figure \ref{fig:foe_combined}  presents the relationship between success probability and the Intrinsic Dimensionality Factor (IDF) for different datasets. Each subplot corresponds to a specific dataset, highlighting the decline in success probability with increasing IDF. Notably, datasets with compact representations, such as ADNI and EEG-FE, exhibit a shallow decline, maintaining high success probabilities even at larger IDFs. Conversely, datasets like Poker Hand and Adult demonstrate a steep decline, signifying their lower amenability to feature ordering. We provide a comprehensive ranking of datasets based on Feature Ordering Effectiveness (FOE), alongside their parameter namely, Optimal $\psi$, Mean IDF, AUC, and overall dataset complexity score. The complexity score measures the inherent complexity of each dataset, offering insights into the challenge solved by feature ordering. Datasets such as EEG-FE and ADNI demonstrate both high FOE and high complexity scores, indicating that while they are amenable to feature ordering, their specialized and focused nature presents unique challenges. Conversely, datasets like Poker Hand and Adult exhibit low complexity scores and FOE values, reflecting their limited %possible of 
potential for improvement with feature ordering.
%%%%%%%%
%2nd table
\begin{table*}[t]
\caption{Success probability at various variance levels for each dataset, evaluated at the corresponding intrinsic dimensionality.}
\label{tab:success_probability_full}
\centering
\footnotesize
\renewcommand{\arraystretch}{0.9}
\setlength{\tabcolsep}{3pt}
\begin{tabular}{lcccc@{\hspace{10pt}}lcccc}
\toprule
\textbf{Dataset} & \textbf{99.75\%} & \textbf{99\%} & \textbf{95\%} & \textbf{90\%} &
\textbf{Dataset} & \textbf{99.75\%} & \textbf{99\%} & \textbf{95\%} & \textbf{90\%} \\
\midrule
AI-D               & 0.6107 & 0.7277 & 0.8880 & 0.9491 &
TOX171             & 0.9708 & 0.9722 & 0.9777 & 0.9830 \\
ADNI               & 0.9886 & 0.9924 & 0.9924 & 0.9962 &
DRIVFACE           & 0.9308 & 0.9530 & 0.9827 & 0.9922 \\
EEG-FE             & 0.9969 & 0.9976 & 0.9980 & 0.9988 &
Dog vs Cat         & 0.1748 & 0.4058 & 0.7979 & 0.8970 \\
EEG-PD             & 0.6876 & 0.8028 & 0.9188 & 0.9546 &
CIFAR-10           & 0.1538 & 0.3691 & 0.7588 & 0.8867 \\
MOF                & 0.7092 & 0.7347 & 0.8061 & 0.8469 &
Fashion-MNIST      & 0.2275 & 0.4629 & 0.8091 & 0.9102 \\
WDBC               & 0.2581 & 0.4194 & 0.6452 & 0.7419 &
MNIST              & 0.2651 & 0.5024 & 0.8213 & 0.9185 \\
CNAE-9             & 0.3376 & 0.4042 & 0.5362 & 0.6168 &
DeepLesion         & 0.9890 & 0.9909 & 0.9933 & 0.9942 \\
Water Potability   & 0.0000 & 0.0000 & 0.0000 & 0.1111 &
HAM10000           & 0.2427 & 0.4712 & 0.8036 & 0.9084 \\
Forest Cover Type  & 0.0556 & 0.0926 & 0.2037 & 0.2593 &
MiniBooNE          & 0.8000 & 0.8600 & 0.9200 & 0.9600 \\
Greenhouse Gas     & 0.7719 & 0.8679 & 0.9543 & 0.9774 &
Adult              & 0.0000 & 0.0000 & 0.0714 & 0.1429 \\
Cargo 2000         & 0.1031 & 0.1753 & 0.3711 & 0.5258 &
Poker Hand         & 0.0000 & 0.0000 & 0.0000 & 0.1000 \\
Arcene             & 0.9803 & 0.9809 & 0.9840 & 0.9871 &
Higgs              & 0.0000 & 0.0714 & 0.1786 & 0.2857 \\
ALLAML             & 0.9900 & 0.9903 & 0.9916 & 0.9928 &
Monks-1            & 0.0000 & 0.0000 & 0.0000 & 0.0000 \\
Colon Cancer       & 0.9700 & 0.9715 & 0.9780 & 0.9830 &
BUPA               & 0.0000 & 0.0000 & 0.1667 & 0.1667 \\
GLI-85             & 0.9962 & 0.9964 & 0.9968 & 0.9973 &
Hayes-Roth         & 0.0000 & 0.0000 & 0.0000 & 0.0000 \\
Lung               & 0.9402 & 0.9432 & 0.9556 & 0.9665 &
Glass              & 0.1111 & 0.2222 & 0.3333 & 0.3333 \\
Prostate-GE        & 0.9836 & 0.9849 & 0.9896 & 0.9928 &
Diabetes           & 0.0000 & 0.0000 & 0.0000 & 0.1250 \\
SMK\_CAN\_187      & 0.9908 & 0.9914 & 0.9936 & 0.9957 &
Iris               & 0.0000 & 0.2500 & 0.5000 & 0.5000 \\
\bottomrule
\end{tabular}
\end{table*}
%%%%%%%
%%%%%%
%%%%%% feature ordering why extension end
\renewcommand{\thesection}{B}
\renewcommand{\thesubsection}{B.\arabic{subsection}}
\setcounter{figure}{0}\renewcommand{\thefigure}{B.\arabic{figure}}
\setcounter{table}{0}\renewcommand{\thetable}{B.\arabic{table}}
\section{Extended Analysis of Datasets}
\label{app:ex}
%\section{Extended Analysis of Datasets}
%%%%%%%%
We provide detailed analysis of all datasets used in this study. The 36 datasets span various regimes: HDLSS, HDHSS, LDHSS, LDLSS, and MixedRegime. Table~\ref{tab:all_datasets_full} summarizes their properties, including sample and feature counts, class distribution, FOE score, AUC, and feature-to-sample ratio.
%%%%%%%
\subsection{MixedRegime datasets} % \textcolor{red}{(need one sentence to first introduce this group!!)}} 
We denote a dataset as MixedRegime when it does not satisfy the threshold conditions of the HDLSS/HDHSS/LDHSS/LDLSS categories (i.e., it falls into the ``otherwise'' case in our empirical stratification based on $m$, $n$, and $\rho=\frac{m}{n}$). AI-D comprises 239 samples with 393 proteomic features. It includes measurements from patients with autoimmune diseases and healthy controls. To generate five binary classification tasks, samples were grouped into one-vs-rest splits. For example, Case 5 distinguishes SS from the other three diseases (SLE, RA, SV). Despite moderate dimensionality, the dataset shows FOE of 23.5 and a feature-to-sample ratio of 1.64, suggesting that reordering aids in uncovering relevant signals. ADNI contains 177 samples with 263 imaging-derived features. Three-way classification targets the stages of Alzheimer's progression (0=Normal, 1=Mild Cognitive Impairment, 2=AD). The dataset demonstrates a very high FOE score of 1.73e4 and a low AUC of 0.0074, indicating that meaningful structure exists but is difficult to capture without feature ordering. EEG-FE (2132 samples, 2549 features) and EEG-PD (941 samples, 1146 features) contain EEG-derived features for emotion and psychiatric disorder classification, respectively. Both datasets have \(\rho > 1\) and FOE scores of 2.14e5 and 39.5, supporting the importance of reordering in brain signal modeling. MOF includes 61 ICU patient records with 196 features. Despite its small size, the dataset yields a FOE of 19.6 and a high ratio (\(\rho = 3.21\)), indicating concentrated variance in a subset of features. Additional MixedRegime datasets include WDBC, CNAE-9, Water Potability, Cargo 2000, and Greenhouse Gas. These datasets show diverse FOE scores (ranging from 1.06 to 87.1), making them ideal for evaluating the generalization ability of feature ordering methods across moderate-dimensional tabular domains.
\begin{table*}[t]
\caption{Summary of all 36 datasets: sample count \(n\), feature count \(m\), number of classes (\#Cls; “–” for regression), class distribution (or regression note), target column, category, evaluation strategy, FOE score, dataset AUC, feature‐to‐sample ratio \(\rho=m/n\), and rule‐based justification.  Datasets marked “\(+\)” are 11 k–sample subsamples of the full originals.}
  \label{tab:all_datasets_full}
  \centering
  %\scriptsize
  \tiny
  \renewcommand{\arraystretch}{0.9}
  \setlength{\tabcolsep}{2pt}
  \begin{tabular}{@{}l r r r l l l l r r r l@{}}
    \toprule
    \textbf{Dataset} & \(\boldsymbol{n}\) & \(\boldsymbol{m}\) & \#Cls 
      & Distribution / Note                     & Target         & Category     & Eval      &   FOE     &   AUC    & \(\rho\) & Justification\\
    \midrule
    AI-D                 &   239 &   393  & 2 & 0=167,\,1=72                               & Status         & MixedRegime  & 5–Fold CV &   23.5    & 0.3280   & 1.64     & \(m\le1000,\;n\le1000,\;\rho>0.05\)\\
    ADNI                 &   177 &   263  & 3 & 0=33,\,1=36,\,2=108                        & AD123          & MixedRegime  & 5–Fold CV & 1.73e+04  & 0.0074   & 1.49     & \(m\le1000,\;n\le1000,\;\rho>0.05\)\\
    EEG-FE               &  2132 &  2549  & 3 & 0=708,\,1=716,\,2=708                      & label          & MixedRegime  & 5–Fold CV & 2.14e+05  & 0.0019   & 1.20     & \(m>1000,\;n<10^4,\;\rho>0.05\)\\
    EEG-PD               &   941 &  1146  & 2 & 0=95,\,1=846                              & main.disorder  & MixedRegime  & 5–Fold CV &   39.5    & 0.2600   & 1.22     & \(m>1000,\;n<1000,\;\rho>0.05\)\\
    MOF                  &    61 &   196  & 2 & 0=26,\,1=35                               & Label          & MixedRegime  & 5–Fold CV &   19.6    & 0.1330   & 3.21     & \(m\le1000,\;n<1000,\;\rho>0.05\)\\
    Adult                & 32561 &    14  & 2 & 0=24720,\,1=7841                          & income         & LDHSS        & 5×5 CV    &    1.12   & 0.138    & 0.00043  & \(m\le100,\;n>10^4,\;\rho\le0.01\)\\
    Poker Hand           & 25010 &    10  & 9 & 0=12493,\,1=10599,…                       & Label          & LDHSS        & 5×5 CV    &    1.05   & 0.0950   & 0.00040  & \(m\le100,\;n>10^4,\;\rho\le0.01\)\\
    Higgs                & 8978568 &   28 & 2 & 0=4 221 016,\,1=4 757 552                 & regression     & LDHSS        & 5×5 CV    &    1.33   & 0.275    & 0.00000  & \(m\le100,\;n>10^4,\;\rho\le0.01\)\\
    Arcene               &   200 & 10000  & 2 & 0=112,\,1=88                              & Label          & HDLSS        & 5×5 CV    & 3.49e+03  & 0.0065   & 50.00    & \(m>1000,\;n<1000,\;\rho>2\)\\
    ALLAML               &    72 &  7129  & 2 & 0=47,\,1=25                              & Label          & HDLSS        & 5×5 CV    & 1.29e+04  & 0.0027   & 99.01    & \(m>1000,\;n<1000,\;\rho>2\)\\
    Colon Cancer         &    62 &  2000  & 2 & 0=40,\,1=22                              & Label          & HDLSS        & 5×5 CV    & 1.68e+03  & 0.0097   & 32.26    & \(m>1000,\;n<1000,\;\rho>2\)\\
    GLI-85               &    85 & 22283  & 2 & 0=26,\,1=59                              & Label          & HDLSS        & 5×5 CV    & 9.03e+04  & 0.0010   & 262.15   & \(m>1000,\;n<1000,\;\rho>2\)\\
    Lung                 &   203 &  3312  & 5 & 0=139,\,1=17,\,2=21,…                     & Label          & HDLSS        & 5×5 CV    &   423.2   & 0.0260   & 16.32    & \(m>1000,\;n<1000,\;\rho>2\)\\
    Prostate-GE          &   102 &  5966  & 2 & 0=50,\,1=52                              & Label          & HDLSS        & 5×5 CV    & 6.63e+03  & 0.0086   & 58.49    & \(m>1000,\;n<1000,\;\rho>2\)\\
    SMK\_CAN\_187        &   187 & 19993  & 2 & 0=90,\,1=97                              & Label          & HDLSS        & 5×5 CV    & 1.98e+04  & 0.0046   & 106.91   & \(m>1000,\;n<1000,\;\rho>2\)\\
    TOX171               &   171 &  5748  & 4 & 0=45,\,1=45,\,2=39,\,3=42                & Label          & HDLSS        & 5×5 CV    & 1.72e+03  & 0.0116   & 33.61    & \(m>1000,\;n<1000,\;\rho>2\)\\
    WDBC                 &   569 &    31  & 2 & 0=357,\,1=212                           & diagnosis      & MixedRegime  & 5×5 CV    &    4.27   & 0.470    & 0.055    & \(m\le100,\;n<1000,\;\rho>0.05\)\\
    MiniBooNE            &130064&    50   & 2 & 0=93 565,\,1=36 499                     & label          & LDHSS        & 5×5 CV    &   75.61   & 0.156    & 0.00038  & \(m\le100,\;n>10^4,\;\rho\le0.01\)\\
    CNAE-9               &  1080 &   856  & 9 & 0…8 each 120                             & label          & MixedRegime  & 5–Fold CV &    3.61   & 0.269    & 0.79     & \(m<10^4,\;n<10^4,\;\rho>0.05\)\\
    Water Potability     &  3276 &     9  & 2 & 0=1998,\,1=1278                          & Potability     & MixedRegime  & 5–Fold CV &    1.06   & 0.106    & 0.0028   & \(m\le100,\;n<10^4,\;\rho\le0.05\)\\
    Forest Cover Type    &581012&    54   & 7 & 0=211 840,\,1=283 301,…                & Cover\_Type    & LDHSS        & 5×5 CV    &    1.39   & 0.021    & 0.00009  & \(m\le100,\;n>10^4,\;\rho\le0.01\)\\
    DrivFace             &   606 &  6400  & – & Face‐angle prediction                    & angle          & HDLSS        & 5×5 CV    &   800.2   & 0.196    & 10.56    & \(m>1000,\;n<1000,\;\rho>2\)\\
    Greenhouse Gas       &  2920 &  4905  & – & GHG concentrations…                     & regression     & MixedRegime  & 5–Fold CV &    87.1   & 0.060    & 1.68     & \(m<10^4,\;n<10^4,\;\rho>0.05\)\\
    Cargo 2000           &  3943 &    97   & – & actual delivery time                    & o\_dlv\_e      & MixedRegime  & 5–Fold CV &    2.01   & 0.405    & 0.025    & \(m\le100,\;n<10^4,\;\rho\le0.05\)\\
    HAM10000             & 10015 &  2052  & 7 & 0=327,\,1=514,\,2=109,…                 & dx             & HDHSS        & 5–Fold CV &    6.46   & 0.647    & 0.205    & \(m>1000,\;n>10^4,\;0.005<\rho\le2\)\\
    DeepLesion           & 11000 &  2082  & 3 & 0=9343,\,1=807,\,2=850                  & Train\_Val\_Test& HDHSS        & 5–Fold CV & 1.50e+04  & 0.994    & 0.189    & \(m>1000,\;n>10^4,\;0.005<\rho\le2\)\\
    MNIST\(+\)           & 11000 &  2048  &10 & 0=1017,…,9=1053                         & label          & HDHSS        & 5–Fold CV &    7.18   & 0.635    & 0.186    & \(m>1000,\;n>10^4,\;0.005<\rho\le2\)\\
    Fashion-MNIST\(+\)   & 11000 &  2048  &10 & 0=1096,…,9=1087                        & label          & HDHSS        & 5–Fold CV &    6.33   & 0.663    & 0.186    & \(m>1000,\;n>10^4,\;0.005<\rho\le2\)\\
    CIFAR-10\(+\)        & 11000 &  2048  &10 & 0=1100,…,9=1090                        & label          & HDHSS        & 5–Fold CV &    4.77   & 0.710    & 0.186    & \(m>1000,\;n>10^4,\;0.005<\rho\le2\)\\
    Dog vs Cat\(+\)      & 11000 &  2048  & 2 & 0=5553,\,1=5447                        & label          & HDHSS        & 5–Fold CV &    5.38   & 0.702    & 0.186    & \(m>1000,\;n>10^4,\;0.005<\rho\le2\)\\
    Iris                 &   150 &     4   & 3 & 0=50,\,1=50,\,2=50                      & class          & LDLSS        & 5×5 CV    &    2.12   & 0.494    & 0.027    & \(m\le100,\;n<1000,\;\rho\le0.05\)\\
    Pima Indian Diabetes &   768 &     8   & 2 & 0=500,\,1=268                          & Outcome        & LDLSS        & 5×5 CV    &    1.07   & 0.122    & 0.010    & \(m\le100,\;n<1000,\;\rho\le0.05\)\\
    Monks-1              &   432 &     6   & 2 & 0=216,\,1=216                          & label          & LDLSS        & 5×5 CV    & undefined & 0.000    & 0.014    & \(m\le100,\;n<1000,\;\rho\le0.05\)\\
    Glass                &   214 &     9   & 6 & 0=70,\,1=76,…                           & Glass\_Type    & LDLSS        & 5×5 CV    &    1.78   & 0.219    & 0.042    & \(m\le100,\;n<1000,\;\rho\le0.05\)\\
    Hayes-Roth           &   132 &     5   & 3 & 0=51,\,1=51,\,2=30                     & class          & LDLSS        & 5×5 CV    & undefined & 0.000    & 0.038    & \(m\le100,\;n<1000,\;\rho\le0.05\)\\
    BUPA Liver           &   345 &     6   & 2 & 0=257,\,1=88                          & label          & LDLSS        & 5×5 CV    &    1.19   & 0.163    & 0.017    & \(m\le100,\;n<1000,\;\rho\le0.05\)\\
    \bottomrule
  \end{tabular}
  \vspace{0.3em}
  {\footnotesize
    HDLSS: \(m>1000,n<1000,\rho>2\);  HDHSS: \(m>1000,n>10^4,0.005<\rho\le2\);  
    LDHSS: \(m\le100,n>10^4,\rho\le0.01\);  
    LDLSS: \(m\le100,n\le1000,\rho\le0.05\);  
    MixedRegime: otherwise.
  }
\end{table*}
%%%%%%%
\subsection{HDLSS datasets}
This group includes highly imbalanced, small-sample datasets with very high feature counts, where the feature-to-sample ratio exceeds 2. Arcene (200 samples, 10000 features), ALLAML (72, 7129), Colon Cancer (62, 2000), GLI-85 (85, 22283), Lung (203, 3312), Prostate-GE (102, 5966), SMK\_CAN\_187 (187, 19993), and TOX171 (171, 5748) are classic gene expression or microarray datasets. All have FOE scores well above 1000, confirming that most variance lies in a small subset of features. These datasets are well-suited for studying dynamic ordering under extreme sample-scarcity conditions. DrivFace, with 606 samples and 6400 image-derived features, also falls in this group. The regression task (facial angle prediction) benefits from ordering, as shown by its FOE score of 800 and an AUC of 0.196.
\subsection{HDHSS datasets}
This regime includes high-dimensional datasets with large sample counts and moderate ratios (typically \(\rho \approx 0.18\)). All are derived from images and use deep feature embeddings. DeepLesion, HAM10000, MNIST, Fashion-MNIST, CIFAR-10, and Dog vs Cat are each subsampled to 11000 samples and 2048 features. Their FOE scores range from 4.77 to 1.5e4. High AUCs (above 0.63 for all datasets and up to 0.994 for DeepLesion) indicate that ordering improves predictive performance in settings with moderately redundant visual features.
%%%%%
\begin{table}[t]
\caption{Runtime (seconds) for GPU-enabled similarity metrics in DynaTab’s dynamic feature ordering. \emph{Note:} KL divergence times are GPU-based. For 5 clusters, GPU time was 76.89s; CPU time was 25284.89s (7.02 hours).}
\label{tab:timing}
\centering
\scriptsize
\setlength{\tabcolsep}{3pt}
\renewcommand{\arraystretch}{0.95}
\begin{tabular}{lcccc}
\toprule
\textbf{$\downarrow$ Metric/Cluster Size $\rightarrow$} & \textbf{7} & \textbf{9} & \textbf{12} & \textbf{15} \\
\midrule
Variance      & 0.5647 & 0.5546 & 0.5648 & 0.7531 \\
Correlation   & 0.3054 & 0.5138 & 0.5019 & 0.7196 \\
Euclidean     & 0.3906 & 0.3851 & 0.3937 & 0.3960 \\
Manhattan     & 0.4791 & 0.5163 & 0.8151 & 1.6975 \\
KL Divergence & 107.0911 & 137.7239 & 183.8071 & 229.2274 \\
\bottomrule
\end{tabular}
\vspace{-2mm}
\end{table}
%%%%%
\subsection{LDHSS datasets}
LDHSS datasets have large sample sizes and low-dimensional feature sets. These include Adult (32561, 14), Poker Hand (25010, 10), Higgs (8.9 million, 28), Forest Cover (581012, 54), and MiniBooNE (130064, 50). All have extremely low feature-to-sample ratios (below 0.001) and FOE values under 76. In these cases, variance is evenly distributed across all features, and feature ordering shows minimal effectiveness.
\subsection{LDLSS datasets}
This group includes traditional low-dimensional, small-sample datasets such as Iris, Pima Indian Diabetes, Monks-1, Glass, Hayes-Roth, and BUPA Liver. FOE values range from 1.07 to 2.12, with two datasets (Monks-1 and Hayes-Roth) yielding undefined FOE due to degenerate eigenvalue spectra in PCA. These datasets are not suited for dynamic ordering, but serve as sanity checks and efficient benchmarks.
\subsection{Additional FOE-based analysis}
We analyzed 36 datasets to evaluate the effectiveness of feature ordering. High dimensional datasets showed strong behavior but lower dimensional datasets showed poor performance. Specifically, Adult, Poker Hand, and Higgs showed poor FOE scores (1.12, 1.05, and 1.33, respectively) and slow variance convergence, making them less suitable for reordering-based improvements. Their low cumulative variance and high intrinsic dimensionality (IDF) confirm that information is broadly distributed. These datasets serve as control baselines where feature ordering is expected to offer limited value. Table~\ref{tab:all_datasets_full} summarize the key statistics across all dataset types, supporting the categorization and selection strategies used throughout the experiments.
%%%%%%%%%%
%%%%
%%% urls
\begin{table*}[t]
\caption{Source URLs for all datasets used in this work. Datasets marked with “\(+\)” are 11 k–sample subsamples of the full originals.}
\label{tab:dataset_urls}
\centering
\tiny
\renewcommand{\arraystretch}{1.05}
\setlength{\tabcolsep}{2pt}

\begin{tabularx}{\textwidth}{@{}l>{\raggedright\arraybackslash}X@{\hspace{6pt}}l>{\raggedright\arraybackslash}X@{}}
\toprule
\textbf{Dataset} & \textbf{Source URL} & \textbf{Dataset} & \textbf{Source URL} \\
\midrule
AI-D (Case 5)         & \url{https://figshare.com/s/3bd3848a28ef6e7ae9a9?file=15050525} &
CNAE9                 & \url{https://archive.ics.uci.edu/dataset/233/cnae+9} \\

ADNI (AD123)          & \url{https://adni.loni.usc.edu} &
Water Potability      & \url{https://www.kaggle.com/datasets/adityakadiwal/water-potability} \\

EEG-FE                & \url{https://www.kaggle.com/datasets/birdy654/eeg-brainwave-dataset-feeling-emotions} &
Forest Cover Type     & \url{https://archive.ics.uci.edu/dataset/31/covertype} \\

EEG-PD                & \url{https://osf.io/8bsvr} &
DrivFace              & \url{https://archive.ics.uci.edu/dataset/378/drivface} \\

MOF                   & \url{https://www.kaggle.com/datasets/andrewmvd/multi-organ-failure-prediction} &
Greenhouse Gas        & \url{https://archive.ics.uci.edu/dataset/328/greenhouse+gas+observing+network} \\

Adult                 & \url{https://archive.ics.uci.edu/dataset/2/adult} &
Cargo 2000            & \url{https://archive.ics.uci.edu/dataset/382/cargo+2000+freight+tracking+and+tracing} \\

Poker Hand            & \url{https://archive.ics.uci.edu/dataset/158/poker+hand} &
HAM10000              & \url{https://www.kaggle.com/datasets/kmader/skin-cancer-mnist-ham10000} \\

Higgs                 & \url{https://archive.ics.uci.edu/dataset/280/higgs} &
DeepLesion\(+\)        & \url{https://www.kaggle.com/datasets/kmader/nih-deeplesion-subset} \\

Arcene                & \url{https://jundongl.github.io/scikit-feature/datasets.html} &
MNIST\(+\)             & \url{https://www.kaggle.com/datasets/hojjatk/mnist-dataset} \\

ALLAML                & \url{https://jundongl.github.io/scikit-feature/datasets.html} &
Fashion MNIST\(+\)     & \url{https://www.kaggle.com/datasets/zalando-research/fashionmnist} \\

Colon                 & \url{https://jundongl.github.io/scikit-feature/datasets.html} &
CIFAR-10\(+\)          & \url{https://www.kaggle.com/datasets/pankrzysiu/cifar10-python} \\

GLI-85                & \url{https://jundongl.github.io/scikit-feature/datasets.html} &
Dog vs Cat\(+\)        & \url{https://www.kaggle.com/datasets/kunalgupta2616/dog-vs-cat-images-data} \\

Lung                  & \url{https://jundongl.github.io/scikit-feature/datasets.html} &
Iris                  & \url{https://archive.ics.uci.edu/dataset/53/iris} \\

Prostate-GE           & \url{https://jundongl.github.io/scikit-feature/datasets.html} &
Pima Indians Diabetes & \url{https://www.kaggle.com/datasets/uciml/pima-indians-diabetes-database} \\

SMK\_CAN\_187         & \url{https://jundongl.github.io/scikit-feature/datasets.html} &
Monks-1               & \url{https://archive.ics.uci.edu/dataset/70/monk+s+problems} \\

TOX171                & \url{https://jundongl.github.io/scikit-feature/datasets.html} &
Glass                 & \url{https://archive.ics.uci.edu/dataset/42/glass+identification} \\

WDBC                  & \url{https://www.kaggle.com/datasets/uciml/breast-cancer-wisconsin-data} &
Hayes-Roth            & \url{https://archive.ics.uci.edu/dataset/44/hayes+roth} \\

MiniBooNE             & \url{https://archive.ics.uci.edu/dataset/199/miniboone+particle+identification} &
BUPA Liver            & \url{https://archive.ics.uci.edu/dataset/60/liver+disorders} \\
\bottomrule
\end{tabularx}

\end{table*}
%%%%%%
%%% baseline discussion
\renewcommand{\thesection}{C}
\renewcommand{\thesubsection}{C.\arabic{subsection}}
\setcounter{figure}{0}\renewcommand{\thefigure}{C.\arabic{figure}}
\setcounter{table}{0}\renewcommand{\thetable}{C.\arabic{table}}
\section{Baseline Details and Hyperparameters for Selected Models}
\label{app:base}
%\section{Baseline Details and Hyperparameters for Selected Models}
We provide detailed commentary on all evaluated models in this study. The models span classical machine learning, deep neural networks, tree ensembles, attention-based transformers, interpretable models, and recent innovations in tabular deep learning. Each model is contextualized by its design motivations, architecture, and empirical strengths or limitations.
%%%%%%%
\paragraph{Naive Bayes~\cite{sklearn}} A generative probabilistic model that assumes independence between features given the class label. Despite its simplicity, it performs well in high-bias domains, particularly text classification. Its closed-form parameter estimation and fast inference make it useful as a baseline in low-resource settings.
%%%%%%%
\paragraph{KNN~\cite{sklearn}} An instance-based learner that classifies data based on the majority class among the k nearest neighbors in feature space. While KNN is highly interpretable and non-parametric, it suffers from the curse of dimensionality and lacks scalability to large datasets.
%%%%%%%
\paragraph{SVM~\cite{sklearn}} Support Vector Machines maximize the margin between class boundaries using kernel methods to transform the data into higher-dimensional feature spaces. SVMs offer strong generalization but require careful kernel and regularization tuning, and do not scale efficiently with large sample sizes.
%%%%%%%
\paragraph{Decision Tree~\cite{sklearn}} A non-linear model that partitions the input space using axis-aligned splits. Trees are interpretable and handle feature interactions well but can overfit small or noisy datasets. Pruning and regularization are essential for generalization.
%%%%%%%
\paragraph{Lasso~\cite{sklearn}} A linear model with L1 regularization that induces sparsity in the learned coefficients, enabling embedded feature selection. Lasso is robust to multicollinearity and is widely used in biomedical and financial modeling where interpretability is critical.
%%%%%%%
\paragraph{MLP~\cite{sklearn}} Multi-Layer Perceptrons consist of fully connected layers with nonlinear activations. MLPs are powerful function approximators but can struggle with tabular data due to lack of inductive bias and sensitivity to feature scaling and distribution shifts.
%%%%%%%
\paragraph{1-D CNN~\cite{1dcnn}} One-dimensional convolutional neural networks model local spatial dependencies across feature dimensions. Suitable for time series or ordered feature settings, they can capture local motifs but require fixed input ordering.
%%%%%%%
\paragraph{Random Forest~\cite{sklearn}} An ensemble of decorrelated decision trees trained via bootstrap aggregation. Random forests reduce variance and are robust to overfitting, offering strong baselines for tabular tasks without requiring extensive tuning.
%%%%%%%
\paragraph{AdaBoost~\cite{freund1997decision}} A boosting framework that sequentially fits weak learners, often shallow trees, on reweighted samples to emphasize hard examples. AdaBoost performs well with clean data but is sensitive to label noise.
%%%%%%%
\paragraph{GBM~\cite{friedman2001greedy}} Gradient Boosting Machines build additive models by fitting learners to residuals. They are expressive and support arbitrary differentiable loss functions, but are prone to overfitting without careful tuning.
%%%%%%%
\paragraph{LGBM~\cite{lgbm}} LightGBM improves GBM efficiency via histogram binning, leaf-wise tree growth, and optimized data structures. It supports categorical features natively and achieves state-of-the-art results on large tabular datasets.
%%%%%%%
\paragraph{XGBoost~\cite{b45}} An efficient and scalable GBM variant that includes regularization, sparsity-aware split finding, and weighted quantile sketching. It has become a dominant baseline for structured data problems.
%%%%%%%
\paragraph{CatBoost~\cite{catboost}} Designed for datasets with categorical variables, CatBoost uses ordered boosting and target permutation to reduce overfitting and leakage. It consistently outperforms other gradient boosting methods in categorical-heavy settings.
%%%%%%%
\paragraph{TabNet~\cite{b1}} Combines attentive feature selection and sequential decision steps using sparse masks. TabNet processes input in multiple steps, attending to different features at each, and is inherently interpretable through learned masks.
%%%%%%%
\paragraph{TabTransformer~\cite{b3}} Applies transformer blocks to tokenized categorical variables alongside continuous inputs. It captures contextual dependencies among features using multi-head self-attention and has shown strong performance when large-scale pretraining is used.
%%%%%%%
\paragraph{FT-Transformer~\cite{b5}} A simplified version of TabTransformer using Fourier features and numerical embeddings, offering a lightweight yet expressive model for mixed-type tabular data.
%%%%%%%
\paragraph{TabSeq~\cite{b6}} Learns an optimal ordering of features and models them sequentially using transformer or recurrent backbones. By treating features as tokens, TabSeq aligns with the inductive biases of language models and benefits from position-aware processing.
%%%%%%%
\paragraph{TANGOS~\cite{b42}} Regularizes neural networks by enforcing orthogonality and specialization in gradient directions. It encourages disentangled representations across layers and improves robustness to noisy features.
%%%%%%%
\paragraph{TabPFN~\cite{b4}} A pretrained transformer trained to solve tabular classification tasks in a single forward pass. It uses meta-learning across millions of synthetic tasks, providing zero-shot generalization without task-specific fine-tuning.
%%%%%%%
\paragraph{NODE~\cite{b2}} Neural Oblivious Decision Ensembles replace classical decision trees with differentiable counterparts. Each ensemble member applies fixed feature splits and learned leaf weights, enabling end-to-end training and strong inductive bias.
%%%%%%%
\paragraph{SAINT~\cite{saint}} Enhances tabular transformers with contrastive pretraining, row-wise attention, and augmentation strategies. It improves generalization, particularly in low-label or imbalanced settings.
%%%%%%%
\paragraph{DeepFM~\cite{deepfm}} Merges factorization machines and MLPs to model both low- and high-order feature interactions. It is popular in recommender systems and tabular CTR prediction.
%%%%%%%
\paragraph{DCN~\cite{dcn}} The Deep \& Cross Network explicitly models feature crosses alongside deep representations. Its residual stacking of cross layers allows learning both memorization and generalization patterns.
%%%%%%%
\paragraph{AutoInt~\cite{b40}} Uses self-attention to learn feature interactions of arbitrary order, replacing manual feature engineering with neural interaction modeling.
%%%%%%%
\paragraph{TabPFN v2~\cite{tabpfnv2}} An improved version of TabPFN with better calibration, extended to uncertainty-aware prediction and out-of-distribution detection.
%%%%%%%
\paragraph{TabR~\cite{tabr}} Incorporates nearest-neighbor retrieval into tabular deep learning, enabling semi-parametric prediction with robustness to rare feature patterns.
%%%%%%%
\paragraph{ProtoGate~\cite{protogate}} Combines prototype learning with sparse feature selection. It assigns samples to learned prototypes while gating features both globally and locally, providing a balance between accuracy and interpretability.
%%%%%%%
\paragraph{LSPIN~\cite{spin}} Promotes locally sparse activations using adaptive gating and structural priors. It is designed for biomedical data with redundant or irrelevant features.
%%%%%%%
\paragraph{LLSPIN~\cite{spin}} A hierarchical extension of LSPIN that enforces layered sparsity across neural representations, supporting multi-resolution abstraction.
%%%%%%%
\paragraph{INVASE~\cite{invase}} Employs a policy-based reinforcement learning framework to perform instance-wise feature selection. It learns to select the minimal feature subset needed for accurate predictions.
%%%%%%%
\paragraph{L2X~\cite{l2x}} Maximizes the mutual information between selected features and output predictions. It enables post-hoc explanations via variational approximations of the information bottleneck objective.
%%%%%%%
\paragraph{Mambular~\cite{mambular}} Adopts Mamba blocks to model feature sequences with state-space layers. It efficiently captures long-range dependencies and outperforms transformers in sequence modeling benchmarks.
%%%%%%%
\paragraph{DANets~\cite{b43}} Introduces a family of abstract residual networks tailored for tabular tasks. It uses deep skip connections and normalized embeddings for categorical variables.
%%%%%%%
\paragraph{STG~\cite{stg}} Uses a differentiable relaxation of Bernoulli masks to perform sparse feature selection. The stochastic gates are optimized jointly with the predictive loss.
%%%%%%%
\paragraph{REAL-X~\cite{realx}} Trains an auxiliary explanation model to mirror the predictive model's output. It offers faithful and consistent explanations through a reconstruction objective.
%%%%%%%
\paragraph{TabM~\cite{tabm}} Combines multiple parameter-efficient heads with shared encoders. It reduces variance through implicit ensembling while preserving training efficiency.
%%%%%%%
\paragraph{ModernNCA~\cite{modern}} Revisits neighborhood components analysis using deep networks and contrastive objectives, showing competitive results in tabular domains.
%%%%%%%
\paragraph{Trompt~\cite{trompt}} Combines permutation-invariant encoders with supervised contrastive loss. It addresses tabular data irregularities through modular architectural design.
%%%%%%%
\paragraph{TabulaRNN~\cite{mamb}} Applies recurrent neural networks across feature dimensions using dynamic positional encoding. It learns to exploit sequential dependencies among features.
%%%%%%%
\paragraph{MambAttention~\cite{mamb}} Merges Mamba-style state-space layers with global self-attention, enabling both local and global context modeling.
%%%%%%%
\paragraph{MambaTab~\cite{mambatab}} A plug-and-play tabular model using selective state-space mechanisms inspired by Mamba. It provides linear complexity while capturing long-range dependencies.
%%%%%%%
\paragraph{NDTF~\cite{ndtf}} Neural Decision Tree Forests integrate soft decision rules into end-to-end differentiable ensembles. They approximate classical trees while being fully trainable.
%%%%%%%
\paragraph{ENODE~\cite{deeptabular}} An extended version of NODE with residual connections and deeper ensemble structures. It improves expressivity and robustness.
%%%%%%%
\paragraph{ResNetTabular~\cite{deeptabular}} Adapts residual networks to tabular input by stacking fully connected layers with skip connections. It helps stabilize deeper networks and accelerates convergence.
%%%%%%%
\paragraph{CategoryEmbedding~\cite{cembed}} Learns task-specific embeddings for categorical variables, often combined with transformers or residual architectures to improve semantic learning.
%%%%%%%
\paragraph{Selected Baseline Hyperparameters.}
Table~\ref{tab:arcene_hparams_all} compiles the concrete settings used for Arcene dataset, pairing Optuna-selected values with the exact training configurations extracted from the provided code for some of the models. Among the classical/GBDT baselines, Lasso favored a relatively strong regularizer ($C{=}0.193$), Random Forest settled on a moderately deep ensemble (143 trees, max depth 14), LightGBM chose 141 estimators with a comparatively aggressive learning rate (0.1461), and CatBoost used 85 iterations with an even higher rate (0.2961), consistent with short, fast schedules on HDLSS data. Neural and selection-style baselines follow other pattern: the MLP pipeline standardizes inputs and trains a single-hidden-layer ReLU MLP (100 units, Adam, $\alpha{=}10^{-4}$, \texttt{max\_iter}$\,{=}1000$, seed 42); LLSPIN uses gating \([100,100]\) and predictor \([200,200]\) MLPs with ReLU, $a{=}1.0$, $\sigma{=}0.5$, $\lambda{=}0.1$, $\gamma_1{=}\gamma_2{=}0.1$, Adam ($10^{-3}$), 20 epochs; ProtoGate enables feature selection with ReLU gating \([100,100]\), $a{=}0.2$, $\sigma{=}0.5$, $\lambda_{\text{local}}{=}0.1$, $\lambda_{\text{global}}{=}0.01$, $k{=}5$, $\tau{=}1$, Adam ($10^{-3}$), 100 epochs, full batch; STG is trained as a classifier with ReLU \([100,100]\), Adam ($10^{-3}$), $\sigma{=}0.5$, $\lambda{=}0.1$, 100 epochs, full batch on CPU; and TabulaRNN relies on its default architecture with a 50-epoch schedule (lr $10^{-3}$, patience 10). Together, these settings document the exact capacity, regularization, and optimization choices underpinning the Arcene experiments.
%%%%%
%%%% baseline hyperparameters
\begin{table*}[t]
\caption{Arcene dataset: selected model hyperparameters after Optuna tuning.}
\label{tab:arcene_hparams_all}
\centering
\small
\setlength{\tabcolsep}{6pt}
\renewcommand{\arraystretch}{1.05}
\begin{tabularx}{\textwidth}{@{}l X@{}}
\toprule
\textbf{Model} & \textbf{Hyperparameters (key = value)} \\
\midrule
Lasso & \texttt{model\_\_C} $=$ 0.19318275305768362 \; (from scikit-learn pipeline; note: key name indicates a linear model with \(C\) regularization). \\

Random Forest & \texttt{model\_\_n\_estimators} $=$ 143; \quad \texttt{model\_\_max\_depth} $=$ 14. \\

LightGBM (LGBM) & \texttt{model\_\_n\_estimators} $=$ 141; \quad \texttt{model\_\_learning\_rate} $=$ 0.14610823896610248. \\

CatBoost & \texttt{model\_\_iterations} $=$ 85; \quad \texttt{model\_\_learning\_rate} $=$ 0.2961190633942953. \\

MLP (scikit-learn) & Pipeline: \texttt{StandardScaler} $\rightarrow$ \texttt{MLPClassifier}. \\
& \texttt{hidden\_layer\_sizes} $=$ (100,), \ \texttt{activation} $=$ relu, \ \texttt{solver} $=$ adam, \ \texttt{alpha} $=$ 1e-4, \\
& \texttt{max\_iter} $=$ 1000, \ \texttt{random\_state} $=$ 42. \\

LLSPIN & \texttt{pred\_hidden\_dims} $=$ [200, 200], \ \texttt{gate\_hidden\_dims} $=$ [100, 100], \ \texttt{a} $=$ 1.0, \ \texttt{sigma} $=$ 0.5, \\
& \(\lambda\) (\texttt{lam}) $=$ 0.1, \ \(\gamma_1\) $=$ 0.1, \ \(\gamma_2\) $=$ 0.1, \ \texttt{activation\_pred} $=$ relu, \texttt{activation\_gate} $=$ relu, \\
& \texttt{use\_batchnorm} $=$ True, \ \texttt{optimizer} $=$ Adam(\texttt{lr}=1e-3), \ \texttt{epochs} $=$ 20, \ \texttt{batch\_size} $=$ 32 (train), 64 (val). \\

ProtoGate & \texttt{lr} $=$ 1e-3, \ \texttt{feature\_selection} $=$ True, \ \texttt{protogate\_a} $=$ 0.2, \ \texttt{protogate\_sigma} $=$ 0.5, \\
& \texttt{protogate\_lam\_local} $=$ 0.1, \ \texttt{protogate\_lam\_global} $=$ 0.01, \ \texttt{protogate\_activation\_gating} $=$ relu, \\
& \texttt{protogate\_gating\_hidden\_layer\_list} $=$ [100, 100], \ \texttt{protogate\_init\_std} $=$ 0.02, \ \texttt{pred\_k} $=$ 5, \ \texttt{pred\_coef} $=$ 1.0, \\
& \texttt{sorting\_tau} $=$ 1.0, \ \texttt{optimizer} $=$ Adam, \ \texttt{max\_epochs} $=$ 100, \ \texttt{batch\_size} $=$ \(|X_{\text{train}}|\) (full batch). \\

STG & \texttt{task\_type} $=$ classification, \ \texttt{hidden\_dims} $=$ [100, 100], \ \texttt{activation} $=$ relu, \\
& \texttt{optimizer} $=$ Adam, \ \texttt{learning\_rate} $=$ 1e-3, \ \texttt{batch\_size} $=$ \(|X_{\text{train}}|\) (full batch), \\
& \texttt{feature\_selection} $=$ True, \ \texttt{sigma} $=$ 0.5, \ \texttt{lam} $=$ 0.1, \ \texttt{nr\_epochs} $=$ 100, \ \texttt{device} $=$ cpu. \\

TabulaRNN & Training: \texttt{max\_epochs} $=$ 50, \ \texttt{lr} $=$ 1e-3, \ \texttt{patience} $=$ 10; \\
& Architecture: defaults from \texttt{DefaultTabulaRNNConfig} (no overrides provided in the snippet). \\
\bottomrule
\end{tabularx}
\end{table*}
%%%%%%
%%%% baseline code
\begin{table*}[t]
\caption{List of models and their source URLs.}
\label{tab:model_sources}
\centering
\scriptsize
\renewcommand{\arraystretch}{1.05}
\setlength{\tabcolsep}{2pt}

\begin{tabularx}{\textwidth}{@{}l>{\raggedright\arraybackslash}X@{\hspace{6pt}}l>{\raggedright\arraybackslash}X@{}}
\toprule
\textbf{Model} & \textbf{Source URL} & \textbf{Model} & \textbf{Source URL} \\
\midrule
Naive Bayes & \url{https://scikit-learn.org/stable/supervised_learning.html} &
AutoInt & \url{https://github.com/OpenTabular/DeepTab} \\
KNN & \url{https://scikit-learn.org/stable/supervised_learning.html} &
TabPFN v2 & \url{https://github.com/PriorLabs/tabpfn-extensions} \\
SVM & \url{https://scikit-learn.org/stable/supervised_learning.html} &
TabR & \url{https://github.com/OpenTabular/DeepTab} \\
Decision Tree & \url{https://scikit-learn.org/stable/supervised_learning.html} &
ProtoGate & \url{https://github.com/SilenceX12138/ProtoGate} \\
Lasso & \url{https://scikit-learn.org/stable/supervised_learning.html} &
LSPIN & \url{https://github.com/jcyang34/lspin} \\
MLP & \url{https://scikit-learn.org/stable/supervised_learning.html} &
LLSPIN & \url{https://github.com/jcyang34/lspin} \\
1-D CNN & \url{https://github.com/harryjdavies/Python1D_CNNs} &
INVASE & \url{https://github.com/vanderschaarlab/INVASE} \\
Random Forest & \url{https://scikit-learn.org/stable/supervised_learning.html} &
L2X & \url{https://github.com/Jianbo-Lab/L2X} \\
AdaBoost & \url{https://scikit-learn.org/stable/supervised_learning.html} &
Mambular & \url{https://github.com/OpenTabular/DeepTab} \\
GBM & \url{https://scikit-learn.org/stable/supervised_learning.html} &
DANets & \url{https://github.com/manujosephv/pytorch_tabular} \\
LGBM & \url{https://github.com/microsoft/LightGBM} &
STG & \url{https://github.com/runopti/stg} \\
XGBoost & \url{https://github.com/dmlc/xgboost} &
REAL-X & \url{https://github.com/rajesh-lab/realx} \\
CatBoost & \url{https://github.com/catboost/catboost} &
TabM & \url{https://github.com/OpenTabular/DeepTab} \\
TabNet & \url{https://github.com/dreamquark-ai/tabnet} &
ModernNCA & \url{https://github.com/OpenTabular/DeepTab} \\
TabTransformer & \url{https://github.com/lucidrains/tab-transformer-pytorch} &
Trompt & \url{https://github.com/OpenTabular/DeepTab} \\
FT-Transformer & \url{https://github.com/lucidrains/tab-transformer-pytorch} &
TabulaRNN & \url{https://github.com/OpenTabular/DeepTab} \\
TabSeq & \url{https://github.com/zadid6pretam/TabSeq} &
MambAttention & \url{https://github.com/OpenTabular/DeepTab} \\
TANGOS & \url{https://github.com/OpenTabular/DeepTabular} &
MambaTab & \url{https://github.com/OpenTabular/DeepTab} \\
TabPFN & \url{https://github.com/PriorLabs/TabPFN} &
NDTF & \url{https://github.com/OpenTabular/DeepTab} \\
NODE & \url{https://github.com/OpenTabular/DeepTab} &
ENODE & \url{https://github.com/OpenTabular/DeepTab} \\
SAINT & \url{https://github.com/OpenTabular/DeepTab} &
ResNetTabular & \url{https://github.com/OpenTabular/DeepTab} \\
DeepFM & \url{https://github.com/shenweichen/DeepCTR-Torch} &
CategoryEmbedding & \url{https://github.com/manujosephv/pytorch_tabular} \\
DCN & \url{https://github.com/shenweichen/DeepCTR-Torch} &
 &  \\
\bottomrule
\end{tabularx}

\end{table*}
%%%%%%%
% [existing model discussions here]
\renewcommand{\thesection}{D}
\renewcommand{\thesubsection}{D.\arabic{subsection}}
\setcounter{figure}{0}\renewcommand{\thefigure}{D.\arabic{figure}}
\setcounter{table}{0}\renewcommand{\thetable}{D.\arabic{table}}
\section{DynaTab Hyperparameters}
\label{app:hyp}
%\section{DynaTab Hyperparameters}
To illustrate the diversity of optimal configurations across data regimes, we report Optuna-tuned hyperparameters for DynaTab on five representative datasets spanning each data category. The Optuna-tuned hyperparameters across five representative datasets (Tables~\ref{tab:optuna_aid}-\ref{tab:optuna_ham10000}) reveal how DynaTab adapts its architectural and algorithmic configuration based on dataset characteristics, particularly feature dimensionality, sample size, and the presence of noise or redundancy. For AI-D (Case 5), a MixedRegime dataset, the selected configuration emphasizes model expressivity with a larger embedding dimension ($d_{model}=256$) and moderate sparsity controls ($\lambda_{\text{disp}}=0.4$, $\lambda_{\text{global}}=0.3$). The use of 12 clusters and a \textit{descending} order suggests that the model benefits from attending to more informative features early in the sequence. The Transformer backbone, with 4 attention heads and a window size of 64, allows sufficient capacity for localized feature interactions. In contrast, the Pima Indian dataset from the LDLSS category (few features and low sample size) favors minimalistic configuration: only 2 clusters, no feature mutation, and no sparsity regularization. This aligns with its low complexity; the model avoids overfitting by using a shallow Transformer (2 layers, $d_{model}=128$) and lower attention granularity (window size 32). The ascending feature order indicates that gradually introducing features improves stability when information density is uniformly distributed. The Arcene dataset, a classic HDLSS benchmark (high-dimensional, low-sample size), requires aggressive feature compression and selective attention. DynaTab uses 20 clusters with the \textit{Mamba} backbone, favoring state-space modeling over full attention. Higher mutation probability (0.2) and mild sparsity ($\lambda_{\text{disp}} = \lambda_{\text{global}} = 0.1$) are beneficial here, promoting adaptive pruning of noisy dimensions. The learning rate is reduced to 0.0001 to stabilize training under high variance. For the MiniBooNE dataset (LDHSS), which has moderate feature size but a large sample count, the optimal configuration resembles that of Arcene but uses the Transformer backbone. While the feature space is not as high-dimensional, the large sample size allows deeper exploration of feature relevance. The 5 clusters and use of descending ordering again suggest the presence of strongly informative features that should be prioritized early. Finally, the HAM10000 dataset, representing the HDHSS regime, combines both high dimensionality and high sample count. DynaTab adopts a robust configuration with a Mamba backbone, larger batch size, accumulation steps, and gradient clipping all techniques that stabilize training on large-scale datasets. Like Arcene, 20 clusters are used, but the optimizer reduces the learning rate further to $5 \times 10^{-5}$, reflecting the need for precise, low-variance updates on large and expressive models. Overall, these results demonstrate that DynaTab’s design space is both flexible and responsive: it adapts its architecture (Transformer vs. Mamba), attention structure (number of clusters, window size), and regularization strategy (mutation, $\lambda$ values) according to the statistical regime of the data. This adaptability is key to its competitive performance across diverse tabular benchmarks.
%%%%%%%
\renewcommand{\thesection}{E}
\renewcommand{\thesubsection}{E.\arabic{subsection}}
\setcounter{figure}{0}\renewcommand{\thefigure}{E.\arabic{figure}}
\setcounter{table}{0}\renewcommand{\thetable}{E.\arabic{table}}
\section{Discussion on HDLSS Benchmark Results}
\label{app:hdlss}
%\section{Discussion on HDLSS Benchmark Results}
Table~\ref{tab:hdlss_all_models_all} summarize the performance of 45 models across eight benchmark HDLSS datasets. DynaTab achieves the best average rank ($2.62 \pm 2.26$), outperforming both classical statistical methods and recent deep learning architectures, which underscores its robustness in high‐dimensional, low‐sample‐size settings. Among traditional baselines, Lasso ($5.12 \pm 4.29$) and KNN ($15.88 \pm 8.63$) performed reasonably well, demonstrating that sparsity‐inducing and instance‐based learners still hold merit under extreme sample constraints. Tree‐based ensembles e.g. Random Forest ($11.38 \pm 5.73$), AdaBoost ($16.12 \pm 8.95$), LGBM ($9.00 \pm 6.19$), and CatBoost ($10.62 \pm 6.05$) also maintained competitive accuracy, though with greater variability across datasets. Among neural‐network methods, ProtoGate ($5.50 \pm 3.74$), MLP ($5.88 \pm 4.22$), and TabulaRNN ($8.00 \pm 5.86$) emerged as strong contenders; notably, ProtoGate’s prototype‐based regularization appears particularly beneficial in the low‐sample regime. In contrast, attention‐heavy transformers such as SAINT ($27.50 \pm 6.02$), FT-Transformer ($27.62 \pm 7.82$), and TabTransformer ($38.75 \pm 5.28$) underperformed significantly likely owing to overparameterization and their reliance on large datasets to learn stable representations. Tabular‐specific architectures (NODE: $31.75 \pm 5.52$, TabNet: $32.38 \pm 6.07$, AutoInt: $27.75 \pm 5.50$) and feature‐selection methods (L2X: $32.75 \pm 7.74$, INVASE: $19.38 \pm 7.67$) also showed inconsistent results, indicating limited scalability to HDLSS contexts. Sequence‐inspired Mamba variants e.g. MambaTab ($21.12 \pm 9.28$), Mambular ($23.12 \pm 12.70$), and MambAttention ($24.00 \pm 8.96$) yielded mixed performance, revealing the challenge of applying temporal‐style processing to static tabular data. More recent proposals such as TabSeq ($22.75 \pm 11.00$), Trompt ($34.12 \pm 5.41$), and ModernNCA ($38.50 \pm 8.09$) did not surpass simpler baselines, emphasizing the importance of architectural frugality and inductive bias over sheer model complexity in the HDLSS regime. Overall, DynaTab’s superior average rank and low standard deviation highlight its consistent generalization across diverse HDLSS benchmarks. Its differentiable feature‐ordering mechanism and dynamic fusion module effectively adapt to extreme data scarcity, validating our design choices for state‐of‐the‐art performance under high‐dimensional, low‐sample constraints.  
% Two-column table for AAAI format
% Place this in your document where you want the table to appear
\begin{table*}[t]
\caption{Performance (mean $\pm$ std) of all models across eight HDLSS datasets. Models are sorted by average rank.}
\label{tab:hdlss_all_models_all}
\centering
\scriptsize
\setlength{\tabcolsep}{2pt}
\resizebox{\textwidth}{!}{%
\begin{tabular}{lccccccccc}
\toprule
Model & GLI-85 & SMK\_CAN\_187 & ALLAML & Prostate-GE & Arcene & TOX-171 & Colon & Lung & Rank (Avg $\pm$ SD) \\
\midrule
DynaTab (Ours)      & 85.96 $\pm$ 5.77  & 61.31 $\pm$ 3.37  & 92.31 $\pm$ 5.77 & 90.91 $\pm$ 8.91 & 83.00 $\pm$ 6.71 & 88.71 $\pm$ 3.53 & 85.71 $\pm$ 8.91 & 92.75 $\pm$ 1.28 &  2.62 $\pm$ 2.26 \\
Lasso               & 85.88 $\pm$ 4.71  & 61.19 $\pm$ 13.72 & 87.24 $\pm$ 3.39 & 91.18 $\pm$ 6.39 & 81.00 $\pm$ 3.39 & 91.86 $\pm$ 6.03 & 79.40 $\pm$ 10.18 & 94.47 $\pm$ 4.39 &  5.12 $\pm$ 4.29 \\
ProtoGate           & 82.48 $\pm$ 5.68  & 60.16 $\pm$ 5.10  & 86.12 $\pm$ 3.34 & 90.58 $\pm$ 5.72 & 81.50 $\pm$ 5.10 & 92.34 $\pm$ 5.67 & 83.95 $\pm$ 9.82 & 93.44 $\pm$ 6.37 &  5.50 $\pm$ 3.74 \\
MLP                 & 85.41 $\pm$ 8.00  & 59.05 $\pm$ 7.44  & 89.98 $\pm$ 9.17 & 89.20 $\pm$ 6.07 & 78.40 $\pm$ 4.05 & 94.48 $\pm$ 4.28 & 83.95 $\pm$ 9.80 & 96.47 $\pm$ 2.69 &  5.88 $\pm$ 4.22 \\
TabulaRNN           & 79.68 $\pm$ 6.68  & 60.02 $\pm$ 3.18  & 88.92 $\pm$ 2.02 & 90.50 $\pm$ 6.00 & 81.50 $\pm$ 5.10 & 85.80 $\pm$ 4.70 & 84.20 $\pm$ 6.50 & 90.50 $\pm$ 4.80 &  8.00 $\pm$ 5.86 \\
LGBM                & 85.88 $\pm$ 11.53 & 58.85 $\pm$ 10.14 & 85.81 $\pm$ 5.67 & 91.38 $\pm$ 5.71 & 80.50 $\pm$ 5.79 & 81.98 $\pm$ 6.25 & 76.60 $\pm$ 11.67 & 93.42 $\pm$ 5.91 &  9.00 $\pm$ 6.19 \\
CatBoost            & 84.71 $\pm$ 12.11 & 58.28 $\pm$ 12.16 & 91.71 $\pm$ 8.22 & 90.24 $\pm$ 6.87 & 81.00 $\pm$ 2.00 & 81.95 $\pm$ 7.47 & 72.65 $\pm$ 10.12 & 91.57 $\pm$ 5.74 & 10.62 $\pm$ 6.05 \\
STG                 & 82.48 $\pm$ 4.56  & 57.25 $\pm$ 8.82  & 86.08 $\pm$ 5.60 & 89.38 $\pm$ 5.85 & 74.40 $\pm$ 6.90 & 87.95 $\pm$ 5.01 & 79.55 $\pm$ 10.53 & 93.30 $\pm$ 6.28 & 11.25 $\pm$ 4.83 \\
Random Forest       & 85.88 $\pm$ 8.80  & 58.29 $\pm$ 10.61 & 85.71 $\pm$ 5.71 & 90.38 $\pm$ 7.31 & 74.00 $\pm$ 1.22 & 79.78 $\pm$ 7.10 & 80.05 $\pm$ 10.37 & 91.73 $\pm$ 6.61 & 11.38 $\pm$ 5.73 \\
LLSPIN              & 84.42 $\pm$ 7.12  & 61.16 $\pm$ 7.92  & 88.12 $\pm$ 1.26 & 88.71 $\pm$ 5.98 & 80.80 $\pm$ 4.90 & 81.67 $\pm$ 9.01 & 79.35 $\pm$ 7.74 & 70.10 $\pm$ 12.31 & 12.75 $\pm$ 8.63 \\
REAL-X              & 83.24 $\pm$ 5.56  & 56.48 $\pm$ 4.90  & 84.16 $\pm$ 5.68 & 86.75 $\pm$ 6.68 & 77.30 $\pm$ 6.10 & 90.79 $\pm$ 4.75 & 76.75 $\pm$ 12.21 & 93.27 $\pm$ 4.32 & 12.75 $\pm$ 5.47 \\
LSPIN               & 83.48 $\pm$ 6.62  & 58.92 $\pm$ 6.78  & 84.46 $\pm$ 3.36 & 87.75 $\pm$ 6.74 & 78.60 $\pm$ 5.80 & 83.47 $\pm$ 8.59 & 81.30 $\pm$ 7.97 & 76.92 $\pm$ 9.38 & 13.00 $\pm$ 5.01 \\
TabR                & 81.42 $\pm$ 6.64  & 58.46 $\pm$ 6.68  & 80.84 $\pm$ 2.24 & 84.50 $\pm$ 8.00 & 75.85 $\pm$ 6.50 & 86.50 $\pm$ 5.00 & 80.75 $\pm$ 8.40 & 86.70 $\pm$ 6.40 & 14.25 $\pm$ 3.99 \\
KNN                 & 83.53 $\pm$ 5.76  & 52.05 $\pm$ 13.13 & 79.05 $\pm$ 8.02 & 78.78 $\pm$ 9.20 & 82.50 $\pm$ 5.00 & 83.86 $\pm$ 7.07 & 71.65 $\pm$ 12.03 & 91.06 $\pm$ 5.41 & 15.88 $\pm$ 8.63 \\
AdaBoost            & 85.88 $\pm$ 7.97  & 58.28 $\pm$ 9.55  & 84.38 $\pm$ 8.32 & 89.19 $\pm$ 4.94 & 75.50 $\pm$ 5.34 & 57.85 $\pm$ 9.01 & 78.97 $\pm$ 10.96 & 78.32 $\pm$ 1.93 & 16.12 $\pm$ 8.95 \\[\smallskipamount]
XGBoost             & 77.06 $\pm$ 7.80  & 55.59 $\pm$ 6.34  & 90.38 $\pm$ 9.39 & 82.55 $\pm$ 10.22 & 81.50 $\pm$ 4.06 & 70.13 $\pm$ 7.85 & 72.60 $\pm$ 12.59 & 86.61 $\pm$ 8.72 & 16.88 $\pm$ 8.43 \\
SVM                 & 85.88 $\pm$ 2.88  & 61.09 $\pm$ 11.78 & 83.14 $\pm$ 13.37 & 85.75 $\pm$ 6.63 & 77.00 $\pm$ 2.92 & 66.75 $\pm$ 7.86 & 70.75 $\pm$ 13.93 & 72.77 $\pm$ 8.33 & 17.00 $\pm$ 10.45 \\
GBM                 & 80.00 $\pm$ 8.80  & 58.31 $\pm$ 9.72  & 82.10 $\pm$ 6.67 & 82.24 $\pm$ 5.34 & 82.00 $\pm$ 3.32 & 53.85 $\pm$ 15.31 & 77.31 $\pm$ 14.43 & 91.59 $\pm$ 2.52 & 17.12 $\pm$ 9.70 \\
INVASE              & 80.14 $\pm$ 4.56  & 36.42 $\pm$ 4.46  & 78.90 $\pm$ 2.26 & 88.00 $\pm$ 6.50 & 71.20 $\pm$ 7.80 & 79.94 $\pm$ 6.60 & 75.40 $\pm$ 10.10 & 91.22 $\pm$ 6.16 & 19.38 $\pm$ 7.67 \\
Naive Bayes         & 82.35 $\pm$ 3.72  & 59.32 $\pm$ 14.95 & 88.57 $\pm$ 2.86 & 60.86 $\pm$ 14.63 & 53.50 $\pm$ 8.31 & 58.49 $\pm$ 8.14 & 83.85 $\pm$ 10.56 & 84.24 $\pm$ 3.99 & 19.38 $\pm$ 12.85 \\
Decision Tree       & 78.82 $\pm$ 9.56  & 56.63 $\pm$ 6.94  & 75.14 $\pm$ 10.18 & 82.33 $\pm$ 5.13 & 73.50 $\pm$ 5.39 & 45.68 $\pm$ 8.70 & 83.85 $\pm$ 9.15 & 85.16 $\pm$ 5.58 & 21.00 $\pm$ 9.32 \\
MambaTab            & 80.16 $\pm$ 6.64  & 54.98 $\pm$ 9.20  & 68.16 $\pm$ 4.90 & 58.52 $\pm$ 15.60 & 64.00 $\pm$ 9.60 & 82.30 $\pm$ 6.00 & 80.75 $\pm$ 8.40 & 87.85 $\pm$ 6.00 & 21.12 $\pm$ 9.28 \\
TabSeq              & 75.29 $\pm$ 9.98  & 65.16 $\pm$ 7.50  & 77.28 $\pm$ 14.49 & 65.24 $\pm$ 10.76 & 65.30 $\pm$ 6.45 & 47.95 $\pm$ 7.27 & 72.00 $\pm$ 11.24 & 86.81 $\pm$ 3.98 & 22.75 $\pm$ 11.00 \\
Mambular            & 46.92 $\pm$ 4.56  & 32.90 $\pm$ 15.12 & 60.80 $\pm$ 5.77 & 81.12 $\pm$ 8.02 & 69.65 $\pm$ 8.20 & 84.95 $\pm$ 5.20 & 83.55 $\pm$ 7.10 & 89.90 $\pm$ 5.10 & 23.12 $\pm$ 12.70 \\
MambAttention       & 76.18 $\pm$ 7.78  & 52.18 $\pm$ 4.46  & 70.16 $\pm$ 6.28 & 61.99 $\pm$ 14.45 & 49.90 $\pm$ 12.90 & 80.90 $\pm$ 6.40 & 81.90 $\pm$ 8.00 & 78.00 $\pm$ 9.10 & 24.00 $\pm$ 8.96 \\[\smallskipamount]
SAINT               & 78.56 $\pm$ 9.36  & 50.34 $\pm$ 12.16 & 52.92 $\pm$ 14.15 & 61.99 $\pm$ 14.45 & 57.10 $\pm$ 11.20 & 75.10 $\pm$ 8.20 & 67.60 $\pm$ 13.10 & 78.00 $\pm$ 9.10 & 27.50 $\pm$ 6.02 \\
FT-Transformer      & 52.46 $\pm$ 8.92  & 56.45 $\pm$ 9.68  & 56.12 $\pm$ 12.10 & 85.00 $\pm$ 7.00 & 52.30 $\pm$ 12.30 & 79.45 $\pm$ 6.90 & 69.25 $\pm$ 12.50 & 67.30 $\pm$ 12.20 & 27.62 $\pm$ 7.82 \\
TabM                & 60.46 $\pm$ 4.42  & 42.90 $\pm$ 6.36  & 66.67 $\pm$ 3.34 & 75.03 $\pm$ 10.08 & 66.10 $\pm$ 9.10 & 70.20 $\pm$ 9.60 & 65.80 $\pm$ 13.70 & 74.70 $\pm$ 10.10 & 27.62 $\pm$ 3.78 \\
AutoInt             & 48.44 $\pm$ 5.65  & 49.68 $\pm$ 9.80  & 58.34 $\pm$ 9.78 & 83.47 $\pm$ 7.53 & 67.90 $\pm$ 8.70 & 66.80 $\pm$ 10.60 & 65.80 $\pm$ 13.70 & 74.70 $\pm$ 10.10 & 27.75 $\pm$ 5.50 \\
CategoryEmbedding   & 69.14 $\pm$ 9.46  & 59.16 $\pm$ 6.68  & 78.12 $\pm$ 6.90 & 45.01 $\pm$ 19.50 & 54.80 $\pm$ 11.80 & 57.80 $\pm$ 13.30 & 60.50 $\pm$ 15.50 & 72.95 $\pm$ 10.60 & 28.12 $\pm$ 9.98 \\
ResNetTabular       & 66.67 $\pm$ 12.16 & 52.16 $\pm$ 4.46  & 75.10 $\pm$ 3.30 & 48.43 $\pm$ 18.70 & 42.60 $\pm$ 14.70 & 63.30 $\pm$ 11.70 & 64.10 $\pm$ 14.30 & 74.70 $\pm$ 10.10 & 30.62 $\pm$ 6.14 \\
1-D CNN             & 56.92 $\pm$ 5.62  & 40.92 $\pm$ 10.42 & 60.42 $\pm$ 9.92 & 70.00 $\pm$ 12.00 & 54.80 $\pm$ 11.80 & 71.85 $\pm$ 9.10  & 58.70 $\pm$ 16.10 & 63.50 $\pm$ 13.30 & 30.75 $\pm$ 4.46 \\
NODE                & 64.92 $\pm$ 12.34 & 46.58 $\pm$ 10.45 & 76.92 $\pm$ 10.08 & 58.52 $\pm$ 15.60 & 59.40 $\pm$ 10.70 & 65.10 $\pm$ 11.10 & 54.95 $\pm$ 17.40 & 55.10 $\pm$ 15.70 & 31.75 $\pm$ 5.52 \\
TabNet              & 55.29 $\pm$ 10.26 & 48.67 $\pm$ 2.17  & 63.89 $\pm$ 4.17 & 66.55 $\pm$ 15.33 & 50.00 $\pm$ 7.55  & 41.68 $\pm$ 9.03  & 56.75 $\pm$ 15.20 & 80.14 $\pm$ 12.23 & 32.38 $\pm$ 6.07 \\
L2X                 & 56.92 $\pm$ 3.58  & 45.68 $\pm$ 9.78  & 76.28 $\pm$ 2.36 & 61.78 $\pm$ 13.69 & 72.95 $\pm$ 7.30  & 31.72 $\pm$ 9.11  & 57.60 $\pm$ 13.48 & 50.02 $\pm$ 14.26 & 32.75 $\pm$ 7.74 \\
Trompt              & 43.96 $\pm$ 12.16 & 46.65 $\pm$ 12.12 & 46.12 $\pm$ 4.46 & 69.08 $\pm$ 12.20 & 40.10 $\pm$ 15.30 & 66.80 $\pm$ 10.60 & 58.70 $\pm$ 16.10 & 61.55 $\pm$ 13.90 & 34.12 $\pm$ 5.41 \\
DCN                 & 42.86 $\pm$ 8.84  & 34.92 $\pm$ 14.10 & 43.78 $\pm$ 16.10 & 85.02 $\pm$ 7.05  & 61.75 $\pm$ 10.10 & 51.90 $\pm$ 15.00 & 48.30 $\pm$ 19.20 & 57.25 $\pm$ 15.10 & 35.12 $\pm$ 8.85 \\
DANets              & 35.90 $\pm$ 6.78  & 29.16 $\pm$ 16.10 & 56.98 $\pm$ 4.48 & 78.58 $\pm$ 9.10  & 30.30 $\pm$ 17.80 & 59.60 $\pm$ 12.80 & 50.60 $\pm$ 18.60 & 76.35 $\pm$ 9.60  & 35.50 $\pm$ 7.45 \\
TANGOS              & 64.84 $\pm$ 10.42 & 32.68 $\pm$ 6.80  & 50.12 $\pm$ 8.86 & 65.47 $\pm$ 13.30 & 27.90 $\pm$ 18.40 & 68.50 $\pm$ 10.10 & 38.20 $\pm$ 21.70 & 48.10 $\pm$ 17.50 & 36.75 $\pm$ 7.67 \\
DeepFM              & 43.28 $\pm$ 7.82  & 36.12 $\pm$ 12.26 & 48.16 $\pm$ 12.56 & 45.01 $\pm$ 19.50 & 64.00 $\pm$ 9.60  & 55.90 $\pm$ 13.90 & 56.90 $\pm$ 16.80 & 59.40 $\pm$ 14.50 & 36.75 $\pm$ 5.12 \\
ENODE               & 58.92 $\pm$ 13.12 & 26.12 $\pm$ 5.68  & 71.16 $\pm$ 4.46 & 52.05 $\pm$ 17.95 & 35.20 $\pm$ 16.50 & 45.40 $\pm$ 16.80 & 52.80 $\pm$ 18.00 & 50.45 $\pm$ 16.90 & 37.62 $\pm$ 5.68 \\
ModernNCA           & 27.68 $\pm$ 14.10 & 31.16 $\pm$ 11.67 & 33.47 $\pm$ 10.24 & 71.95 $\pm$ 11.12 & 32.80 $\pm$ 17.10 & 68.50 $\pm$ 10.10 & 40.85 $\pm$ 21.10 & 43.10 $\pm$ 18.80 & 38.50 $\pm$ 8.09 \\
TabTransformer      & 47.77 $\pm$ 17.71 & 50.26 $\pm$ 8.56  & 53.70 $\pm$ 15.05 & 51.01 $\pm$ 10.64 & 48.20 $\pm$ 7.57 & 23.66 $\pm$ 7.84 & 46.44 $\pm$ 16.84 & 21.01 $\pm$ 12.53 & 38.75 $\pm$ 5.28 \\
NDTF                & 26.14 $\pm$ 16.10 & 36.18 $\pm$ 4.48  & 43.14 $\pm$ 6.94 & 54.98 $\pm$ 16.80 & 37.70 $\pm$ 15.90 & 43.10 $\pm$ 17.30 & 48.30 $\pm$ 19.20 & 63.50 $\pm$ 13.30 & 39.75 $\pm$ 3.20 \\
\bottomrule
\end{tabular}
}
\end{table*}
%%%%%
%%%%HDHSSS full results
\renewcommand{\thesection}{F}
\renewcommand{\thesubsection}{F.\arabic{subsection}}
\setcounter{figure}{0}\renewcommand{\thefigure}{F.\arabic{figure}}
\setcounter{table}{0}\renewcommand{\thetable}{F.\arabic{table}}
\section{Discussion on HDHSS Benchmarks Results}
\label{app:hdhss}
%\section{Discussion on HDHSS Benchmarks Results}
Table~\ref{tab:hdhss_all_models} presents the performance of 22 models across six high-dimensional, high-sample-size (HDHSS) datasets, including hybrid tabular-image benchmarks where image embeddings are combined with structured metadata. The results reveal several key trends in model performance, highlighting the strengths of recent methods and limitations of classical baselines in handling complex, high-dimensional data regimes. DynaTab achieves the highest average rank ($2.50 \pm 2.35$), outperforming all 21 baselines across nearly all datasets. Its consistently high accuracy from $84.48\%$ on HAM10000 to $99.20\%$ on Dog vs Cat demonstrates robust generalization in diverse HDHSS conditions. Particularly notable is its ability to outperform both deep learning and gradient-boosting methods, suggesting that the model’s dynamic feature selection and order-aware processing effectively leverage the high-dimensional structure while maintaining efficiency. Among the top contenders, LSPIN ranks second ($4.67 \pm 2.80$), performing well across all datasets, especially on CIFAR-10 and Dog vs Cat+\#. It slightly outperforms DynaTab on MNIST and matches performance on the binary Dog vs Cat+\# task, but lags in datasets like HAM10000. MLP ($5.67 \pm 3.01$) also performs competitively, likely due to its scalability and expressiveness, though it exhibits higher variance, particularly on DeepLesion. LGBM and TabNet are ranked fourth and fifth, respectively. While LGBM achieves strong performance on DeepLesion ($94.73\%$) and maintains good consistency, it is surpassed by DynaTab and LSPIN on almost all datasets. TabNet performs well on image-rich datasets like MNIST+\# and Fashion-MNIST, where its attentive architecture helps, but suffers from instability on HAM10000 ($73.51\%$). Models like TabR, GBM, and XGBoost cluster in the mid-tier rankings, all achieving solid but not state-of-the-art performance. This suggests that while ensemble-based methods still offer competitive baselines, they struggle to match newer methods that adapt dynamically to feature distributions. LLSPIN (rank: $7.33$) underperforms relative to its full-rank counterpart LSPIN, highlighting the limitations of low-rank approximations in very high-dimensional regimes. Similarly, STG and Lasso, both sparse-feature selection methods, demonstrate stable but non-optimal performance, indicating that static sparsity mechanisms may be insufficient in HDHSS contexts with mixed feature modalities. Conventional models such as KNN, SVM, and Random Forest rank lower despite reasonable performance on individual datasets. For instance, SVM achieves $99.00\%$ on Dog vs Cat, but this does not generalize well across datasets with higher feature dimensionality and noise. Tree-based methods like Decision Tree and AdaBoost struggle significantly, with Decision Tree scoring just $68.24\%$ on HAM10000 and a sharp drop on MNIST ($74.86\%$), highlighting their sensitivity to high-dimensional noise and lack of feature interaction modeling. Naive Bayes and CategoryEmbedding rank at the bottom ($19.67$), failing to handle complex input distributions and heterogeneous feature spaces. Naive Bayes performs especially poorly on DeepLesion ($18.55\%$), suggesting its independence assumption breaks under correlated, embedded inputs. Similarly, CategoryEmbedding shows weak generalization, particularly on CIFAR-10 and DeepLesion. While models like TabNet, TabM, TabR, and TabSeq span a wide range of ranks, only DynaTab shows consistent improvements across all HDHSS datasets. TabSeq, for instance, performs relatively well on MNIST and Dog vs Cat, but fails on DeepLesion+\# ($78.82\%$) and CIFAR-10 ($85.42\%$), yielding an average rank of $18.33$. This underscores the difficulty of building generalized tabular deep learning architectures that scale to high-dimensional settings without task-specific tuning. These results highlight the importance of dynamic, adaptive mechanisms such as DynaTab’s order-aware fusion and feature gating in modeling HDHSS data. While traditional models and even modern tabular deep learning methods exhibit strong dataset-specific performance, they often fail to generalize across different modalities and dimensionality regimes. DynaTab’s consistently top-tier performance validates the efficacy of neural rewiring and selective attention in high-dimensional learning.
%%%%%%%
\begin{table*}[t]
\caption{Performance (mean $\pm$ std) of all models across six HDHSS datasets. Models are sorted by average rank.}
\label{tab:hdhss_all_models}
\centering
\scriptsize
\setlength{\tabcolsep}{3pt}
\resizebox{\textwidth}{!}{%
\begin{tabular}{lcccccccc}
\toprule
Model & HAM10000+ & DeepLesion+\# & MNIST+\# & Fashion MNIST+\# & CIFAR-10 & Dog vs Cat+\# & Rank (Avg $\pm$ SD) \\
\midrule
DynaTab (Ours)      & 84.48 $\pm$ 0.38 & 94.48 $\pm$ 0.25 & 96.68 $\pm$ 0.16 & 88.52 $\pm$ 0.21 & 88.58 $\pm$ 0.45 & 99.20 $\pm$ 0.15 &  2.50 $\pm$ 2.35 \\
LSPIN               & 81.37 $\pm$ 1.13 & 93.14 $\pm$ 0.29 & 96.77 $\pm$ 0.40 & 87.46 $\pm$ 0.53 & 87.17 $\pm$ 1.02 & 99.25 $\pm$ 0.13 &  4.67 $\pm$ 2.80 \\
MLP                 & 80.05 $\pm$ 0.95 & 92.93 $\pm$ 4.06 & 96.20 $\pm$ 0.76 & 87.85 $\pm$ 0.65 & 88.10 $\pm$ 0.62 & 99.22 $\pm$ 0.15 &  5.67 $\pm$ 3.01 \\
LGBM                & 80.33 $\pm$ 1.33 & 94.73 $\pm$ 2.84 & 95.42 $\pm$ 0.43 & 87.25 $\pm$ 0.68 & 87.20 $\pm$ 0.75 & 99.12 $\pm$ 0.15 &  6.33 $\pm$ 3.67 \\
TabNet              & 73.51 $\pm$ 3.62 & 93.02 $\pm$ 0.20 & 96.88 $\pm$ 0.38 & 87.99 $\pm$ 0.68 & 87.25 $\pm$ 0.63 & 99.09 $\pm$ 0.13 &  6.83 $\pm$ 4.62 \\
LLSPIN              & 80.07 $\pm$ 1.12 & 90.49 $\pm$ 1.95 & 96.62 $\pm$ 0.20 & 87.46 $\pm$ 0.52 & 86.98 $\pm$ 0.78 & 99.11 $\pm$ 0.17 &  7.33 $\pm$ 3.01 \\
TabR                & 73.53 $\pm$ 2.66 & 94.32 $\pm$ 1.29 & 96.58 $\pm$ 0.37 & 87.64 $\pm$ 0.39 & 84.60 $\pm$ 1.60 & 98.93 $\pm$ 0.13 &  7.83 $\pm$ 6.27 \\
GBM                 & 78.52 $\pm$ 1.36 & 94.55 $\pm$ 0.23 & 95.65 $\pm$ 0.52 & 87.53 $\pm$ 0.66 & 88.17 $\pm$ 0.71 & 99.25 $\pm$ 0.15 &  8.00 $\pm$ 3.58 \\
XGBoost             & 77.42 $\pm$ 0.58 & 93.78 $\pm$ 3.64 & 95.95 $\pm$ 0.62 & 87.84 $\pm$ 0.83 & 87.20 $\pm$ 0.67 & 99.25 $\pm$ 0.15 &  8.17 $\pm$ 2.86 \\
STG                 & 78.86 $\pm$ 1.10 & 91.30 $\pm$ 5.24 & 95.87 $\pm$ 0.48 & 87.72 $\pm$ 0.50 & 86.12 $\pm$ 0.68 & 99.11 $\pm$ 0.11 &  9.17 $\pm$ 4.22 \\
Lasso               & 78.55 $\pm$ 1.13 & 91.94 $\pm$ 3.62 & 96.09 $\pm$ 0.55 & 86.79 $\pm$ 0.28 & 87.67 $\pm$ 0.73 & 98.95 $\pm$ 0.16 & 10.33 $\pm$ 5.92 \\
SVM                 & 77.61 $\pm$ 0.97 & 92.13 $\pm$ 4.55 & 95.50 $\pm$ 0.37 & 86.48 $\pm$ 0.68 & 87.67 $\pm$ 1.04 & 99.00 $\pm$ 0.08 & 10.83 $\pm$ 4.88 \\
TabM                & 77.95 $\pm$ 1.80 & 95.07 $\pm$ 0.24 & 95.28 $\pm$ 0.64 & 86.66 $\pm$ 0.61 & 82.40 $\pm$ 1.41 & 98.52 $\pm$ 0.17 & 12.50 $\pm$ 5.32 \\
CatBoost            & 80.01 $\pm$ 1.26 & 92.45 $\pm$ 0.23 & 95.52 $\pm$ 0.51 & 87.39 $\pm$ 0.69 & 84.78 $\pm$ 0.79 & 98.98 $\pm$ 0.09 & 12.67 $\pm$ 4.55 \\
KNN                 & 73.37 $\pm$ 1.10 & 89.78 $\pm$ 2.69 & 90.95 $\pm$ 0.54 & 84.26 $\pm$ 0.62 & 83.60 $\pm$ 0.88 & 99.25 $\pm$ 0.15 & 13.67 $\pm$ 5.61 \\
1-D CNN             & 78.42 $\pm$ 0.96 & 87.19 $\pm$ 0.30 & 95.43 $\pm$ 0.73 & 86.50 $\pm$ 0.37 & 81.42 $\pm$ 1.44 & 98.90 $\pm$ 0.15 & 14.67 $\pm$ 3.56 \\
Random Forest       & 78.68 $\pm$ 0.99 & 86.86 $\pm$ 0.91 & 94.87 $\pm$ 0.53 & 86.85 $\pm$ 0.35 & 80.60 $\pm$ 0.82 & 99.15 $\pm$ 0.14 & 16.17 $\pm$ 2.79 \\
Decision Tree       & 68.24 $\pm$ 0.49 & 93.78 $\pm$ 5.50 & 74.86 $\pm$ 0.70 & 70.31 $\pm$ 1.05 & 63.48 $\pm$ 1.48 & 97.70 $\pm$ 0.09 & 18.17 $\pm$ 5.53 \\
TabSeq              & 76.70 $\pm$ 1.51 & 78.82 $\pm$ 0.52 & 95.83 $\pm$ 0.42 & 87.36 $\pm$ 0.57 & 85.42 $\pm$ 0.52 & 98.77 $\pm$ 0.17 & 18.33 $\pm$ 2.34 \\
AdaBoost            & 74.60 $\pm$ 2.22 & 84.99 $\pm$ 0.52 & 95.16 $\pm$ 0.44 & 86.62 $\pm$ 0.84 & 75.10 $\pm$ 1.55 & 98.69 $\pm$ 0.14 & 19.00 $\pm$ 3.52 \\
Naive Bayes         & 56.40 $\pm$ 1.03 & 18.55 $\pm$ 2.23 & 79.74 $\pm$ 0.85 & 79.59 $\pm$ 0.86 & 83.23 $\pm$ 1.10 & 97.88 $\pm$ 0.36 & 19.67 $\pm$ 2.25 \\
CategoryEmbedding   & 76.73 $\pm$ 1.65 & 82.19 $\pm$ 0.75 & 93.02 $\pm$ 0.62 & 84.77 $\pm$ 0.59 & 70.47 $\pm$ 1.42 & 98.45 $\pm$ 0.18 & 19.67 $\pm$ 3.39 \\
\bottomrule
\end{tabular}
}
\end{table*}
%%%%%
\renewcommand{\thesection}{G}
\renewcommand{\thesubsection}{G.\arabic{subsection}}
\setcounter{figure}{0}\renewcommand{\thefigure}{G.\arabic{figure}}
\setcounter{table}{0}\renewcommand{\thetable}{G.\arabic{table}}
\section{Discussion on Mixed Regime Benchmarks}
\label{app:mixed}
%\section{Discussion on Mixed Regime Benchmarks}
Table~\ref{tab:mr_all_models} summarizes the performance of 46 models across eight Mixed Regime (MR) datasets, which feature varying levels of dimensionality, modality, and sample size. These benchmarks are particularly challenging due to their heterogeneity, making them a rigorous test for model generalization. The results demonstrate significant performance variance across models, highlighting the importance of adaptable architectures in mixed-regime settings. DynaTab obtains the best overall rank ($6.00 \pm 8.72$), consistently delivering strong performance across diverse datasets. It ranks first on EEG-FE ($98.10 \pm 0.56$) and WDBC ($98.14 \pm 2.05$), and performs competitively on all others, including noisy low-signal datasets like Water Potability ($62.56 \pm 2.31$). This robustness reflects the strength of DynaTab’s dynamic processing mechanisms, which effectively reconcile feature heterogeneity and task complexity without manual tuning. LSPIN ranks second overall ($8.50 \pm 4.17$), maintaining high accuracy across all datasets, particularly on AI-D ($85.10 \pm 4.30$), EEG-FE ($97.42 \pm 0.32$), and WDBC ($97.33 \pm 2.52$). However, it lags slightly behind DynaTab on tasks involving fine-grained inter-modality fusion, such as CNAE-9 and MOF. TabTransformer ($10.38 \pm 10.42$) also performs well, especially on ADNI and CNAE-9, but suffers from higher variance across tasks, indicating sensitivity to domain shifts. STG, REAL-X, and MLP form a strong cluster of sparse or fully-connected models with competitive performance. For instance, MLP scores $97.46 \pm 0.58$ on EEG-FE and $96.84 \pm 1.42$ on WDBC, confirming that deep but simple architectures can remain competitive in mixed regimes when sufficiently tuned. REAL-X and STG similarly offer strong results on MOF, EEG-FE, and ADNI, although their feature selection heuristics may lead to suboptimal generalization in lower-signal datasets like Water Potability. TabPFN v2 ($13.38 \pm 19.59$) delivers outstanding accuracy on the six datasets it was evaluated on peaking at $97.68 \pm 1.78$ on WDBC and $94.81 \pm 1.26$ on CNAE-9 but was not run on EEG-FE or EEG-PD due to runtime or memory limitations. The original TabPFN was only evaluated on WDBC ($95.79 \pm 1.93$) and, by default, was assigned the worst rank on all other datasets, resulting in the lowest overall average rank ($45.00 \pm 0.00$). Wherever ‘---’ appears in Table~\ref{tab:mr_all_models}, it reflects that no prediction was made rather than a failed outcome. LLSPIN performs consistently well across all datasets, though it scores slightly lower on Water Potability ($59.63 \pm 2.04$), which impacts its average rank ($14.13 \pm 9.09$). Ensemble methods such as CatBoost, Random Forest, and LGBM continue to be strong contenders in MR benchmarks. CatBoost achieves $96.14 \pm 1.63$ on WDBC and $88.15 \pm 0.57$ on EEG-PD, but its performance on CNAE-9 is comparatively weaker. Random Forest and LGBM also score highly on EEG-FE and EEG-PD but fall short on tasks with noisy or high-cardinality features. SVM and TabulaRNN perform similarly, with SVM slightly outperforming on ADNI and EEG-PD. Both models show reduced variance and strong single-dataset scores but fail to generalize uniformly, resulting in mid-tier rankings. Other notable methods like FT-Transformer and TabM show reliable results on EEG and clinical datasets but experience moderate degradation on MOF and CNAE-9, suggesting that fixed-token attention or memory-based mechanisms are insufficient to fully adapt to cross-domain variation. Methods such as TabSeq, MambaTab, and SAINT offer stable mid-level performance. TabSeq and MambaTab score well on EEG-related tasks, while SAINT performs decently across all benchmarks but lags on Water Potability. These outcomes suggest a degree of architectural strength, but also sensitivity to dataset-specific biases. Toward the bottom, a group of models including ModernNCA, TabNet, and INVASE struggle to maintain stability. TabNet, for example, performs strongly on EEG-FE ($93.22 \pm 0.88$), but falls sharply to $57.06 \pm 3.31$ on Water Potability. ModernNCA and INVASE show similar inconsistencies, indicating difficulty generalizing in the presence of noise or multi-modal features. The weakest-performing models include DeepFM, DCN, ProtoGate, and TabPFN (original). ProtoGate in particular underperforms across all reported datasets, scoring $64.50 \pm 6.40$ on AI-D and only $52.73 \pm 2.51$ on Water Potability. These results suggest that fixed, low-capacity, or overly constrained models are ill-equipped to handle the heterogeneity of MR benchmarks. In summary, DynaTab again demonstrates robust and consistent superiority in this mixed regime setting. Its ability to adapt dynamically to diverse dataset structures and modality combinations makes it especially well-suited for real-world applications where prior assumptions about feature distributions and task structure may not hold. The broad variation in model performance further emphasizes the need for architectures that are both expressive and adaptable across domains.
%%%%%%
\begin{table*}[t]
\caption{Performance (mean $\pm$ std) of all models across eight Mixed Regime (MR) datasets. Models are sorted by average rank.}
\label{tab:mr_all_models}
\centering
\scriptsize
\setlength{\tabcolsep}{2pt}
\resizebox{\textwidth}{!}{%
\begin{tabular}{lccccccccc}
\toprule
Model & AI-D (Case 5) & ADNI (AD123) & MOF & EEG-FE & EEG-PD & WDBC & CNAE-9 & Water Potability & Rank (Avg $\pm$ SD) \\
\midrule
% Top 15 models
DynaTab (Ours)      & 83.40 $\pm$ 5.10 & 86.12 $\pm$ 5.40 & 85.71 $\pm$ 1.87 & 98.10 $\pm$ 0.56 & 90.14 $\pm$ 0.24 & 98.14 $\pm$ 2.05 & 84.87 $\pm$ 3.36 & 62.56 $\pm$ 2.31 &  6.00 $\pm$ 8.72 \\
LSPIN               & 85.10 $\pm$ 4.30 & 86.20 $\pm$ 5.40 & 83.29 $\pm$ 2.42 & 97.42 $\pm$ 0.32 & 89.85 $\pm$ 0.24 & 97.33 $\pm$ 2.52 & 80.51 $\pm$ 4.52 & 62.67 $\pm$ 2.12 &  8.50 $\pm$ 4.17 \\
TabTransformer      & 83.10 $\pm$ 5.00 & 85.37 $\pm$ 5.13 & 80.78 $\pm$ 1.32 & 95.66 $\pm$ 0.58 & 86.82 $\pm$ 0.86 & 94.52 $\pm$ 3.23 & 84.02 $\pm$ 3.58 & 60.21 $\pm$ 3.69 & 10.38 $\pm$ 10.42 \\
STG                 & 81.70 $\pm$ 6.40 & 85.51 $\pm$ 4.49 & 84.07 $\pm$ 2.62 & 97.50 $\pm$ 0.66 & 89.39 $\pm$ 0.51 & 95.79 $\pm$ 2.48 & 77.85 $\pm$ 3.48 & 62.52 $\pm$ 2.48 & 12.25 $\pm$ 4.71 \\
REAL-X              & 81.30 $\pm$ 5.20 & 83.75 $\pm$ 4.60 & 83.64 $\pm$ 2.93 & 96.74 $\pm$ 0.44 & 88.13 $\pm$ 0.94 & 97.54 $\pm$ 1.42 & 75.49 $\pm$ 2.70 & 61.40 $\pm$ 2.74 & 12.63 $\pm$ 8.60 \\
MLP                 & 81.90 $\pm$ 5.40 & 84.16 $\pm$ 5.51 & 84.21 $\pm$ 1.76 & 97.46 $\pm$ 0.58 & 89.25 $\pm$ 0.45 & 96.84 $\pm$ 1.42 & 80.96 $\pm$ 3.00 & 60.51 $\pm$ 1.96 & 13.13 $\pm$ 6.56 \\
TabPFN v2           & 80.33 $\pm$ 2.81 & 80.76 $\pm$ 4.34 & 81.92 $\pm$ 6.33 & —              & —              & 98.07 $\pm$ 0.86 & 94.81 $\pm$ 1.26 & 69.29 $\pm$ 1.96 & 13.63 $\pm$ 20.05 \\
LLSPIN              & 84.50 $\pm$ 3.70 & 84.83 $\pm$ 4.42 & 83.36 $\pm$ 1.78 & 95.82 $\pm$ 0.60 & 89.24 $\pm$ 0.34 & 95.79 $\pm$ 1.93 & 77.62 $\pm$ 3.28 & 59.63 $\pm$ 2.04 & 14.13 $\pm$ 9.09 \\
CatBoost            & 82.30 $\pm$ 4.60 & 85.41 $\pm$ 5.10 & 84.07 $\pm$ 1.49 & 96.36 $\pm$ 0.52 & 88.15 $\pm$ 0.57 & 96.14 $\pm$ 1.63 & 77.48 $\pm$ 3.20 & 60.33 $\pm$ 2.09 & 14.63 $\pm$ 8.25 \\
Lasso               & 81.30 $\pm$ 5.30 & 85.51 $\pm$ 4.49 & 82.21 $\pm$ 1.42 & 97.30 $\pm$ 0.60 & 88.23 $\pm$ 0.69 & 97.02 $\pm$ 1.51 & 72.59 $\pm$ 4.84 & 60.97 $\pm$ 2.33 & 15.25 $\pm$ 10.98 \\
L2X                 & 78.90 $\pm$ 5.70 & 81.64 $\pm$ 6.59 & 83.71 $\pm$ 2.64 & 94.94 $\pm$ 0.84 & 87.11 $\pm$ 0.49 & 97.37 $\pm$ 1.60 & 71.60 $\pm$ 2.82 & 61.59 $\pm$ 2.48 & 15.38 $\pm$ 11.10 \\
Random Forest       & 82.50 $\pm$ 5.30 & 82.48 $\pm$ 4.45 & 84.21 $\pm$ 2.47 & 97.42 $\pm$ 0.24 & 88.88 $\pm$ 0.45 & 94.91 $\pm$ 2.84 & 76.20 $\pm$ 3.22 & 61.09 $\pm$ 2.23 & 16.25 $\pm$ 9.48 \\
SVM                 & 77.70 $\pm$ 5.40 & 86.32 $\pm$ 4.22 & 81.21 $\pm$ 1.21 & 97.18 $\pm$ 0.38 & 89.09 $\pm$ 0.40 & 96.49 $\pm$ 1.60 & 77.62 $\pm$ 2.98 & 59.24 $\pm$ 2.51 & 16.88 $\pm$ 11.78 \\
TabulaRNN           & 81.80 $\pm$ 4.90 & 82.24 $\pm$ 5.37 & 83.57 $\pm$ 1.85 & 96.02 $\pm$ 0.60 & 84.52 $\pm$ 0.92 & 93.51 $\pm$ 1.70 & 78.89 $\pm$ 3.66 & 60.52 $\pm$ 2.29 & 16.88 $\pm$ 12.65 \\
LGBM                & 82.50 $\pm$ 4.30 & 83.30 $\pm$ 5.52 & 82.93 $\pm$ 2.00 & 97.46 $\pm$ 0.40 & 89.15 $\pm$ 0.59 & 95.96 $\pm$ 2.84 & 73.10 $\pm$ 3.84 & 60.91 $\pm$ 2.21 & 17.88 $\pm$ 6.20 \\[\smallskipamount]
% Models 16-30
GBM                 & 80.70 $\pm$ 5.00 & 82.77 $\pm$ 6.14 & 84.21 $\pm$ 2.85 & 97.02 $\pm$ 0.46 & 88.72 $\pm$ 0.59 & 95.44 $\pm$ 2.40 & 75.32 $\pm$ 3.48 & 60.79 $\pm$ 2.36 & 18.00 $\pm$ 6.41 \\
XGBoost             & 81.70 $\pm$ 5.00 & 85.51 $\pm$ 4.49 & 82.64 $\pm$ 1.66 & 97.14 $\pm$ 0.36 & 88.68 $\pm$ 0.48 & 94.56 $\pm$ 2.39 & 75.44 $\pm$ 3.24 & 60.85 $\pm$ 2.43 & 18.25 $\pm$ 9.13 \\
TabR                & 79.60 $\pm$ 4.80 & 82.97 $\pm$ 5.47 & 83.71 $\pm$ 1.85 & 96.64 $\pm$ 0.54 & 85.91 $\pm$ 0.51 & 95.09 $\pm$ 1.57 & 76.34 $\pm$ 4.00 & 60.09 $\pm$ 2.39 & 19.13 $\pm$ 12.68 \\
FT-Transformer      & 81.30 $\pm$ 5.70 & 85.17 $\pm$ 4.78 & 81.86 $\pm$ 1.77 & 96.50 $\pm$ 0.58 & 84.65 $\pm$ 1.23 & 94.03 $\pm$ 3.20 & 77.48 $\pm$ 3.62 & 60.00 $\pm$ 2.23 & 19.50 $\pm$ 12.92 \\
TabM                & 81.30 $\pm$ 5.00 & 84.02 $\pm$ 4.76 & 83.00 $\pm$ 1.48 & 95.82 $\pm$ 0.78 & 85.75 $\pm$ 0.82 & 95.44 $\pm$ 1.42 & 78.23 $\pm$ 2.92 & 59.18 $\pm$ 2.46 & 19.75 $\pm$ 13.55 \\
AdaBoost            & 82.30 $\pm$ 5.40 & 80.59 $\pm$ 4.57 & 83.71 $\pm$ 1.51 & 96.82 $\pm$ 0.28 & 88.48 $\pm$ 0.51 & 95.61 $\pm$ 2.52 & 73.56 $\pm$ 3.62 & 60.24 $\pm$ 2.46 & 20.25 $\pm$ 9.02 \\
MambaTab            & 80.60 $\pm$ 4.90 & 82.28 $\pm$ 4.52 & 83.43 $\pm$ 2.20 & 95.50 $\pm$ 0.88 & 86.48 $\pm$ 0.82 & 94.56 $\pm$ 2.50 & 75.71 $\pm$ 3.84 & 59.24 $\pm$ 2.59 & 21.00 $\pm$ 12.11 \\
KNN                 & 75.40 $\pm$ 5.00 & 83.40 $\pm$ 4.45 & 80.71 $\pm$ 1.40 & 96.88 $\pm$ 0.44 & 88.32 $\pm$ 0.77 & 95.26 $\pm$ 2.69 & 76.20 $\pm$ 3.12 & 59.27 $\pm$ 2.46 & 21.38 $\pm$ 7.65 \\
TabSeq              & 80.60 $\pm$ 5.20 & 82.19 $\pm$ 5.74 & 83.14 $\pm$ 2.51 & 95.72 $\pm$ 0.82 & 86.51 $\pm$ 0.92 & 93.68 $\pm$ 2.30 & 77.76 $\pm$ 3.48 & 58.55 $\pm$ 2.31 & 21.38 $\pm$ 13.29 \\
Mambular            & 79.30 $\pm$ 5.60 & 84.21 $\pm$ 4.22 & 81.71 $\pm$ 1.85 & 94.66 $\pm$ 0.72 & 86.48 $\pm$ 0.80 & 93.33 $\pm$ 2.60 & 75.62 $\pm$ 3.26 & 59.18 $\pm$ 2.33 & 21.50 $\pm$ 11.55 \\
SAINT               & 81.50 $\pm$ 5.40 & 80.98 $\pm$ 5.06 & 82.71 $\pm$ 1.48 & 95.58 $\pm$ 0.74 & 85.49 $\pm$ 0.70 & 94.21 $\pm$ 2.03 & 77.67 $\pm$ 3.78 & 58.18 $\pm$ 2.36 & 22.25 $\pm$ 11.51 \\
MambAttention       & 79.90 $\pm$ 4.90 & 83.60 $\pm$ 5.44 & 81.43 $\pm$ 1.51 & 95.14 $\pm$ 0.62 & 85.05 $\pm$ 1.23 & 92.63 $\pm$ 2.42 & 75.39 $\pm$ 3.68 & 59.24 $\pm$ 2.14 & 22.88 $\pm$ 9.30 \\
INVASE              & 77.50 $\pm$ 4.80 & 80.93 $\pm$ 4.80 & 79.71 $\pm$ 1.48 & 95.40 $\pm$ 0.58 & 86.45 $\pm$ 0.77 & 95.79 $\pm$ 1.93 & 73.93 $\pm$ 2.26 & 57.70 $\pm$ 2.00 & 25.38 $\pm$ 11.53 \\
TabNet              & 79.10 $\pm$ 5.20 & 82.38 $\pm$ 5.44 & 83.50 $\pm$ 1.93 & 93.22 $\pm$ 0.88 & 85.01 $\pm$ 0.82 & 92.98 $\pm$ 2.60 & 77.25 $\pm$ 3.72 & 57.06 $\pm$ 3.31 & 26.63 $\pm$ 14.72 \\
ModernNCA           & 76.90 $\pm$ 5.10 & 82.09 $\pm$ 6.12 & 79.21 $\pm$ 1.67 & 94.94 $\pm$ 0.68 & 84.68 $\pm$ 0.88 & 91.05 $\pm$ 1.42 & 75.90 $\pm$ 3.12 & 56.76 $\pm$ 2.80 & 26.63 $\pm$ 16.73 \\[\smallskipamount]
% Models 31-46
Decision Tree       & 75.10 $\pm$ 5.40 & 83.11 $\pm$ 4.76 & 79.00 $\pm$ 1.51 & 95.84 $\pm$ 0.60 & 87.66 $\pm$ 0.42 & 94.74 $\pm$ 2.33 & 67.04 $\pm$ 4.16 & 58.61 $\pm$ 2.66 & 27.50 $\pm$ 10.31 \\
CategoryEmbedding   & 79.50 $\pm$ 5.80 & 81.15 $\pm$ 4.16 & 82.57 $\pm$ 1.76 & 93.42 $\pm$ 1.10 & 84.07 $\pm$ 1.32 & 93.51 $\pm$ 1.60 & 76.76 $\pm$ 3.42 & 56.36 $\pm$ 2.59 & 27.75 $\pm$ 10.63 \\
1-D CNN             & 77.10 $\pm$ 6.00 & 81.78 $\pm$ 4.22 & 82.79 $\pm$ 1.49 & 93.06 $\pm$ 0.78 & 84.87 $\pm$ 0.86 & 93.68 $\pm$ 1.70 & 76.02 $\pm$ 3.52 & 56.33 $\pm$ 2.68 & 28.63 $\pm$ 8.18 \\
AutoInt             & 78.90 $\pm$ 5.20 & 83.11 $\pm$ 5.51 & 82.14 $\pm$ 1.87 & 94.10 $\pm$ 0.80 & 82.66 $\pm$ 1.00 & 92.81 $\pm$ 2.60 & 75.85 $\pm$ 3.62 & 56.64 $\pm$ 2.48 & 29.75 $\pm$ 10.86 \\
Naive Bayes         & 74.40 $\pm$ 5.70 & 85.41 $\pm$ 4.49 & 74.07 $\pm$ 1.53 & 94.60 $\pm$ 0.82 & 85.17 $\pm$ 0.88 & 94.74 $\pm$ 2.42 & 64.32 $\pm$ 4.84 & 59.52 $\pm$ 2.21 & 31.13 $\pm$ 7.24 \\
DANets              & 78.90 $\pm$ 5.30 & 80.44 $\pm$ 6.14 & 80.64 $\pm$ 1.64 & 92.66 $\pm$ 1.02 & 82.89 $\pm$ 1.20 & 90.88 $\pm$ 2.52 & 75.76 $\pm$ 3.54 & 55.94 $\pm$ 2.77 & 32.25 $\pm$ 12.43 \\
NDTF                & 77.10 $\pm$ 5.60 & 79.60 $\pm$ 4.65 & 82.36 $\pm$ 1.55 & 92.22 $\pm$ 0.90 & 82.21 $\pm$ 1.16 & 92.63 $\pm$ 2.12 & 74.74 $\pm$ 3.88 & 55.36 $\pm$ 3.14 & 32.88 $\pm$ 10.84 \\
ResNetTabular       & 75.10 $\pm$ 6.00 & 80.78 $\pm$ 4.88 & 81.93 $\pm$ 1.17 & 91.06 $\pm$ 0.84 & 83.52 $\pm$ 0.57 & 91.93 $\pm$ 2.30 & 74.83 $\pm$ 3.64 & 56.18 $\pm$ 2.62 & 34.13 $\pm$ 4.55 \\
TANGOS              & 72.30 $\pm$ 5.90 & 81.49 $\pm$ 4.76 & 78.64 $\pm$ 1.48 & 92.66 $\pm$ 1.00 & 83.33 $\pm$ 0.69 & 92.11 $\pm$ 1.93 & 73.56 $\pm$ 3.42 & 56.24 $\pm$ 2.51 & 34.25 $\pm$ 5.52 \\
NODE                & 73.10 $\pm$ 5.40 & 81.20 $\pm$ 5.51 & 81.93 $\pm$ 1.72 & 91.30 $\pm$ 0.88 & 83.20 $\pm$ 0.77 & 92.28 $\pm$ 1.60 & 74.69 $\pm$ 3.42 & 56.09 $\pm$ 2.95 & 34.38 $\pm$ 4.66 \\
Trompt              & 73.10 $\pm$ 5.40 & 80.78 $\pm$ 4.88 & 80.21 $\pm$ 1.51 & 90.68 $\pm$ 1.50 & 83.49 $\pm$ 0.77 & 91.40 $\pm$ 2.39 & 75.16 $\pm$ 3.60 & 55.91 $\pm$ 2.62 & 35.00 $\pm$ 5.42 \\
ENODE               & 72.30 $\pm$ 5.70 & 79.84 $\pm$ 5.06 & 81.14 $\pm$ 1.64 & 89.42 $\pm$ 0.84 & 82.50 $\pm$ 0.88 & 91.23 $\pm$ 3.37 & 74.42 $\pm$ 3.96 & 55.73 $\pm$ 2.68 & 36.38 $\pm$ 4.53 \\
DeepFM              & 76.90 $\pm$ 5.60 & 82.58 $\pm$ 5.52 & 79.50 $\pm$ 1.67 & 91.84 $\pm$ 0.78 & 79.70 $\pm$ 1.62 & 90.35 $\pm$ 2.84 & 71.60 $\pm$ 3.64 & 55.64 $\pm$ 2.59 & 37.50 $\pm$ 12.29 \\
DCN                 & 70.50 $\pm$ 6.00 & 80.54 $\pm$ 5.06 & 77.86 $\pm$ 1.49 & 88.20 $\pm$ 1.56 & 82.80 $\pm$ 0.92 & 88.95 $\pm$ 2.42 & 72.68 $\pm$ 3.52 & 55.76 $\pm$ 3.08 & 39.88 $\pm$ 5.19 \\
TabPFN              & — & — & — & — & — & 97.19 $\pm$ 1.02 & — & — & 41.50 $\pm$ 12.73 \\
ProtoGate           & 64.50 $\pm$ 6.40 & 63.14 $\pm$ 5.51 & 70.14 $\pm$ 1.12 & 83.70 $\pm$ 1.22 & 74.40 $\pm$ 1.45 & 79.12 $\pm$ 3.37 & 55.90 $\pm$ 5.28 & 52.73 $\pm$ 2.51 & 44.88 $\pm$ 0.64 \\
\bottomrule
\end{tabular}
}
\end{table*}
%%%%%%
%%%% ldhss datasets
\renewcommand{\thesection}{H}
\renewcommand{\thesubsection}{H.\arabic{subsection}}
\setcounter{figure}{0}\renewcommand{\thefigure}{H.\arabic{figure}}
\setcounter{table}{0}\renewcommand{\thetable}{H.\arabic{table}}
\section{Discussion on LDHSS Benchmarks Results}
\label{app:ldhss}
%\section{Discussion on LDHSS Benchmarks Results}
Table~\ref{tab:ldhss_all_models} reports the performance of models across five low-dimensional, high-sample-size (LDHSS) datasets. These datasets emphasize large sample sizes and relatively moderate input dimensionality, placing importance on training stability, generalization under data imbalance, and computational efficiency. Tree-based ensemble methods dominate the top ranks. CatBoost achieves the highest average rank ($1.60 \pm 0.55$), consistently outperforming other models across all datasets. It scores particularly well on MiniBooNE ($94.19 \pm 0.29$), Adult ($87.32 \pm 0.37$), and Poker Hand ($80.53 \pm 0.91$), reflecting its strong inductive bias and robustness to tabular structure. XGBoost follows closely with strong performance across the board, achieving near-identical results to CatBoost on MiniBooNE and Adult, although slightly weaker on Poker Hand ($71.98 \pm 0.72$). Random Forest and LGBM also rank among the top five, with LGBM outperforming on Higgs ($72.45 \pm 0.33$), while Random Forest leads on Covertype ($89.29 \pm 0.17$). However, Random Forest underperforms on Poker Hand ($60.23 \pm 0.44$), which impacts its average rank. MLP is the top deep learning-based model in this regime, showing consistent performance across all datasets with no catastrophic failures, scoring as high as $93.34 \pm 0.32$ on MiniBooNE. GBM and TabR also demonstrate strong overall utility, with TabR achieving the highest score on Adult ($87.50 \pm 0.45$), though its high standard deviation on Covertype suggests less stability in larger feature spaces. TabSeq and TabTransformer, representing early-stage and transformer-based tabular models respectively, achieve mid-range ranks. While TabSeq performs reliably on Adult and Higgs, it struggles on Covertype and Poker Hand. TabTransformer suffers from relatively high variance on MiniBooNE ($86.48 \pm 3.42$), which penalizes its rank. SVM achieves competitive scores on Adult and MiniBooNE but is limited by relatively lower accuracy on Covertype and Poker Hand. DynaTab performs solidly on MiniBooNE ($91.23 \pm 0.80$) and Higgs ($71.93 \pm 0.35$), but underperforms on Covertype ($68.80 \pm 2.12$) and Poker Hand ($52.91 \pm 3.70$), resulting in a mid-tier average rank ($11.40 \pm 5.90$). This reflects a trade-off between expressiveness and training dynamics when scaled to very large datasets with shallow feature representations. TabNet, CategoryEmbedding, and TabM also fall into the mid-tier group. TabNet exhibits stable but non-leading results, while CategoryEmbedding and TabM struggle on datasets with complex distributions, such as Covertype and Poker Hand. KNN performs surprisingly well on Covertype ($85.15 \pm 0.25$), but shows a steep drop on Higgs ($62.02 \pm 0.23$), which lowers its overall rank. Transformer-based and neural feature selection models like FT-Transformer, SAINT, AutoInt, and Mambular show mixed results. FT-Transformer performs decently on MiniBooNE and Adult but lags on Higgs. SAINT and AutoInt display inconsistent behavior, with relatively poor generalization on Poker Hand and Covertype. Mambular and 1-D CNN models are even less competitive, with both methods showing significant drop-offs across most tasks. ModernNCA, DCN, LLSPIN, and MambaTab struggle with the scale and shallowness of LDHSS data, reflecting their architectural biases toward high-dimensional feature spaces. DeepFM, TANGOS, DANets, and NDTF exhibit similarly poor performance, often failing to surpass simpler baselines. Notably, models designed for feature selection such as REAL-X, L2X, and INVASE perform poorly in this setting, as their complexity does not translate well to LDHSS distributions. The bottom ranks are dominated by TabulaRNN, ProtoGate, and various deep hybrid models. Naive Bayes and Decision Tree perform poorly on datasets with high cardinality or class imbalance, such as Covertype and Poker Hand. ProtoGate and INVASE rank lowest, indicating their limited applicability to large-sample, shallow-feature tabular domains. Overall, the LDHSS benchmarks strongly favor traditional gradient-boosted ensembles and robust shallow models. While deep learning models like DynaTab and MLP show competitive behavior on some datasets, their rank is hindered by underperformance on datasets that require highly optimized memory or input encoding strategies. These results reaffirm the dominance of ensemble models in low-dimensional settings, while highlighting the importance of future work on scaling deep learning methods to such regimes without sacrificing stability or sample efficiency.
%%%%%%%%
\begin{table*}[t]
\caption{Performance (mean $\pm$ std) of all models across five LDHSS datasets. Models are sorted by average rank. (* = 30K subsamples taken from original dataset).}
\label{tab:ldhss_all_models}
\centering
\scriptsize
\setlength{\tabcolsep}{4pt}
\resizebox{\textwidth}{!}{%
\begin{tabular}{lcccccc}
\toprule
Model & MiniBooNE* & Covertype* & Adult & Poker Hand & Higgs* & Rank (Avg $\pm$ SD) \\
\midrule
% Top 15 models
CatBoost            & 94.19 $\pm$ 0.29 & 87.72 $\pm$ 0.14 & 87.32 $\pm$ 0.37 & 80.53 $\pm$ 0.91 & 72.73 $\pm$ 0.48 &  1.60 $\pm$ 0.55 \\
XGBoost             & 94.21 $\pm$ 0.19 & 85.47 $\pm$ 0.12 & 86.99 $\pm$ 0.48 & 71.98 $\pm$ 0.72 & 71.98 $\pm$ 0.38 &  2.60 $\pm$ 1.14 \\
Random Forest       & 93.37 $\pm$ 0.26 & 89.29 $\pm$ 0.17 & 85.74 $\pm$ 0.43 & 60.23 $\pm$ 0.44 & 71.75 $\pm$ 0.36 &  4.60 $\pm$ 2.30 \\
LGBM                & 93.93 $\pm$ 0.22 & 84.94 $\pm$ 0.21 & 87.21 $\pm$ 0.38 & 52.51 $\pm$ 4.02 & 72.45 $\pm$ 0.33 &  5.20 $\pm$ 4.49 \\
MLP                 & 93.34 $\pm$ 0.32 & 83.88 $\pm$ 0.40 & 84.87 $\pm$ 0.49 & 63.44 $\pm$ 2.47 & 71.26 $\pm$ 0.35 &  6.80 $\pm$ 3.35 \\
GBM                 & 92.74 $\pm$ 0.28 & 77.15 $\pm$ 0.28 & 86.76 $\pm$ 0.37 & 60.39 $\pm$ 0.75 & 71.06 $\pm$ 0.49 &  7.60 $\pm$ 3.51 \\
TabR                & 90.50 $\pm$ 0.30 & 81.85 $\pm$ 6.50 & 87.50 $\pm$ 0.45 & 54.80 $\pm$ 1.10 & 69.50 $\pm$ 0.40 &  8.40 $\pm$ 4.72 \\
TabSeq              & 90.28 $\pm$ 0.20 & 79.93 $\pm$ 0.10 & 84.98 $\pm$ 0.60 & 54.74 $\pm$ 0.49 & 70.63 $\pm$ 0.29 & 10.40 $\pm$ 1.14 \\
TabTransformer      & 86.48 $\pm$ 3.42 & 82.46 $\pm$ 0.48 & 85.00 $\pm$ 0.51 & 55.68 $\pm$ 1.25 & 70.47 $\pm$ 0.25 & 11.00 $\pm$ 5.39 \\
SVM                 & 88.95 $\pm$ 0.24 & 76.51 $\pm$ 0.25 & 85.12 $\pm$ 0.44 & 55.14 $\pm$ 0.43 & 69.73 $\pm$ 0.25 & 11.40 $\pm$ 3.21 \\
DynaTab (Ours)      & 91.23 $\pm$ 0.80 & 68.80 $\pm$ 2.12 & 83.26 $\pm$ 0.53 & 52.91 $\pm$ 3.70 & 71.93 $\pm$ 0.35 & 11.40 $\pm$ 5.90 \\
TabNet              & 85.32 $\pm$ 2.09 & 81.86 $\pm$ 0.62 & 84.80 $\pm$ 0.47 & 54.62 $\pm$ 0.76 & 71.77 $\pm$ 0.46 & 11.80 $\pm$ 6.38 \\
CategoryEmbedding   & 89.75 $\pm$ 0.40 & 69.50 $\pm$ 2.30 & 83.40 $\pm$ 0.40 & 50.75 $\pm$ 1.60 & 71.26 $\pm$ 0.75 & 13.40 $\pm$ 4.04 \\
TabM                & 87.90 $\pm$ 0.55 & 68.80 $\pm$ 2.30 & 86.80 $\pm$ 0.42 & 52.10 $\pm$ 1.40 & 70.22 $\pm$ 1.24 & 13.60 $\pm$ 5.59 \\
KNN                 & 88.83 $\pm$ 0.28 & 85.15 $\pm$ 0.25 & 83.07 $\pm$ 0.35 & 50.81 $\pm$ 0.61 & 62.02 $\pm$ 0.23 & 15.00 $\pm$ 6.63 \\[\smallskipamount]
% Models 16-30
FT-Transformer      & 88.60 $\pm$ 0.50 & 66.40 $\pm$ 2.55 & 83.00 $\pm$ 0.62 & 53.50 $\pm$ 1.25 & 68.10 $\pm$ 0.45 & 17.00 $\pm$ 4.74 \\
Decision Tree       & 88.93 $\pm$ 0.16 & 84.64 $\pm$ 0.22 & 81.38 $\pm$ 0.54 & 48.99 $\pm$ 2.06 & 62.76 $\pm$ 0.30 & 17.20 $\pm$ 6.91 \\
Lasso               & 85.63 $\pm$ 0.32 & 72.49 $\pm$ 0.49 & 83.65 $\pm$ 0.51 & 49.96 $\pm$ 6.05 & 64.10 $\pm$ 0.30 & 17.80 $\pm$ 3.11 \\
AdaBoost            & 90.68 $\pm$ 0.18 & 61.07 $\pm$ 1.63 & 85.30 $\pm$ 0.25 & 47.10 $\pm$ 1.50 & 67.64 $\pm$ 0.42 & 17.80 $\pm$ 10.59 \\
NODE                & 90.10 $\pm$ 0.35 & 80.90 $\pm$ 5.90 & 78.30 $\pm$ 1.00 & 40.55 $\pm$ 3.95 & 66.75 $\pm$ 0.50 & 19.80 $\pm$ 9.65 \\
SAINT               & 83.30 $\pm$ 1.20 & 65.10 $\pm$ 2.85 & 82.60 $\pm$ 0.65 & 52.10 $\pm$ 1.40 & 65.20 $\pm$ 0.55 & 20.20 $\pm$ 4.82 \\
Trompt              & 80.75 $\pm$ 1.70 & 67.95 $\pm$ 2.90 & 81.30 $\pm$ 0.75 & 50.75 $\pm$ 1.60 & 60.90 $\pm$ 0.70 & 22.20 $\pm$ 3.63 \\
AutoInt             & 89.20 $\pm$ 0.45 & 60.65 $\pm$ 4.05 & 80.20 $\pm$ 0.85 & 43.60 $\pm$ 3.10 & 63.80 $\pm$ 0.60 & 24.00 $\pm$ 7.71 \\
Mambular            & 79.20 $\pm$ 2.00 & 69.50 $\pm$ 2.25 & 81.80 $\pm$ 0.72 & 50.75 $\pm$ 1.60 & 41.75 $\pm$ 2.75 & 24.20 $\pm$ 8.87 \\
1-D CNN             & 88.60 $\pm$ 0.50 & 67.60 $\pm$ 2.30 & 69.20 $\pm$ 2.55 & 43.60 $\pm$ 3.10 & 60.90 $\pm$ 0.70 & 25.40 $\pm$ 7.57 \\
ResNetTabular       & 84.50 $\pm$ 1.00 & 64.55 $\pm$ 4.45 & 80.80 $\pm$ 0.80 & 49.30 $\pm$ 1.85 & 46.30 $\pm$ 2.00 & 25.60 $\pm$ 4.22 \\
MambaTab            & 72.25 $\pm$ 3.60 & 68.10 $\pm$ 2.75 & 79.60 $\pm$ 0.90 & 48.00 $\pm$ 2.10 & 56.10 $\pm$ 0.95 & 26.40 $\pm$ 6.02 \\
ModernNCA           & 70.30 $\pm$ 4.10 & 61.20 $\pm$ 6.50 & 82.20 $\pm$ 0.62 & 45.10 $\pm$ 2.75 & 50.50 $\pm$ 1.45 & 28.60 $\pm$ 6.19 \\
DCN                 & 80.75 $\pm$ 1.70 & 53.45 $\pm$ 6.50 & 76.80 $\pm$ 1.20 & 48.00 $\pm$ 2.10 & 46.30 $\pm$ 2.00 & 30.60 $\pm$ 6.35 \\
LLSPIN              & 80.75 $\pm$ 1.70 & 51.35 $\pm$ 7.30 & 79.60 $\pm$ 0.90 & 37.40 $\pm$ 5.00 & 50.50 $\pm$ 1.45 & 31.40 $\pm$ 6.11 \\[\smallskipamount]
% Models 31-46
MambAttention       & 61.50 $\pm$ 6.60 & 65.90 $\pm$ 3.30 & 74.30 $\pm$ 1.60 & 40.55 $\pm$ 3.95 & 48.45 $\pm$ 1.70 & 31.60 $\pm$ 5.08 \\
DeepFM              & 82.10 $\pm$ 1.40 & 62.20 $\pm$ 3.60 & 70.40 $\pm$ 2.30 & 35.70 $\pm$ 5.60 & 44.10 $\pm$ 2.35 & 31.80 $\pm$ 4.66 \\
TANGOS              & 75.90 $\pm$ 2.70 & 57.25 $\pm$ 5.15 & 74.30 $\pm$ 1.60 & 42.10 $\pm$ 3.50 & 56.10 $\pm$ 0.95 & 32.20 $\pm$ 4.49 \\
Naive Bayes         & 28.34 $\pm$ 0.04 & 9.27 $\pm$ 0.16 & 81.16 $\pm$ 0.73 & 46.92 $\pm$ 1.53 & 60.32 $\pm$ 0.23 & 32.60 $\pm$ 10.43 \\
NDTF                & 77.60 $\pm$ 2.30 & 67.10 $\pm$ 3.20 & 56.52 $\pm$ 8.68 & 34.00 $\pm$ 6.25 & 44.10 $\pm$ 2.35 & 32.80 $\pm$ 5.97 \\
LSPIN               & 77.60 $\pm$ 2.30 & 44.10 $\pm$ 10.10 & 78.30 $\pm$ 1.00 & 35.70 $\pm$ 5.60 & 48.45 $\pm$ 1.70 & 33.80 $\pm$ 5.63 \\
TabulaRNN           & 56.40 $\pm$ 8.10 & 64.55 $\pm$ 4.45 & 75.20 $\pm$ 1.45 & 39.00 $\pm$ 4.45 & 33.60 $\pm$ 4.30 & 34.00 $\pm$ 5.15 \\
DANets              & 82.10 $\pm$ 1.40 & 60.45 $\pm$ 6.90 & 71.50 $\pm$ 2.10 & 30.50 $\pm$ 9.30 & 36.50 $\pm$ 3.75 & 34.40 $\pm$ 5.90 \\
STG                 & 75.90 $\pm$ 2.70 & 62.00 $\pm$ 5.95 & 56.52 $\pm$ 5.66 & 32.30 $\pm$ 7.00 & 24.00 $\pm$ 6.40 & 36.20 $\pm$ 4.27 \\
ENODE               & 80.75 $\pm$ 1.70 & 60.10 $\pm$ 6.42 & 46.92 $\pm$ 12.10 & 30.01 $\pm$ 12.90 & 30.50 $\pm$ 4.90 & 36.80 $\pm$ 5.89 \\
REAL-X              & 53.70 $\pm$ 8.90 & 60.45 $\pm$ 6.90 & 36.48 $\pm$ 9.82 & 31.50 $\pm$ 7.60 & 27.20 $\pm$ 5.60 & 39.40 $\pm$ 3.29 \\
L2X                 & 56.40 $\pm$ 8.10 & 60.10 $\pm$ 5.80 & 46.92 $\pm$ 6.88 & 31.10 $\pm$ 7.60 & 20.50 $\pm$ 8.70 & 39.60 $\pm$ 2.19 \\
INVASE              & 66.10 $\pm$ 5.30 & 55.40 $\pm$ 5.80 & 34.68 $\pm$ 10.10 & 26.35 $\pm$ 6.45 & 20.20 $\pm$ 9.50 & 41.40 $\pm$ 2.70 \\
ProtoGate           & 47.75 $\pm$ 10.50 & 46.70 $\pm$ 9.10 & 34.56 $\pm$ 12.10 & 30.00 $\pm$ 15.30 & 20.00 $\pm$ 10.30 & 43.20 $\pm$ 0.84 \\
\bottomrule
\end{tabular}
}
\end{table*}
%
%%%%
%%%%LDLSS datasets
\renewcommand{\thesection}{I}
\renewcommand{\thesubsection}{I.\arabic{subsection}}
\setcounter{figure}{0}\renewcommand{\thefigure}{I.\arabic{figure}}
\setcounter{table}{0}\renewcommand{\thetable}{I.\arabic{table}}
\section{Discussion on LDLSS Benchmarks Results}
\label{app:ldlss}
%\section{Discussion on LDLSS Benchmarks Results}
Table~\ref{tab:ldlss_all_models} presents the performance of 46 models across six low-dimensional, low-sample-size (LDLSS) datasets, a regime where statistical robustness, overfitting control, and generalization from sparse supervision are crucial. The results exhibit large variance across models, reflecting the difficulty of learning meaningful representations from small datasets with limited feature complexity. TabPFN v2 ranks highest overall ($4.50 \pm 5.09$), achieving perfect or near-perfect scores on Monks-1 and Liver Disorder, and competitive performance on all other datasets. Its pretrained zero-shot prior enables strong few-shot generalization without the need for extensive tuning. CatBoost also performs exceptionally well ($5.33 \pm 6.41$), nearly matching TabPFN v2 across all tasks, particularly on Iris ($95.33 \pm 3.39$), Glass ($81.32 \pm 4.62$), and the two datasets with perfect performance. Tree ensembles such as GBM and Random Forest maintain strong average rankings. GBM achieves $99.31 \pm 1.38$ on Monks-1 and $100.00 \pm 0.00$ on Liver Disorder, while Random Forest performs well across the board, particularly on Hayes-Roth and Glass. However, both models are susceptible to performance fluctuations on very small datasets like Pima Indian and Hayes-Roth. TabulaRNN, AutoInt, FT-Transformer, NODE, and TabR form a strong group of neural and attention-based models adapted to tabular structure. Despite moderate variance, these models show competitive performance on most datasets. TabulaRNN and AutoInt both achieve $100.00 \pm 0.00$ on Monks-1 and Liver Disorder, while maintaining reasonable generalization elsewhere. FT-Transformer exhibits relatively high variance, especially on Hayes-Roth ($77.27 \pm 13.51$), but still ranks in the top third. DynaTab ranks mid-tier overall ($14.17 \pm 12.92$), performing solidly on Hayes-Roth ($84.29 \pm 5.35$) and Liver Disorder, but slightly underperforming on Iris and Pima Indian. While its architecture is optimized for high-dimensional and mixed-regime inputs, its inductive biases are less tailored to small tabular datasets, where simpler models or pretrained solutions excel. XGBoost, TabPFN (original), and Mambular also show strong individual results but slightly lower average ranks due to inconsistencies on certain datasets. TabPFN performs well on all but Monks-1, where it slightly lags behind v2. Mambular, despite achieving perfect scores on two datasets, underperforms on Hayes-Roth ($58.26 \pm 13.87$), reducing its overall standing. Methods like MambAttention, LSPIN, ResNetTabular, Decision Tree, and LLSPIN show stable but non-leading performance. Many of these models are not explicitly optimized for low-sample training and may overfit or underfit, especially on highly noisy or class-imbalanced datasets. TabM, INVASE, and SVM display a similar trend strong accuracy on specific datasets but large drop-offs elsewhere. Mid- to lower-ranked models, including CategoryEmbedding, TabSeq, SAINT, 1-D CNN, MLP, and MambaTab, tend to suffer from higher variance across tasks. While most achieve high scores on Monks-1 and Liver Disorder, they often struggle on Pima Indian or Glass, which contain more challenging class structures and fewer informative features. Models such as ModernNCA, TabNet, Lasso, TabTransformer, and STG demonstrate modest performance, with moderate to high instability across datasets. These models often rely on larger training sets or complex feature interactions that are difficult to capture in LDLSS scenarios. The weakest-performing models include DeepFM, DANets, NDTF, DCN, KNN, ENODE, Naive Bayes, TANGOS, L2X, and ProtoGate. ProtoGate performs especially poorly ($40.50 \pm 7.71$), with accuracy as low as $31.30 \pm 6.40$ on Glass and $65.51 \pm 4.41$ on Liver Disorder, suggesting that complex selection mechanisms or constrained model capacity hinder generalization in this regime. In summary, the LDLSS setting clearly favors pretrained models like TabPFN v2 and robust tree ensembles such as CatBoost and GBM. Simpler architectures and models with few parameters tend to generalize better than deep neural networks when data is scarce. While DynaTab remains competitive, its design favors more complex data scenarios. These results highlight the unique challenges of the LDLSS regime and the need for models that can generalize reliably from limited supervision.
%%%%%%%
\begin{table*}[t]
\caption{Performance (mean $\pm$ std) of all 46 models across six LDLSS datasets. Models are sorted by average rank.}
\label{tab:ldlss_all_models}
\centering
\scriptsize
\setlength{\tabcolsep}{4pt}
\resizebox{\textwidth}{!}{%
\begin{tabular}{lccccccc}
\toprule
Model & Iris & Pima Indian & Glass & Hayes-Roth & Monks-1 & Liver Disorder & Rank (Avg $\pm$ SD) \\
\midrule
% Top 15 models
TabPFN v2           & 97.33 $\pm$ 3.27 & 76.17 $\pm$ 1.92 & 80.82 $\pm$ 3.56 & 81.00 $\pm$ 7.70 & 100.00 $\pm$ 0.00 & 100.00 $\pm$ 0.00 &  4.50 $\pm$ 5.09 \\
CatBoost            & 95.33 $\pm$ 3.39 & 77.60 $\pm$ 1.91 & 81.32 $\pm$ 4.62 & 80.26 $\pm$ 8.60 & 100.00 $\pm$ 0.00 & 100.00 $\pm$ 0.00 &  5.33 $\pm$ 6.41 \\
GBM                 & 95.33 $\pm$ 3.39 & 76.17 $\pm$ 1.45 & 75.70 $\pm$ 1.82 & 82.56 $\pm$ 5.15 & 99.31 $\pm$ 1.38 & 100.00 $\pm$ 0.00 &  9.50 $\pm$ 6.19 \\
TabulaRNN           & 96.00 $\pm$ 3.89 & 75.78 $\pm$ 3.62 & 65.39 $\pm$ 8.24 & 80.97 $\pm$ 10.01 & 100.00 $\pm$ 0.00 & 100.00 $\pm$ 0.00 &  9.67 $\pm$ 8.89 \\
AutoInt             & 95.33 $\pm$ 3.39 & 76.04 $\pm$ 2.07 & 62.62 $\pm$ 1.23 & 82.59 $\pm$ 8.00 & 100.00 $\pm$ 0.00 & 100.00 $\pm$ 0.00 & 11.00 $\pm$ 10.64 \\
FT-Transformer      & 94.67 $\pm$ 5.42 & 77.86 $\pm$ 1.99 & 75.68 $\pm$ 3.32 & 77.27 $\pm$ 13.51 & 95.86 $\pm$ 8.28 & 100.00 $\pm$ 0.00 & 12.00 $\pm$ 10.77 \\
NODE                & 95.33 $\pm$ 4.52 & 76.56 $\pm$ 2.23 & 71.48 $\pm$ 5.44 & 65.07 $\pm$ 14.41 & 99.77 $\pm$ 0.47 & 100.00 $\pm$ 0.00 & 12.33 $\pm$ 7.74 \\
TabR                & 95.33 $\pm$ 4.52 & 75.65 $\pm$ 1.92 & 72.43 $\pm$ 4.53 & 60.51 $\pm$ 7.51 & 100.00 $\pm$ 0.00 & 100.00 $\pm$ 0.00 & 12.50 $\pm$ 10.91 \\
Random Forest       & 96.00 $\pm$ 3.89 & 74.63 $\pm$ 1.19 & 78.51 $\pm$ 3.08 & 82.59 $\pm$ 3.83 & 98.85 $\pm$ 1.26 & 100.00 $\pm$ 0.00 & 13.00 $\pm$ 17.24 \\
LGBM                & 94.67 $\pm$ 4.00 & 75.51 $\pm$ 2.49 & 78.49 $\pm$ 3.48 & 67.38 $\pm$ 10.24 & 100.00 $\pm$ 0.00 & 100.00 $\pm$ 0.00 & 13.00 $\pm$ 12.17 \\
Trompt              & 96.00 $\pm$ 3.89 & 75.78 $\pm$ 2.81 & 63.10 $\pm$ 7.76 & 82.54 $\pm$ 6.66 & 97.46 $\pm$ 2.12 & 100.00 $\pm$ 0.00 & 13.00 $\pm$ 9.76 \\
XGBoost             & 94.00 $\pm$ 3.27 & 73.17 $\pm$ 2.81 & 75.23 $\pm$ 4.82 & 83.33 $\pm$ 6.63 & 100.00 $\pm$ 0.00 & 100.00 $\pm$ 0.00 & 13.17 $\pm$ 16.40 \\
TabPFN              & 95.33 $\pm$ 4.52 & 77.21 $\pm$ 1.77 & 71.94 $\pm$ 4.08 & 81.85 $\pm$ 11.47 & 98.16 $\pm$ 1.87 & 99.71 $\pm$ 0.58 & 13.17 $\pm$ 6.74 \\
DynaTab (Ours)      & 94.07 $\pm$ 6.02 & 75.84 $\pm$ 2.11 & 75.01 $\pm$ 7.14 & 84.29 $\pm$ 5.35 & 85.45 $\pm$ 0.19 & 100.00 $\pm$ 0.00 & 14.17 $\pm$ 12.92 \\
Mambular            & 95.33 $\pm$ 3.39 & 75.65 $\pm$ 2.04 & 64.95 $\pm$ 3.62 & 58.26 $\pm$ 13.87 & 100.00 $\pm$ 0.00 & 100.00 $\pm$ 0.00 & 15.00 $\pm$ 11.73 \\[\smallskipamount]
% Models 16-30
MambAttention       & 95.33 $\pm$ 3.39 & 74.08 $\pm$ 4.18 & 61.16 $\pm$ 6.56 & 79.57 $\pm$ 8.67 & 100.00 $\pm$ 0.00 & 100.00 $\pm$ 0.00 & 15.33 $\pm$ 13.53 \\
LSPIN               & 96.00 $\pm$ 3.89 & 74.61 $\pm$ 2.32 & 65.88 $\pm$ 2.46 & 77.98 $\pm$ 6.27 & 100.00 $\pm$ 0.00 & 99.42 $\pm$ 0.71 & 15.83 $\pm$ 10.76 \\
ResNetTabular       & 96.00 $\pm$ 3.89 & 75.12 $\pm$ 3.14 & 57.01 $\pm$ 5.40 & 83.30 $\pm$ 8.92 & 88.64 $\pm$ 9.56 & 100.00 $\pm$ 0.00 & 16.00 $\pm$ 14.06 \\
Decision Tree       & 95.33 $\pm$ 3.39 & 69.52 $\pm$ 4.18 & 69.62 $\pm$ 3.90 & 83.33 $\pm$ 10.19 & 92.36 $\pm$ 6.73 & 100.00 $\pm$ 0.00 & 16.17 $\pm$ 14.37 \\
LLSPIN              & 94.00 $\pm$ 4.89 & 76.17 $\pm$ 1.96 & 70.54 $\pm$ 7.07 & 72.82 $\pm$ 8.49 & 100.00 $\pm$ 0.00 & 99.71 $\pm$ 0.58 & 16.50 $\pm$ 10.25 \\
TabM                & 96.67 $\pm$ 4.22 & 73.94 $\pm$ 5.04 & 48.54 $\pm$ 8.75 & 62.62 $\pm$ 18.69 & 100.00 $\pm$ 0.00 & 100.00 $\pm$ 0.00 & 16.67 $\pm$ 16.86 \\
INVASE              & 95.33 $\pm$ 4.52 & 74.61 $\pm$ 2.36 & 69.14 $\pm$ 5.11 & 75.78 $\pm$ 9.63 & 100.00 $\pm$ 0.00 & 98.55 $\pm$ 1.83 & 17.50 $\pm$ 9.81 \\
SVM                 & 96.00 $\pm$ 3.89 & 76.17 $\pm$ 1.73 & 68.69 $\pm$ 3.12 & 78.01 $\pm$ 6.67 & 91.43 $\pm$ 2.03 & 96.52 $\pm$ 1.48 & 18.00 $\pm$ 9.74 \\
CategoryEmbedding   & 94.00 $\pm$ 6.53 & 76.04 $\pm$ 1.73 & 62.13 $\pm$ 5.71 & 69.62 $\pm$ 10.29 & 100.00 $\pm$ 0.00 & 100.00 $\pm$ 0.00 & 18.17 $\pm$ 15.20 \\
TabSeq              & 94.00 $\pm$ 4.89 & 77.34 $\pm$ 2.55 & 65.42 $\pm$ 2.52 & 65.85 $\pm$ 12.59 & 99.31 $\pm$ 1.38 & 100.00 $\pm$ 0.00 & 18.50 $\pm$ 14.76 \\
SAINT               & 94.00 $\pm$ 3.27 & 75.78 $\pm$ 1.68 & 68.18 $\pm$ 2.86 & 58.20 $\pm$ 11.42 & 99.77 $\pm$ 0.47 & 100.00 $\pm$ 0.00 & 18.83 $\pm$ 13.32 \\
1-D CNN             & 94.00 $\pm$ 4.89 & 76.43 $\pm$ 2.16 & 57.48 $\pm$ 8.42 & 62.82 $\pm$ 9.23 & 100.00 $\pm$ 0.00 & 100.00 $\pm$ 0.00 & 19.33 $\pm$ 16.20 \\
MLP                 & 93.33 $\pm$ 5.42 & 76.82 $\pm$ 1.78 & 67.26 $\pm$ 6.56 & 58.93 $\pm$ 18.38 & 100.00 $\pm$ 0.00 & 100.00 $\pm$ 0.00 & 19.50 $\pm$ 17.10 \\
MambaTab            & 92.67 $\pm$ 6.11 & 77.21 $\pm$ 1.92 & 57.94 $\pm$ 4.82 & 65.16 $\pm$ 21.39 & 100.00 $\pm$ 0.00 & 100.00 $\pm$ 0.00 & 20.00 $\pm$ 17.66 \\
ModernNCA           & 90.67 $\pm$ 4.99 & 74.34 $\pm$ 2.16 & 75.68 $\pm$ 4.42 & 68.65 $\pm$ 13.51 & 94.01 $\pm$ 8.19 & 100.00 $\pm$ 0.00 & 20.83 $\pm$ 11.40 \\[\smallskipamount]
% Models 31-46
TabNet              & 93.33 $\pm$ 4.00 & 75.65 $\pm$ 2.48 & 66.82 $\pm$ 2.73 & 62.58 $\pm$ 7.16 & 93.54 $\pm$ 9.17 & 100.00 $\pm$ 0.00 & 21.17 $\pm$ 13.17 \\
Lasso               & 94.67 $\pm$ 4.00 & 77.47 $\pm$ 1.87 & 59.32 $\pm$ 6.02 & 58.91 $\pm$ 4.62 & 100.00 $\pm$ 0.00 & 97.10 $\pm$ 0.78 & 22.00 $\pm$ 15.15 \\
TabTransformer      & 93.33 $\pm$ 4.00 & 74.08 $\pm$ 4.02 & 66.82 $\pm$ 4.22 & 61.09 $\pm$ 17.69 & 93.08 $\pm$ 6.42 & 100.00 $\pm$ 0.00 & 22.17 $\pm$ 13.42 \\
STG                 & 92.00 $\pm$ 4.89 & 76.56 $\pm$ 2.23 & 61.66 $\pm$ 8.42 & 64.43 $\pm$ 10.42 & 97.92 $\pm$ 1.81 & 99.71 $\pm$ 0.58 & 22.50 $\pm$ 13.97 \\
AdaBoost            & 94.00 $\pm$ 4.89 & 77.21 $\pm$ 2.04 & 75.20 $\pm$ 4.40 & 48.37 $\pm$ 10.52 & 86.35 $\pm$ 7.17 & 96.52 $\pm$ 2.77 & 23.17 $\pm$ 14.36 \\
REAL-X              & 92.00 $\pm$ 8.00 & 76.95 $\pm$ 2.36 & 68.18 $\pm$ 5.84 & 62.56 $\pm$ 7.82 & 88.65 $\pm$ 12.46 & 97.68 $\pm$ 2.23 & 24.00 $\pm$ 11.01 \\
DeepFM              & 87.33 $\pm$ 6.11 & 71.48 $\pm$ 2.74 & 69.61 $\pm$ 5.48 & 62.59 $\pm$ 6.76 & 100.00 $\pm$ 0.00 & 100.00 $\pm$ 0.00 & 25.00 $\pm$ 14.60 \\
DANets              & 92.67 $\pm$ 4.00 & 75.91 $\pm$ 3.31 & 63.54 $\pm$ 6.16 & 62.59 $\pm$ 7.80 & 78.94 $\pm$ 10.90 & 99.42 $\pm$ 0.71 & 26.17 $\pm$ 11.64 \\
NDTF                & 90.00 $\pm$ 9.80 & 73.95 $\pm$ 2.47 & 60.26 $\pm$ 3.42 & 63.67 $\pm$ 11.11 & 94.01 $\pm$ 8.19 & 100.00 $\pm$ 0.00 & 26.83 $\pm$ 14.50 \\
DCN                 & 93.33 $\pm$ 4.00 & 74.60 $\pm$ 2.44 & 56.08 $\pm$ 3.62 & 57.41 $\pm$ 9.18 & 94.93 $\pm$ 5.87 & 99.42 $\pm$ 0.71 & 27.00 $\pm$ 15.52 \\
KNN                 & 97.33 $\pm$ 2.49 & 72.78 $\pm$ 1.96 & 64.43 $\pm$ 6.86 & 47.72 $\pm$ 7.72 & 83.33 $\pm$ 5.96 & 92.75 $\pm$ 2.43 & 28.00 $\pm$ 14.28 \\
ENODE               & 90.00 $\pm$ 5.66 & 75.39 $\pm$ 2.40 & 51.45 $\pm$ 6.02 & 59.38 $\pm$ 9.03 & 81.02 $\pm$ 21.62 & 99.42 $\pm$ 0.71 & 30.00 $\pm$ 13.88 \\
Naive Bayes         & 94.67 $\pm$ 4.00 & 75.52 $\pm$ 3.38 & 44.85 $\pm$ 11.48 & 66.75 $\pm$ 8.33 & 66.66 $\pm$ 2.50 & 93.62 $\pm$ 2.53 & 31.33 $\pm$ 8.16 \\
TANGOS              & 86.00 $\pm$ 6.53 & 75.52 $\pm$ 2.24 & 62.15 $\pm$ 4.66 & 60.43 $\pm$ 7.62 & 69.21 $\pm$ 17.67 & 96.52 $\pm$ 0.45 & 31.50 $\pm$ 9.85 \\
L2X                 & 85.33 $\pm$ 5.07 & 71.74 $\pm$ 2.88 & 56.53 $\pm$ 6.42 & 60.49 $\pm$ 2.90 & 75.46 $\pm$ 9.32 & 99.42 $\pm$ 0.71 & 34.67 $\pm$ 10.56 \\
ProtoGate           & 78.67 $\pm$ 7.02 & 58.45 $\pm$ 2.39 & 31.30 $\pm$ 6.40 & 55.15 $\pm$ 14.55 & 75.00 $\pm$ 16.14 & 65.51 $\pm$ 4.41 & 40.50 $\pm$ 7.71 \\
\bottomrule
\end{tabular}
}
\end{table*}
%%%%%
\renewcommand{\thesection}{J}
\renewcommand{\thesubsection}{J.\arabic{subsection}}
\setcounter{figure}{0}\renewcommand{\thefigure}{J.\arabic{figure}}
\setcounter{table}{0}\renewcommand{\thetable}{J.\arabic{table}}
\section{Extended Analysis for Statistical Significance}
\label{app:stat}
%\section{Extended Analysis for Statistical Significance}
%%%%%%
\subsection{Statistical Significance for HDLSS}  
We evaluated top six models on HDLSS datasets using average rank analysis (Fif.~\ref{fig:hldss_cd_significance}). DynaTab achieved the best overall performance with the lowest average rank (2.12), followed by Lasso (3.06), ProtoGate (3.50), and MLP (3.56). TabulaRNN (4.19) and LGBM (4.56) performed worse in comparison. A critical difference (CD) of 2.40 was computed using the Nemenyi test at $\alpha = 0.05$. Models connected by horizontal lines fall within this CD and thus do not differ significantly. Pairwise comparisons with DynaTab, corrected using the Holm method, revealed that only TabulaRNN was significantly worse ($p < 0.05$), denoted by an asterisk. The overall Friedman test statistic ($\chi^2 = 8.5$, $p = 0.13$) was not significant, indicating no strong evidence to reject the null hypothesis of equal performance across models (see supplementary material for details on other dataset categories and figures).
%%%%%
%%%%%
\subsection{Statistical Significance for LDHSS}
\begin{figure*}[t]
\centering

% -------- Row 1: HDLSS + HDHSS + LDHSS --------
\begin{subfigure}[t]{0.32\textwidth}
  \centering
  \includegraphics[width=\linewidth]{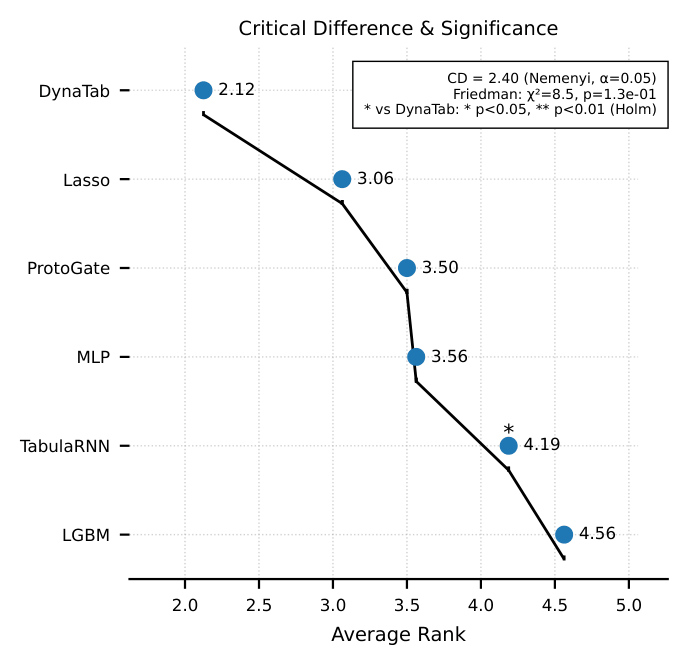}
  \caption{HDLSS}
  \label{fig:hldss_cd_significance}
\end{subfigure}\hfill
\begin{subfigure}[t]{0.32\textwidth}
  \centering
  \includegraphics[width=\linewidth]{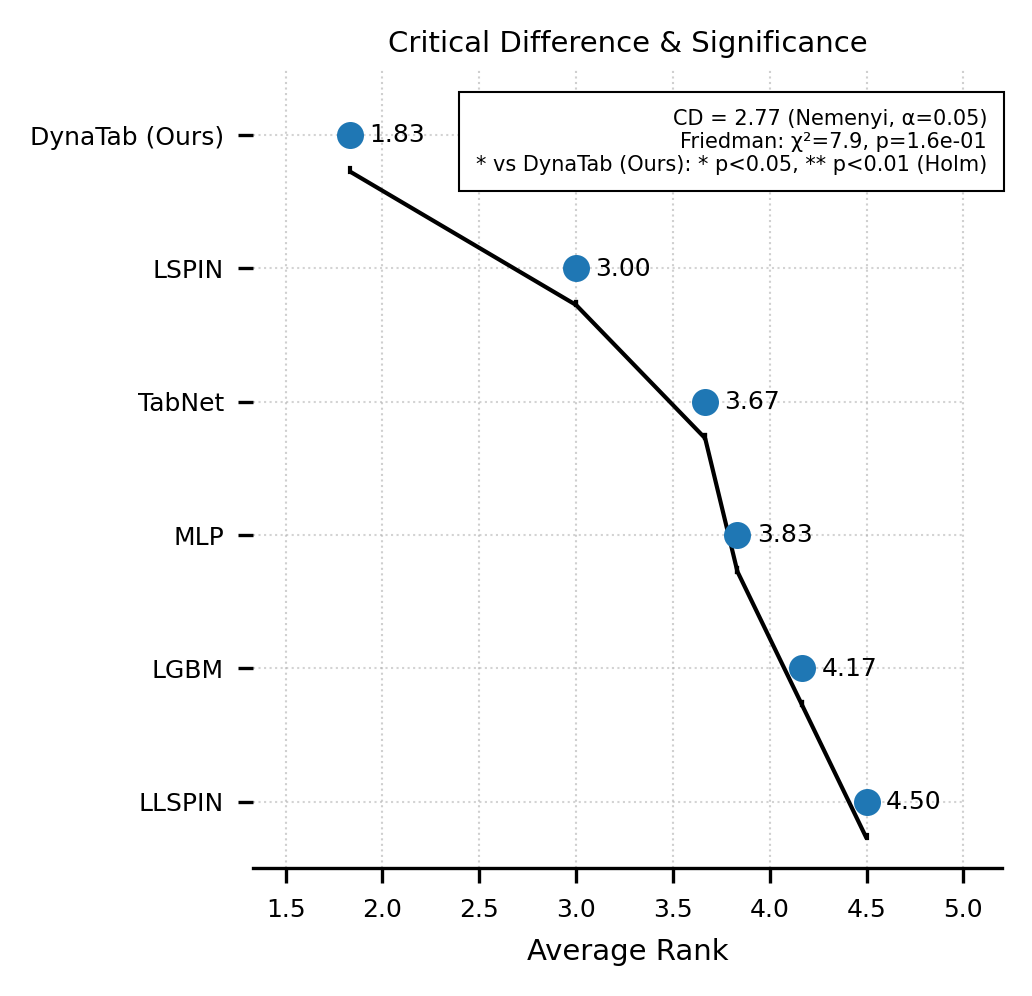}
  \caption{HDHSS}
  \label{fig:cd_hdhss}
\end{subfigure}\hfill
\begin{subfigure}[t]{0.32\textwidth}
  \centering
  \includegraphics[width=\linewidth]{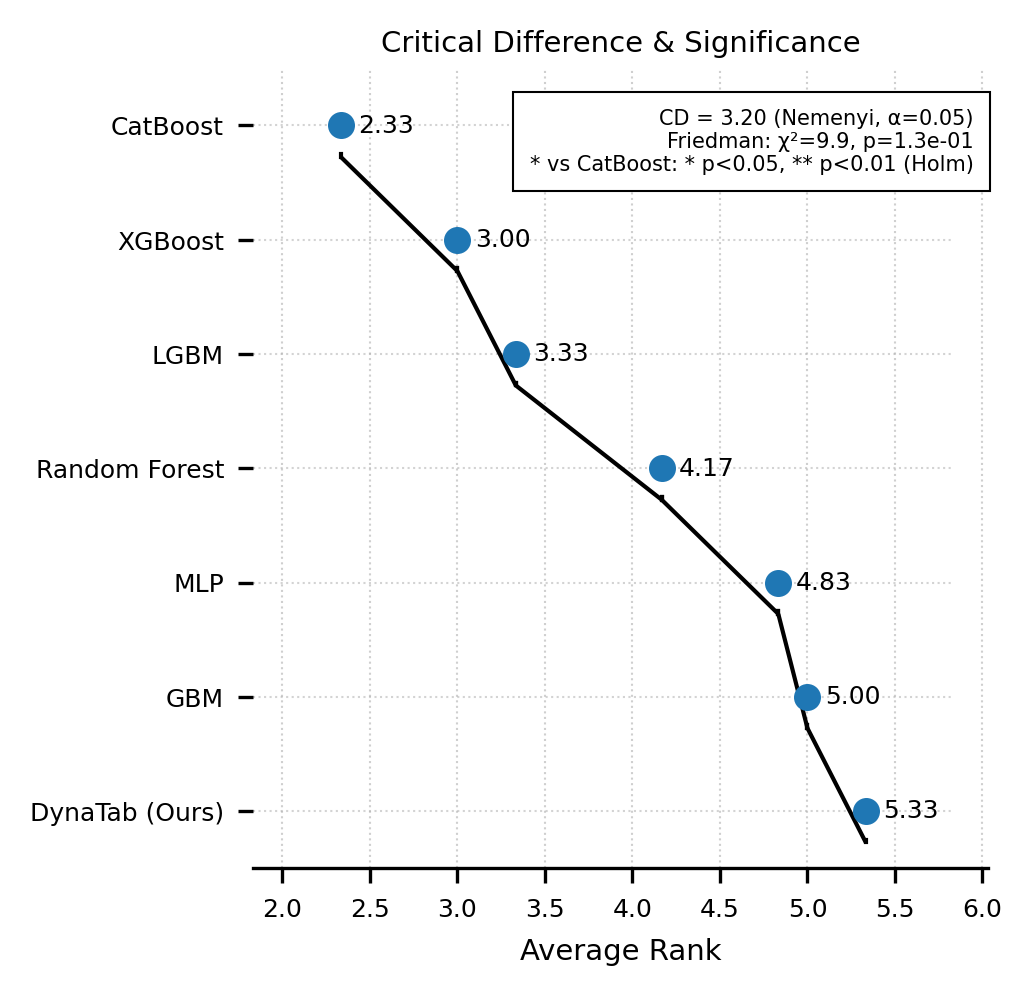}
  \caption{LDHSS}
  \label{fig:cd_ldhss}
\end{subfigure}

\vspace{2mm}

% -------- Row 2: LDLSS + Mixed Regime + (empty) --------
\begin{subfigure}[t]{0.32\textwidth}
  \centering
  \includegraphics[width=\linewidth]{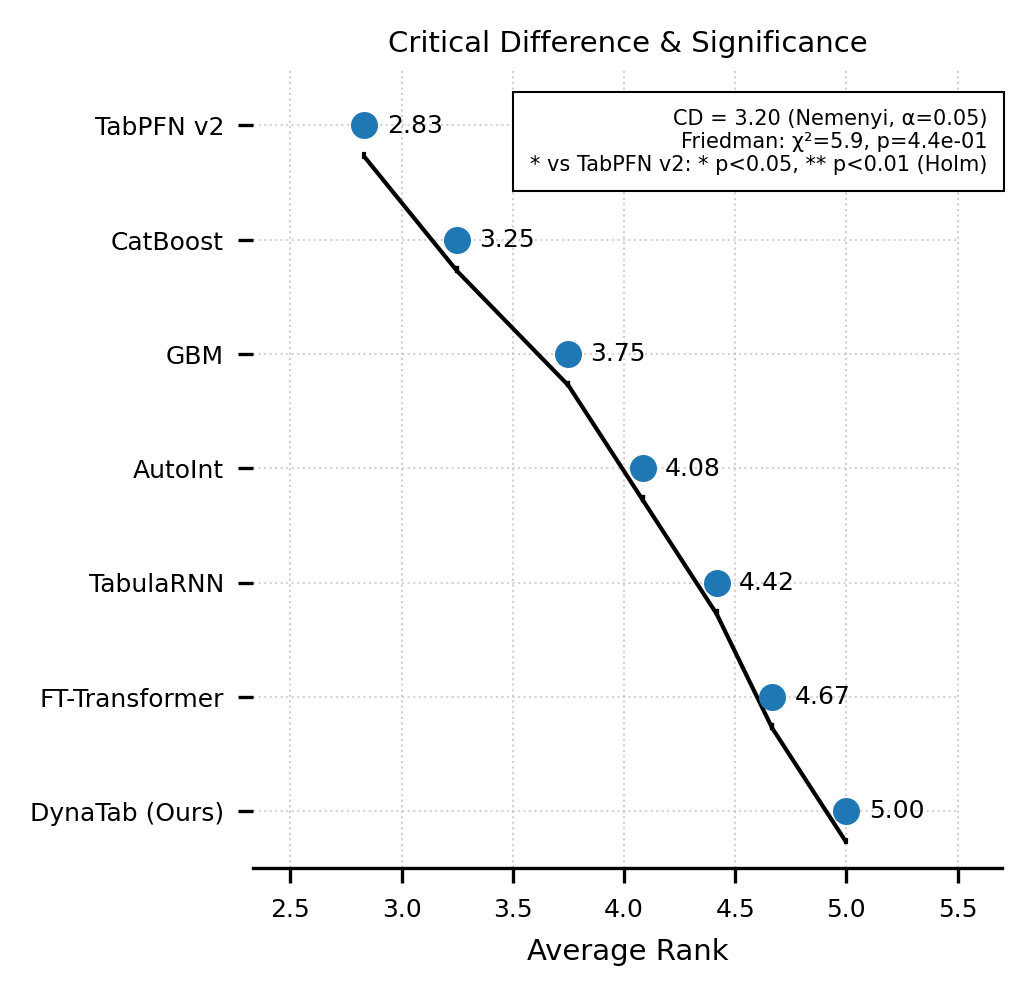}
  \caption{LDLSS}
  \label{fig:cd_ldlss}
\end{subfigure}\hfill
\begin{subfigure}[t]{0.32\textwidth}
  \centering
  \includegraphics[width=\linewidth]{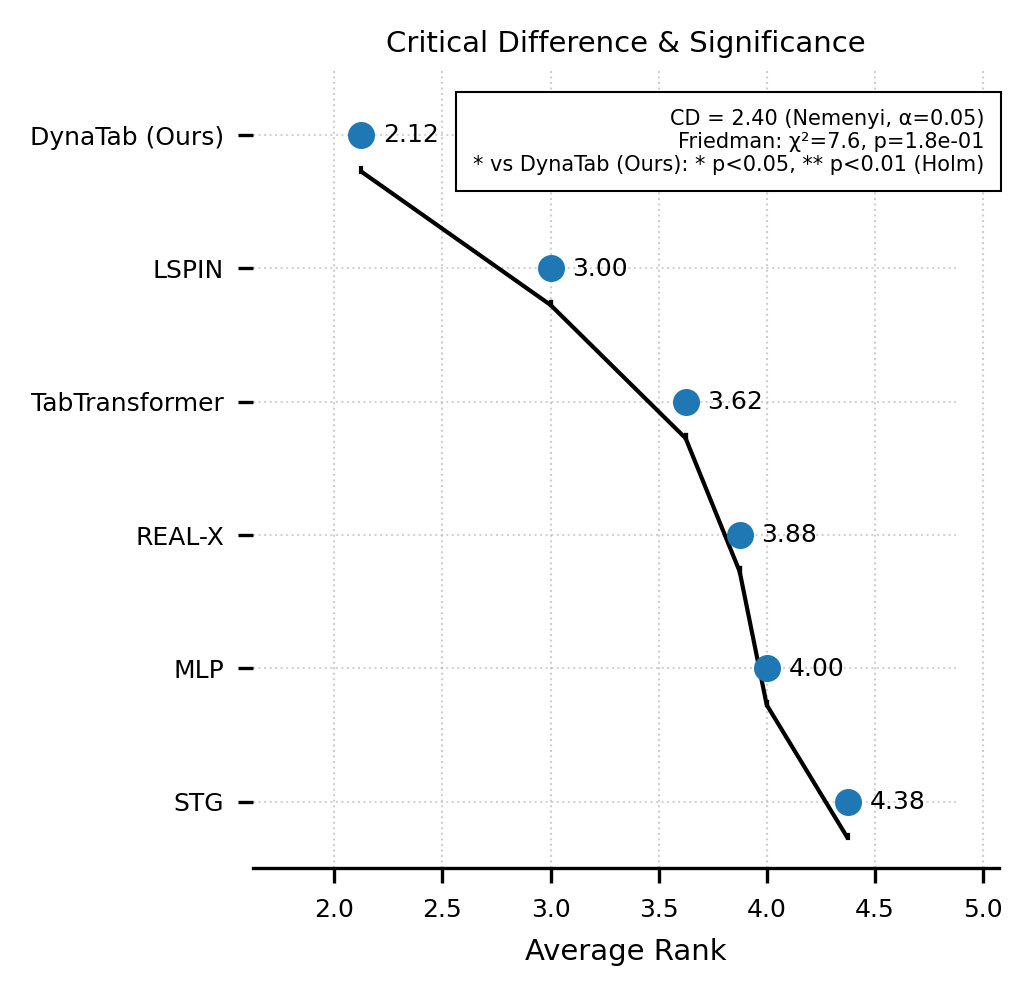}
  \caption{Mixed Regime}
  \label{fig:cd_mr}
\end{subfigure}\hfill
\begin{subfigure}[t]{0.32\textwidth}
  \centering
  % intentionally empty to keep 2x3 grid balanced
\end{subfigure}

\caption{Critical difference diagrams for different dataset regimes (top-6 ranks). Non-significant ties (Nemenyi) and pairwise significance (Wilcoxon-Holm) are indicated.}
\label{fig:cd_all_regimes}
\end{figure*}

We evaluated the top six models on LDHSS datasets using average rank analysis (Figure~\ref{fig:cd_ldhss}). DynaTab scored the lowest average rank (5.33), slightly trailing behind GBM (5.00) and MLP (4.83). CatBoost ranked first overall (2.33), followed by XGBoost (3.00) and LGBM (3.33). A critical difference (CD) of 3.20 was computed using the Nemenyi test at $\alpha = 0.05$. Several models, including DynaTab and GBM, fall within this CD range, indicating no statistically significant difference in performance. Pairwise Holm-corrected Wilcoxon tests confirmed that none of the models were significantly worse than CatBoost at the 0.05 level. The Friedman test yielded $\chi^2 = 9.9$ with $p = 0.13$, indicating no statistically significant difference in overall rankings.
%%%%%
%%%%%
\subsection{Statistical Significance for LDLSS}
On LDLSS datasets (Figure~\ref{fig:cd_ldlss}), DynaTab obtained the highest average rank (5.00), while TabPFN v2 achieved the lowest (2.83), followed closely by CatBoost (3.25). GBM (3.75) and AutoInt (4.08) formed the mid-cluster. A critical difference (CD) of 3.20 was used for significance testing with $\alpha = 0.05$. According to the Nemenyi test, no model pairs exceeded this CD, indicating statistical similarity. Pairwise Wilcoxon tests against TabPFN v2, adjusted via the Holm method, revealed no significant differences at $p < 0.05$. The overall Friedman test was also non-significant ($\chi^2 = 5.9$, $p = 0.44$), reinforcing the lack of strong statistical separation among models.
%%%%%%
\subsection{Statistical Significance for Mixed Regime}
For mixed-regime datasets (Figure~\ref{fig:cd_mr}), DynaTab achieved the best average rank (2.12), followed by LSPIN (3.00) and TabTransformer (3.62). STG, REAL-X, and MLP clustered near the bottom, with ranks ranging from 3.88 to 4.38. The Nemenyi test yielded a CD of 2.40 at $\alpha = 0.05$, under which only STG was significantly worse than DynaTab in pairwise comparison. Holm-corrected Wilcoxon tests confirmed statistical significance between DynaTab and STG ($p < 0.05$). The Friedman test statistic was $\chi^2 = 7.6$ with $p = 0.18$, indicating no overall significant difference, but notable relative advantages for DynaTab.
%%%%
\subsection{Statistical Significance for HDHSS}
On HDHSS benchmarks (Figure~\ref{fig:cd_hdhss}), DynaTab again led with the lowest average rank (1.83), followed by LSPIN (3.00), TabNet (3.67), and MLP (3.83). LGBM and LLSPIN placed slightly lower with ranks above 4. A critical difference of 2.77 was calculated via the Nemenyi test ($\alpha = 0.05$). Several models fall outside this CD range compared to DynaTab. Pairwise Wilcoxon–Holm tests confirmed that LGBM and LLSPIN were significantly worse than DynaTab at $p < 0.05$. The Friedman test result ($\chi^2 = 7.9$, $p = 0.16$) indicates moderate but not statistically significant evidence of rank differences across models.
%%%%%
%%%%%%
\renewcommand{\thesection}{K}
\renewcommand{\thesubsection}{K.\arabic{subsection}}
\setcounter{figure}{0}\renewcommand{\thefigure}{K.\arabic{figure}}
\setcounter{table}{0}\renewcommand{\thetable}{K.\arabic{table}}
\section{Extended Analysis on Ablation Studies}
\label{app:abal}
\begin{figure*}[t]
\centering

\begin{subfigure}[t]{0.49\textwidth}
  \centering
  \includegraphics[width=\linewidth]{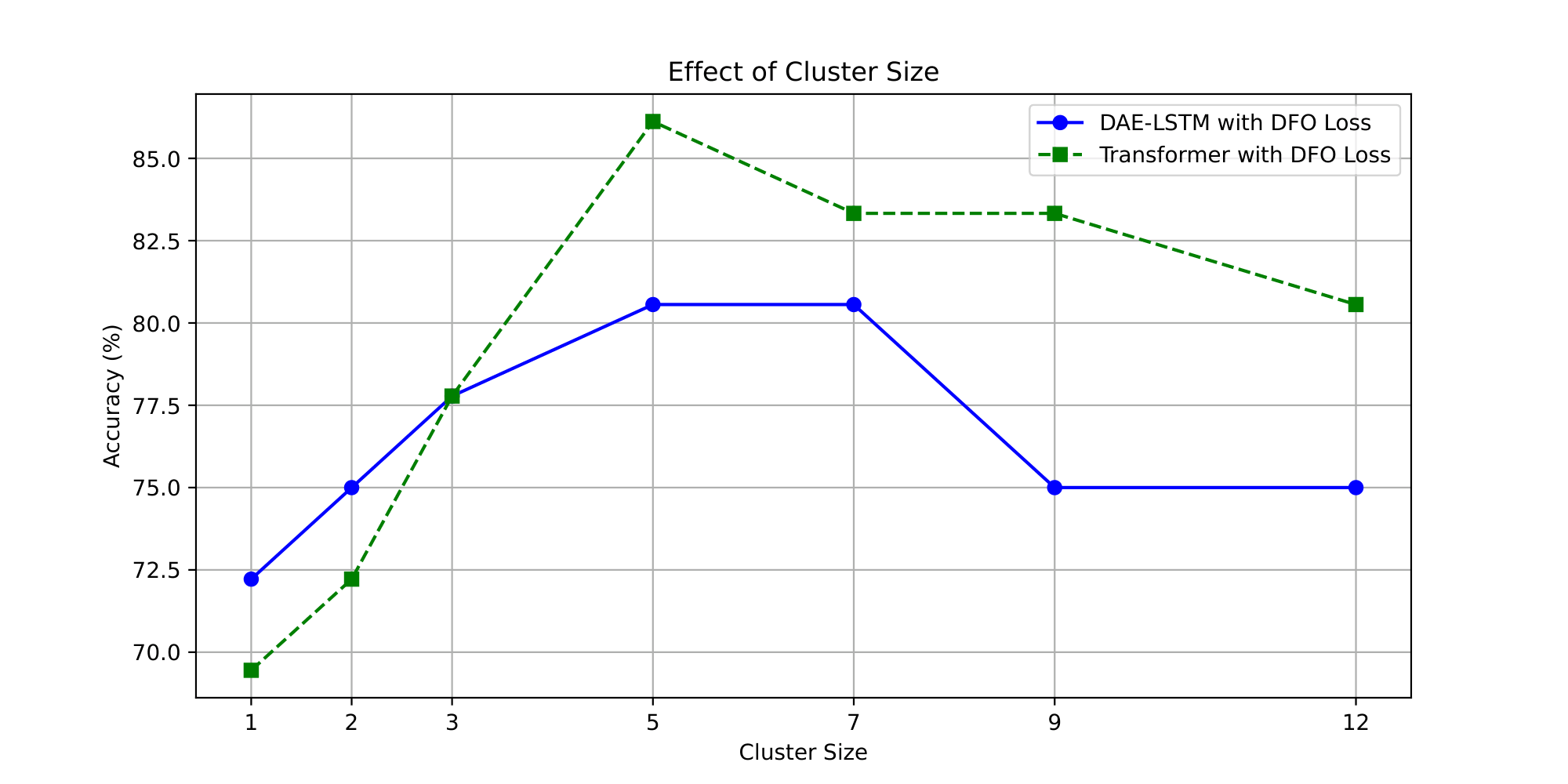}
  \caption{}
  \label{fig:ablation_cluster}
\end{subfigure}\hfill
\begin{subfigure}[t]{0.49\textwidth}
  \centering
  \includegraphics[width=\linewidth]{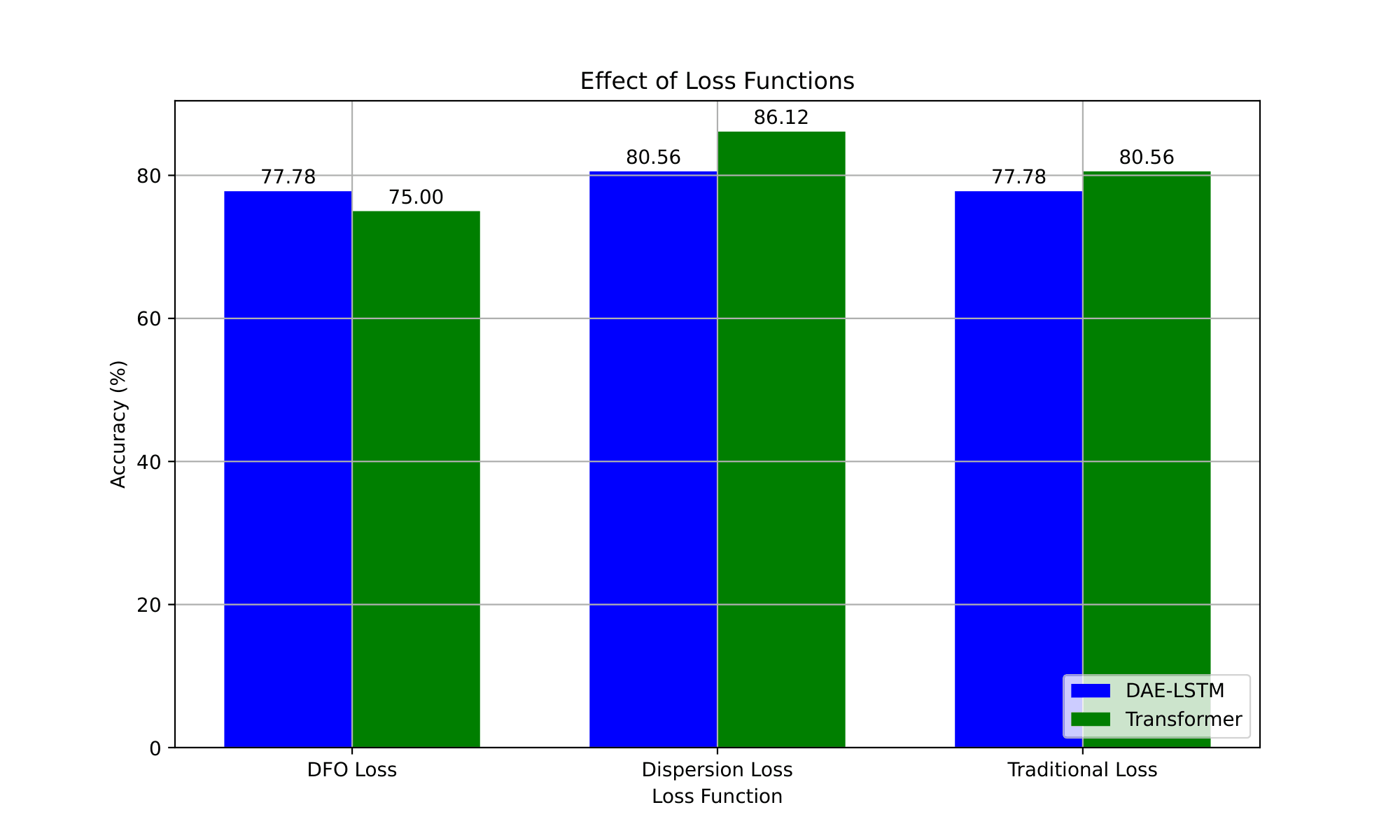}
  \caption{}
  \label{fig:ablation_loss}
\end{subfigure}

\vspace{-1mm}

\begin{subfigure}[t]{0.49\textwidth}
  \centering
  \includegraphics[width=\linewidth]{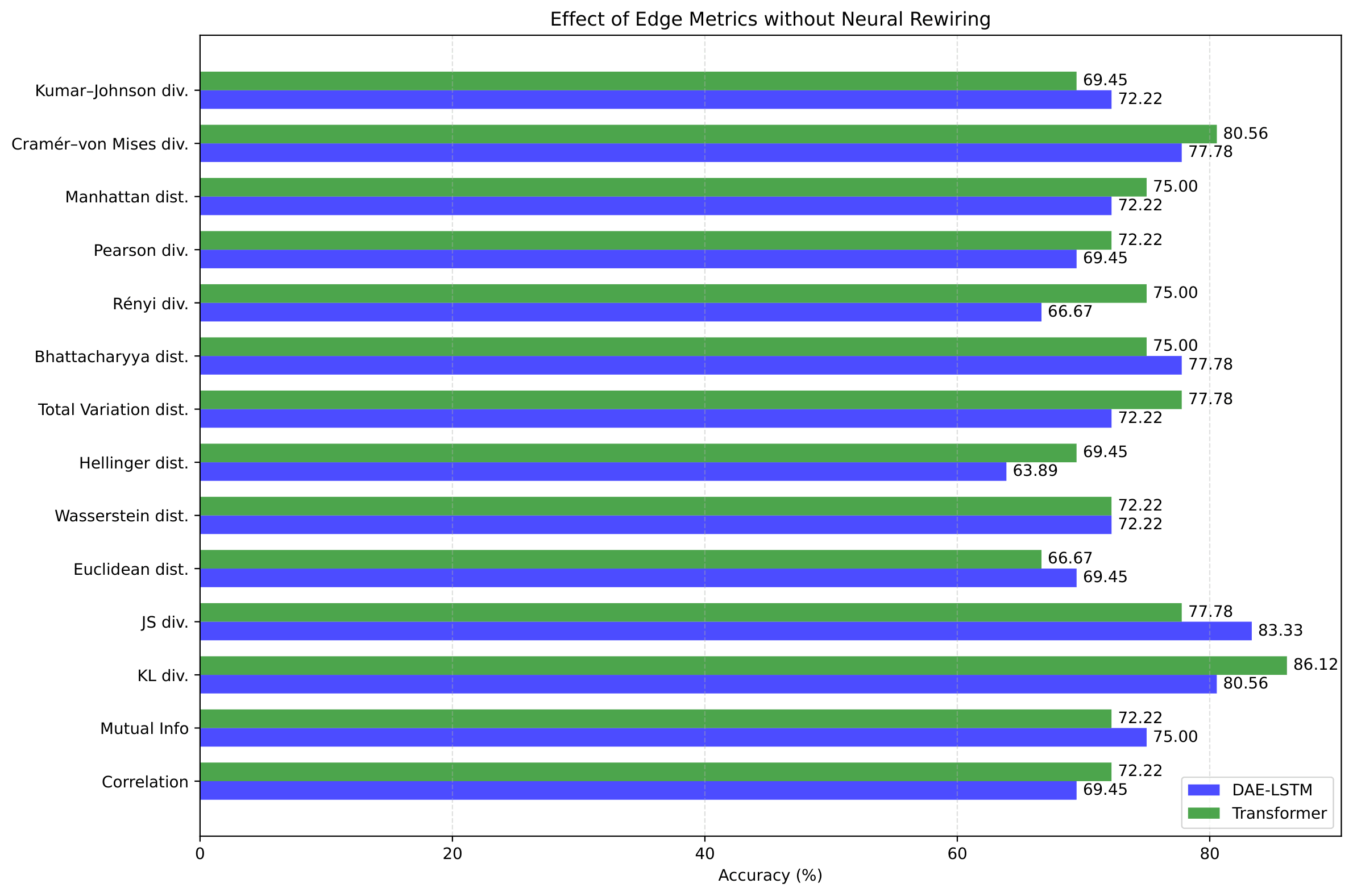}
  \caption{}
  \label{fig:ablation_edge}
\end{subfigure}\hfill
\begin{subfigure}[t]{0.49\textwidth}
  \centering
  \includegraphics[width=\linewidth]{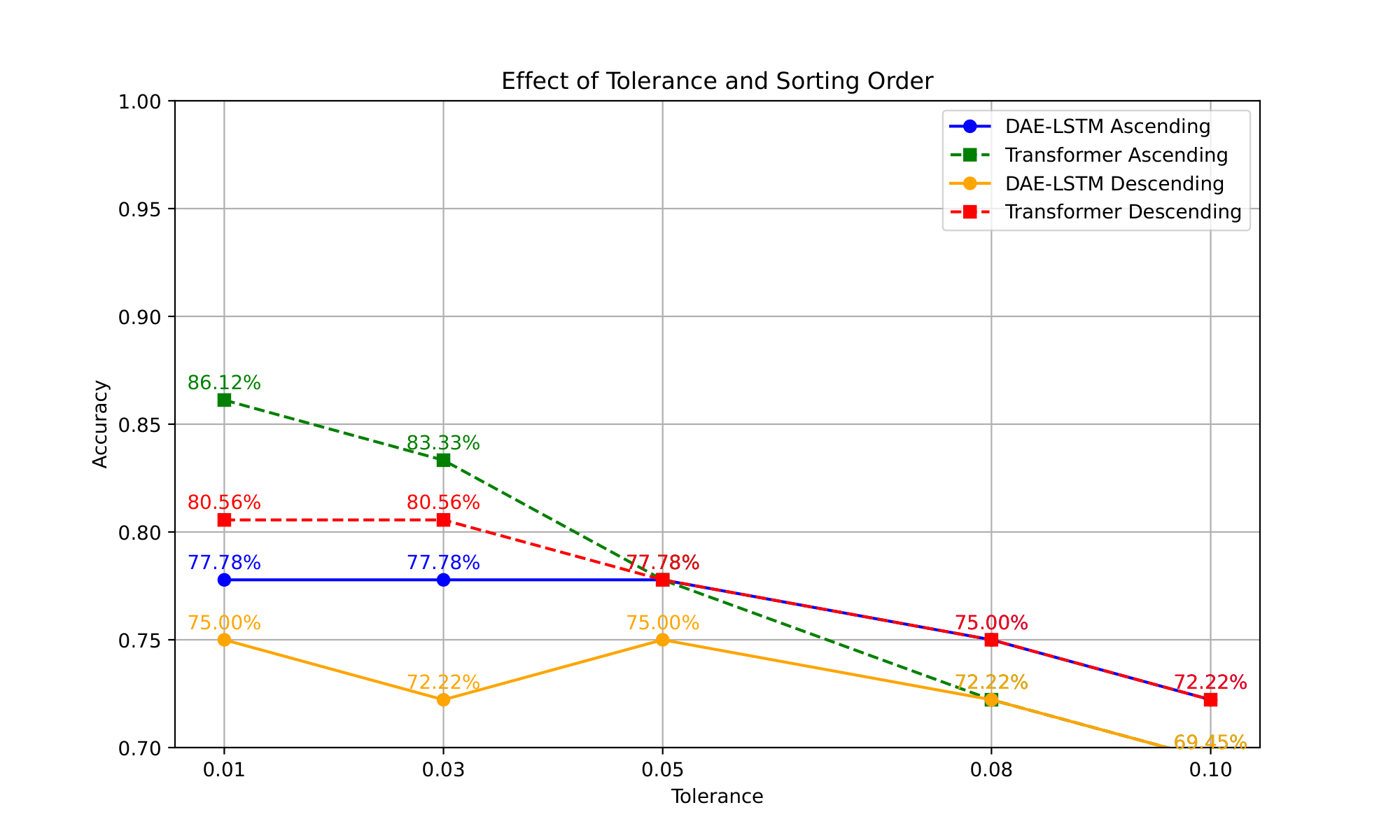}
  \caption{}
  \label{fig:ablation_tolerance}
\end{subfigure}

\vspace{-2mm}
\caption{Ablations: (a) effect of cluster size on accuracy (cluster size 5 is optimal); (b) comparison of loss functions (Dispersion and DFO outperform traditional losses); (c) accuracy under different edge metrics without neural rewiring (KL divergence is most effective); (d) impact of tolerance threshold and sorting order (lower tolerance and ascending order are more robust).} %\textcolor{red}{Use Uppercase for first letters for the names, or acronyms, eg, KL, JS, Wasserstein, etc. Use "dist." and "div." -- it's better to see the names (e.g. "Wasserstein" or "Battacharya" )  , than to see say "divergence". }}
\label{fig:ablation_four}
\end{figure*}

%\section{Extended Analysis on Ablation Studies}
We conduct controlled ablation studies on DynaTab using the AI-D Case 5 dataset to assess the impact of core design components. Each setting is evaluated on both Transformer and DAE-MHA-LSTM backbones.\newline
%%%
\begin{figure}[t]
  \centering
  \includegraphics[width=0.75\linewidth]{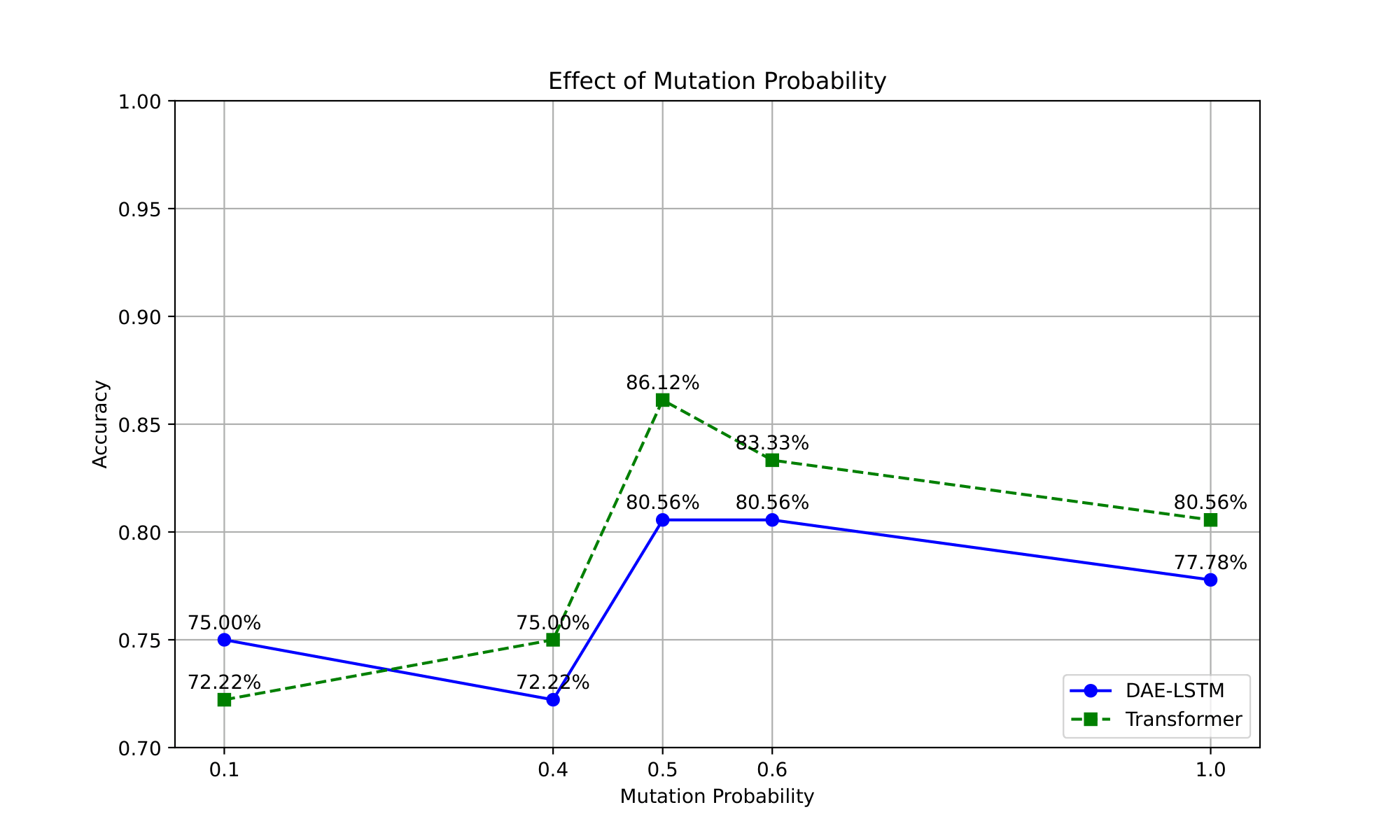}
  \caption{Effect of mutation probability on performance.}
  \label{fig:ablation_mutprob}
  \vspace{-2mm}
\end{figure}
%%%%%
\textbf{Mutation Probability.} Figure~\ref{fig:ablation_mutprob} illustrates that a mutation probability of 0.5 is optimal, yielding 86.12\% for Transformer and 80.56\% for DAE-MHA-LSTM. Lower probabilities (e.g., 0.1) under-explore feature permutations, while higher values (e.g., 1.0) destabilize learning.\newline
%%%%%%%
\textbf{Cluster Size.} As shown in Figure~\ref{fig:ablation_cluster}, a cluster size of 5 yields the highest accuracy across both backbones, balancing feature grouping granularity and model generalization. Performance drops for very small or large clusters, confirming the necessity of moderate grouping for efficient relational encoding.\newline
%%%%%%%
\textbf{Loss Functions.} Figure~\ref{fig:ablation_loss} shows that our proposed Dispersion Loss and DFO Loss outperform traditional objectives. For example, with the Transformer backbone, Dispersion Loss achieves 86.12\% accuracy versus 80.56\% with traditional loss. These custom objectives enhance ordered feature coherence and stabilize learning.\newline
%%%%%
%%%%%%%
\textbf{Edge Metrics without Rewiring.} In Figure~\ref{fig:ablation_edge}, KL divergence emerges as the most effective edge metric, reaching 86.12\% accuracy on the Transformer backbone. Other metrics like Bhattacharyya and Total Variation achieve close but lower performance. Simpler metrics (e.g., Euclidean, Hellinger) underperform, indicating the benefit of information-theoretic divergence for capturing inter-feature relations.\newline
%%%%%
\textbf{Tolerance and Sorting Order.} As seen in Figure~\ref{fig:ablation_tolerance}, lower tolerance values (e.g., 0.03) consistently yield better accuracy across both orderings and backbones. Ascending order slightly outperforms descending in most settings. These results suggest that stricter dispersion thresholds help converge toward stable and informative feature orders.\newline
%%%%%
\textbf{Computational Time for Dynamic Feature Ordering}
We benchmark the runtime of different GPU-enabled similarity metrics used in DynaTab’s dynamic feature ordering module. Table~\ref{tab:timing} reports the execution time (in seconds) for cluster sizes 7, 9, 12, and 15 on the AI-D Case-5 dataset ($316 \times 393$ matrix). Traditional metrics such as variance and correlation are efficient, completing in under a second across all settings. In contrast, KL divergence, while the most effective (see Figure~\ref{fig:ablation_edge}), is computationally expensive even with GPU support. KL divergence takes 107 seconds at cluster size 7 and grows to 229 seconds at size 15. On CPU, the same operation requires over 7 hours. This highlights the tradeoff between metric precision and computational cost in DynaTab’s feature ordering.
%%%%%%

%%%% time
%Having the supplementary compiled together with the main paper means that:
% 
%\begin{itemize}
%\item The supplementary can back-reference sections of the main paper, for example, we can refer to \cref{sec:intro};
%\item The main paper can forward reference sub-sections within the supplementary explicitly (e.g. referring to a particular experiment); 
%\item When submitted to arXiv, the supplementary will already included at the end of the paper.
%\end{itemize}
% 
%To split the supplementary pages from the main paper, you can use \href{https://support.apple.com/en-ca/guide/preview/prvw11793/mac#:~:text=Delete%20a%20page%20from%20a,or%20choose%20Edit%20%3E%20Delete).}{Preview (on macOS)}, \href{https://www.adobe.com/acrobat/how-to/delete-pages-from-pdf.html#:~:text=Choose%20%E2%80%9CTools%E2%80%9D%20%3E%20%E2%80%9COrganize,or%20pages%20from%20the%20file.}{Adobe Acrobat} (on all OSs), as well as \href{https://superuser.com/questions/517986/is-it-possible-to-delete-some-pages-of-a-pdf-document}{command line tools}.
%\bibliography{aaai2026}

\end{document}